\documentclass[10pt,twocolumn,letterpaper]{article}

\usepackage[pagenumbers]{cvpr} 

\definecolor{cvprblue}{rgb}{0.21,0.49,0.74}
\usepackage[pagebackref,breaklinks,colorlinks,allcolors=cvprblue]{hyperref}

\usepackage{multirow}
\usepackage{pifont}
\usepackage{fontawesome}
\usepackage{booktabs}      
\usepackage{makecell}      
\usepackage[table]{xcolor} 

\usepackage[notransparent]{svg}
\NewDocumentCommand{\inlineimage}{O{0.5} m}{%
  \raisebox{-0.2\baselineskip}{\includegraphics[height=#1\baselineskip]{#2}}\hspace{-3pt}
}

\newcommand{\logo}{\raisebox{0\height}{\inlineimage[1.0]{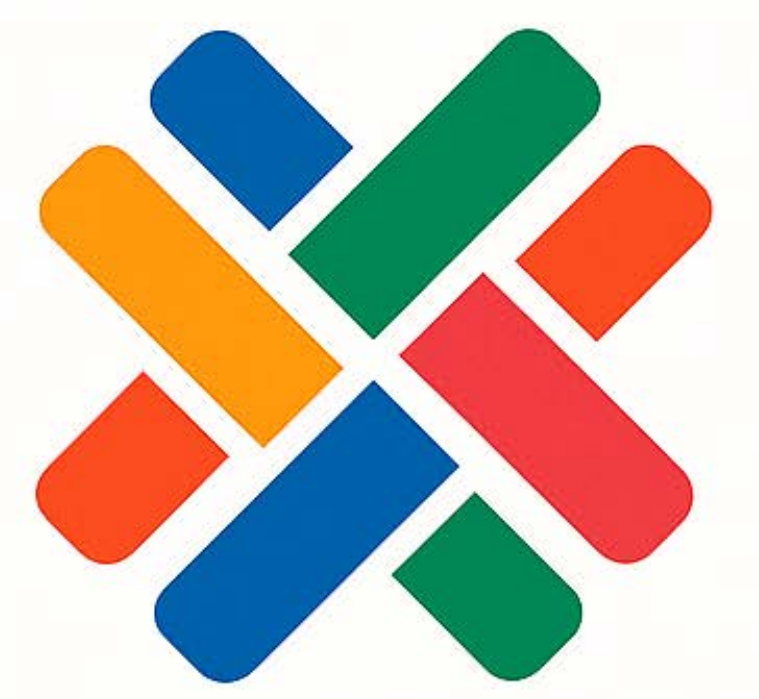}}}

\definecolor{weaveW}{HTML}{4A9FF5}  
\definecolor{weaveE}{HTML}{F7B731}  
\definecolor{weaveA}{HTML}{F4845F}  
\definecolor{weaveV}{HTML}{4ECDC4}  
\definecolor{my_green}{RGB}{51,102,0}
\definecolor{my_red}{RGB}{204, 0, 0}
\renewcommand{\checkmark}{\textcolor{my_green}{\ding{51}}} 
\newcommand{\crossmark}{\textcolor{my_red}{\ding{55}}} 

\def\paperID{1873} 
\def\confName{CVPR}
\def\confYear{2026}

\newcommand{\name}{Weave}
\newcommand{\colorname}{\textcolor{weaveW}{W}%
\textcolor{weaveE}{E}%
\textcolor{weaveA}{A}%
\textcolor{weaveV}{V}%
\textcolor{weaveE}{E}%
}
\newcommand{\trainname}{\colorname{}-100k}
\newcommand{\testname}{\colorname{}Bench}

\title{\logo \colorname{}: Unleashing and Benchmarking the In-context Interleaved Comprehension and Generation}

\author{Wei Chow$^1$\footnotemark[1] \quad Jiachun Pan$^1$\footnotemark[1] \quad Yongyuan Liang$^3$ \quad Mingze Zhou$^4$ \quad Liyu Jia$^2$ \\
Saining Zhang$^2$ \quad Xue Song$^2$ \quad Siliang Tang$^4$ \quad Juncheng Li$^4$ \quad Fengda Zhang$^2$\footnotemark[2] \\
Weijia Wu$^1$\footnotemark[2] \quad Hanwang Zhang$^2$ \quad Tat-Seng Chua$^1$ \\
\small $^1$National University of Singapore, $^2$Nanyang Technological University, \\
\small $^3$University of Maryland, College Park, $^4$Zhejiang University \\
{\faGlobe}\href{https://weichow23.github.io/weave/}{\textcolor{my_red}{\textsl{https://weichow23.github.io/weave}}}
}

\begin{document}
\maketitle
\renewcommand{\thefootnote}{\fnsymbol{footnote}} 
\footnotetext[1]{Equal Contribution.} \footnotetext[2]{Corresponding Author.}
\renewcommand{\thefootnote}{\arabic{footnote}}

\vspace{-0.5cm}
\begin{abstract}
Recent advances in unified multimodal models (UMMs) have enabled impressive progress in visual comprehension and generation.
However, existing datasets and benchmarks focus primarily on single-turn interactions, failing to capture the multi-turn, context-dependent nature of real-world image creation and editing.
To address this gap, we present \colorname{}, the first suite for in-context interleaved cross-modality comprehension and generation.
Our suite consists of two complementary parts. 
\trainname{} is a large-scale dataset of $100$K interleaved samples spanning over $370$K dialogue turns and $500$K images, covering comprehension, editing, and generation tasks that require reasoning over historical context. \testname{} is a human-annotated benchmark with $100$ tasks based on $480$ images, featuring a hybrid VLM judger evaluation framework based on both the reference image and the combination of the original image with editing instructions that assesses models' abilities in multi-turn generation, visual memory, and world-knowledge reasoning across diverse domains.
Experiments demonstrate that training on \trainname{} enables vision comprehension, image editing, and comprehension-generation collaboration capabilities. Furthermore, it facilitates UMMs to develop emergent visual-memory capabilities, while extensive evaluations on \testname{} expose the persistent limitations and challenges of current approaches in multi-turn, context-aware image generation and editing.
We believe \colorname{} provides a view and foundation for studying in-context interleaved comprehension and generation for multi-modal community.
\end{abstract}    
\section{Introduction}\label{sec:intro}
\begin{figure*}[t]
    \centering
    \vspace{-0.3cm}
    \includegraphics[width=0.99\linewidth]{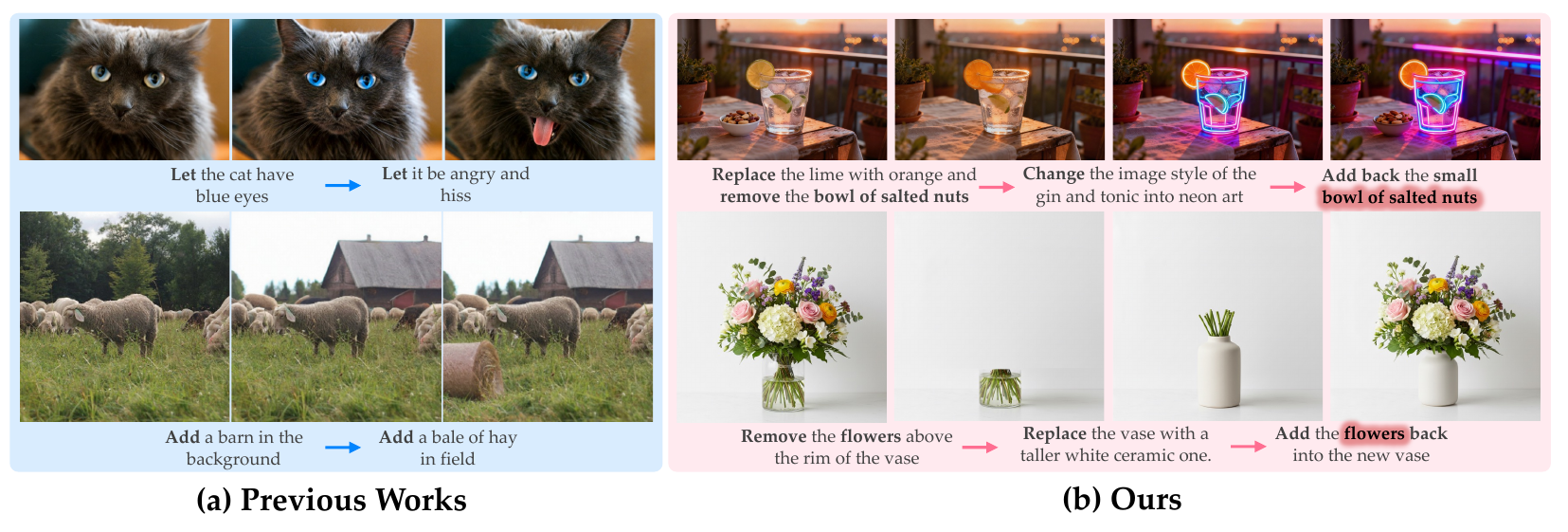}
    \vspace{-0.3cm}
    \caption{\textbf{Comparisons among and existing datasets}. \textbf{(a) Previous Works:}  Simple overlay of single-turn edits. \textbf{(b) Ours:} Multi-turn edits involving in-context visual memory recall. }
    \label{fig-comparison}
    \vspace{-0.5cm}
\end{figure*}

Recent advances in unified multimodal models (UMMs)~\citep{deng2025bagel,xie2025show,zhou2024transfusion,ma2025janusflow} have significantly reshaped the landscape of visual understanding and generation.
This unified formulation enables models to describe and edit visual content through language, integrate visual references, and perform iterative editing across multiple images.
Recent works have shown its remarkable potential for image editing~\citep{liu2025step1x,song2025query,wang2025seededit}, and multi-image composition~\citep{seedream2025seedream,xia2025dreamomni2}. 

However, real-world image creation is rarely a one-shot process. Human creators typically engage in reversible refinement, reusing or reverting to previous results as needed. Moreover, creating a comic or visual story inherently involves multiple rounds of progressive refinement to maintain visual consistency, where each frame must remain coherent with previous scenes in terms of character appearance, lighting, and narrative flow~\citep{huang2025memory,hendrycks2025definition}.

While some closed-source models~\citep{gemini2025flashimage,seedream2025seedream} have recently demonstrated promising capabilities in multi-turn reasoning and editing, such as maintaining visual memory and context coherence, most open-source models~\citep{huang2025diffusion,brooks2023instructpix2pix,yu2025anyedit} remain confined to single-turn editing.
This gap stems from the absence of high-quality interleaved datasets capturing the temporal dependencies and iterative workflows of real-world multi-turn editing.
Existing datasets~\citep{zhang2024magicbrush,wang2025gpt,ye2025echo,ye2025imgedit} are fundamentally single-turn, treating each edit as an isolated instruction and thus failing to represent the long-horizon reasoning required for authentic interactive image creation, as illustrated in Figure~\ref{fig-comparison}(a).
This lack has hindered systematic exploration, and benchmarks for evaluating multi-turn editing with historical context remain absent.

To address this gap, we present \colorname{}, the first suite for in-context interleaved cross-modality comprehension and generation.
Our suite consists of two complementary parts. 
\trainname{} is a large-scale dataset containing $100$K interleaved samples, spanning over $370$K dialogue turns and $500$K images across comprehension, editing, and generation tasks that require reasoning over historical context.
As shown in Figure~\ref{fig-comparison}, effective multi-turn editing tasks demand strong visual memory to retrieve and reuse objects, layouts, and styles from previous rounds, for instance by removing an item in one turn and accurately restoring it later.
This interleaved design captures the iterative nature of realistic multi-turn image editing, in which each modification can rely on information from prior rounds.

\testname{} is a human-annotated benchmark of $100$ tasks with $480$ images, featuring a hybrid VLM judge evaluation framework with four metrics that evaluate alignment with reference images and fidelity to original images and correctness for editing instructions. The benchmark assesses models' capabilities in multi-turn generation, visual memory, and world-knowledge reasoning across diverse domains, including science, creation, logic, and games. \testname{} reveals that current models struggle with in-context interleaved generation and exhibit performance degradation as content length increases, indicating substantial room for improvement.

Experiments demonstrate that training on \trainname{} yields substantial improvements in vision comprehension ($9.8$\% on MMMU), image editing ($4.8$\% on GEditBench), and comprehension-generation collaboration (approximately $50$\% on RISE). Moreover, training facilitates the emergence of visual memory capabilities in UMMs, while evaluations on \testname{} reveal persistent limitations in multi-turn, context-aware image generation.

To summarize, our contributions are threefold:
\begin{itemize}[leftmargin=*]
    \item We introduce \trainname{}, the first large-scale dataset for multi-turn, context-aware image understanding and generation, comprising over $100$K samples, $370$K dialogue turns, and $500$K images across comprehension, editing, and generation tasks.
    
    \item We present \testname{}, the first human-annotated benchmark for interleaved multimodal comprehension and generation, featuring $100$ carefully designed cases with $480$ images and a hybrid VLM judge evaluation framework that assesses multi-turn generation, visual memory, and world-knowledge reasoning.
    
    \item Through extensive experiments, we demonstrate that training on \trainname{} significantly improves performance on established benchmarks and facilitates the emergence of visual memory capabilities, while evaluation on \testname{} reveals persistent limitations in multi-turn, context-aware generation.
\end{itemize}

\section{Related Works}
\begin{table*}[!t]
    \centering
    \vspace{-0.4cm}
    \caption{\textbf{Summary of Multimodal Reasoning Benchmarks.} We compare existing works from aspects including: $^1$interleave, $^2$multi-turn, $ ^3$vision memory, $^4$multidimensional evaluation, $^5$hybrid evaluation, and $^6$whether manual annotations and filtering are applied. \faImage  \faMagic \faEdit means text to image, image edit and image comprehension.}
    \vspace*{-5pt}
    \label{table_benchmark_comparison}
    \resizebox{\textwidth}{!}{
    \begin{tabular}{rrcccccccr}
    \toprule
    \multirow{2}{*}{\textbf{Benchmark}} &  \multirow{2}{*}{\textbf{Venue}} & \multirow{2}{*}{\textbf{Inter.}} & \textbf{Multi-} & \textbf{Vision} & \textbf{Multi-} & \textbf{Hybrid} & \multirow{2}{*}{\textbf{\# Domain}}& \multirow{2}{*}{\textbf{\# Num}} & \multirow{2}{*}{\textbf{\#Types}}
    \\
    & & & \textbf{Turn} & \textbf{Mem.} & \textbf{Dim.} & \textbf{Eval} & 
    \\
    \midrule\midrule
    ReasonPix2Pix~\citep{jin2024reasonpix2pix} & \textcolor{gray}{{\small arXiv'24}} &\crossmark & \crossmark & \crossmark & \crossmark & \crossmark & \faEdit & $40,212$ & $1$  
    \\
    ReasonEdit~\citep{huang2024smartedit} &\textcolor{gray}{{\small CVPR'24}} & \crossmark & \crossmark & \crossmark & \crossmark & \crossmark& \faEdit & $219$ & $1$
    \\
    Reason50K~\citep{he2025reasoning} & \textcolor{gray}{{\small arXiv'25}} & \crossmark & \crossmark & \crossmark & \crossmark & \crossmark & \faEdit & $51,039$ & $4$
    \\
    Zebra-CoT~\citep{li2025zebra} & \textcolor{gray}{{\small arXiv'25}} & \checkmark & \checkmark & \crossmark & \crossmark & \crossmark & \faMagic \faEdit & $182,384$ & $4$ \\
    KRIS-Bench~\citep{wu2025kris} & \textcolor{gray}{{\small NeurIPS'25}} & \crossmark & \crossmark & \crossmark &\checkmark &\checkmark & \faMagic & $1,267$ & $7$
    \\
    RISEBench~\cite{zhao2025envisioning} & \textcolor{gray}{{\small NeurIPS'25}} &\crossmark & \crossmark & \crossmark &\checkmark &\checkmark & \faMagic & $360$ & $4$
    \\
    CoMM~\citep{chen2025comm} &\textcolor{gray}{{\small CVPR'25}} &\checkmark  &\crossmark & \crossmark & \crossmark & \crossmark & \faImage \faEdit & $227,000$ & $1$ 
    \\
    IRG-300k~\citep{huang2025interleaving} & \textcolor{gray}{{\small arXiv'25}} & \checkmark & \crossmark & \crossmark & \crossmark & \crossmark & \faImage & $1$ & $1$  
    \\
    Echo-4o~\citep{ye2025echo} & \textcolor{gray}{{\small arXiv'25}} & \crossmark & \crossmark & \crossmark & \crossmark & \crossmark & \faImage & $179,000$ & $3$ 
    \\
    ROVER~\citep{liang2025rover} & \textcolor{gray}{{\small arXiv'25}} & \checkmark & \crossmark & \crossmark & \checkmark & \checkmark & \faMagic \faEdit & $1,312$ & $23$ 
    \\
    \midrule
    \textbf{\colorname} & {\textbf{Ours}} & \checkmark & \checkmark & \checkmark & \checkmark & \checkmark & \faImage  \faMagic \faEdit & 100,100 & $16$ 
    \\ 
    \bottomrule         
\end{tabular}
}
\end{table*}

\noindent\textbf{Unified Multimodal Models} represent a paradigm designed to seamlessly integrate multimodal comprehension and generation capabilities within a single framework.
To achieve this unified objective, seminal works~\citep{karypis1999chameleon,wu2025janus,chen2025janus,pan2024auto} leverage image tokenization and autoregressive next-token prediction to generate visual tokens.
Subsequent developments, driven by the pursuit of enhanced image synthesis quality, incorporate diffusion-based or flow-matching heads~\citep{lipman2022flow} integrated with shared transformer architectures~\citep{deng2025bagel,ma2025janusflow,zhou2024transfusion,qin2025uni}.
Recent works have demonstrated remarkable potential for instruction-based image editing~\citep{liu2025step1x,song2025query,wang2025seededit} and multi-image composition~\citep{seedream2025seedream,xia2025dreamomni2}. However, the capabilities of UMMs for in-context interleaved comprehension and generation remain largely unexplored.

\noindent\textbf{Image Editing.}
Recent text-guided image editing has achieved substantial progress~\citep{bai2024humanedit, liu2025step1x, ye2025imgedit}. For instance, AnyEdit~\citep{yu2025anyedit} provides general-purpose editing datasets that unify diverse edit types, such as insertion, replacement, and style modification. GPT-Image-Edit-1.5M~\citep{wang2025gpt} and Echo-4o~\citep{ye2025echo} expand data scale and instruction diversity by leveraging GPT-4o. However, these approaches remain limited to one-shot edits without historical context or iterative refinement. While MagicBrush~\citep{zhang2024magicbrush} introduces multi-turn editing, each instruction is treated as an independent request without multi-turn dependencies. As shown in Table~\ref{table_benchmark_comparison}, \trainname{} introduces the first in-context interleaved cross-modal dataset that explicitly captures multi-turn editing and context-dependent generation, enabling models to learn visual memory and consistent reasoning.
More dicussion for the related works can be found in Appendix~\ref{app:related}.

\begin{figure*}[!t]
    \centering
    \includegraphics[width=0.9\linewidth]{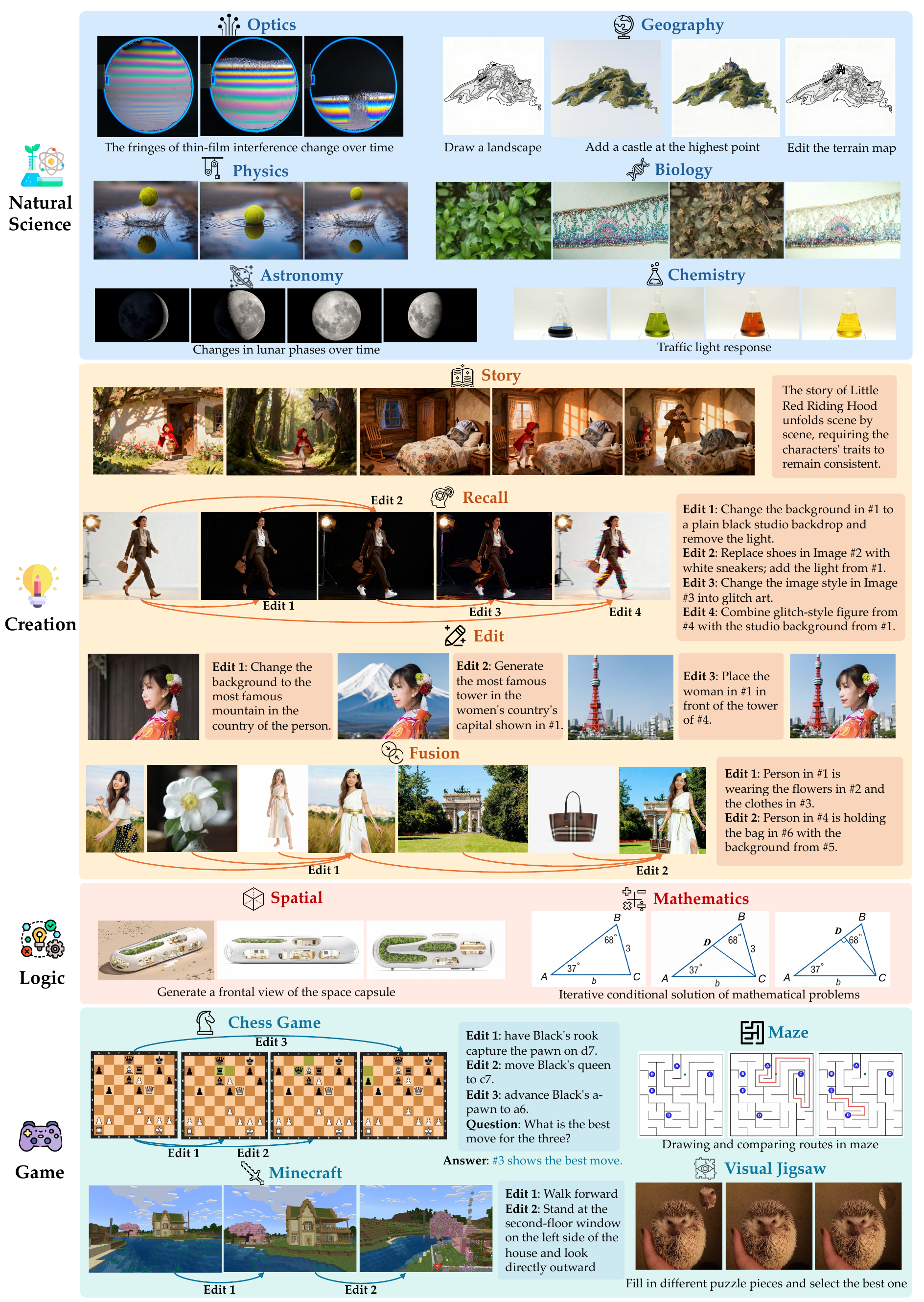}
    \vspace{-0.4cm}
    \caption{\textbf{Overview for \testname{}}. We have shown only a subset of the \colorname{}; further details are in the Appendix~\ref{app:case-test}.}
    \label{fig:test}
\end{figure*}
\begin{figure*}[!h]
    \centering
    \vspace{-0.4cm}
    \includegraphics[width=1.\linewidth]{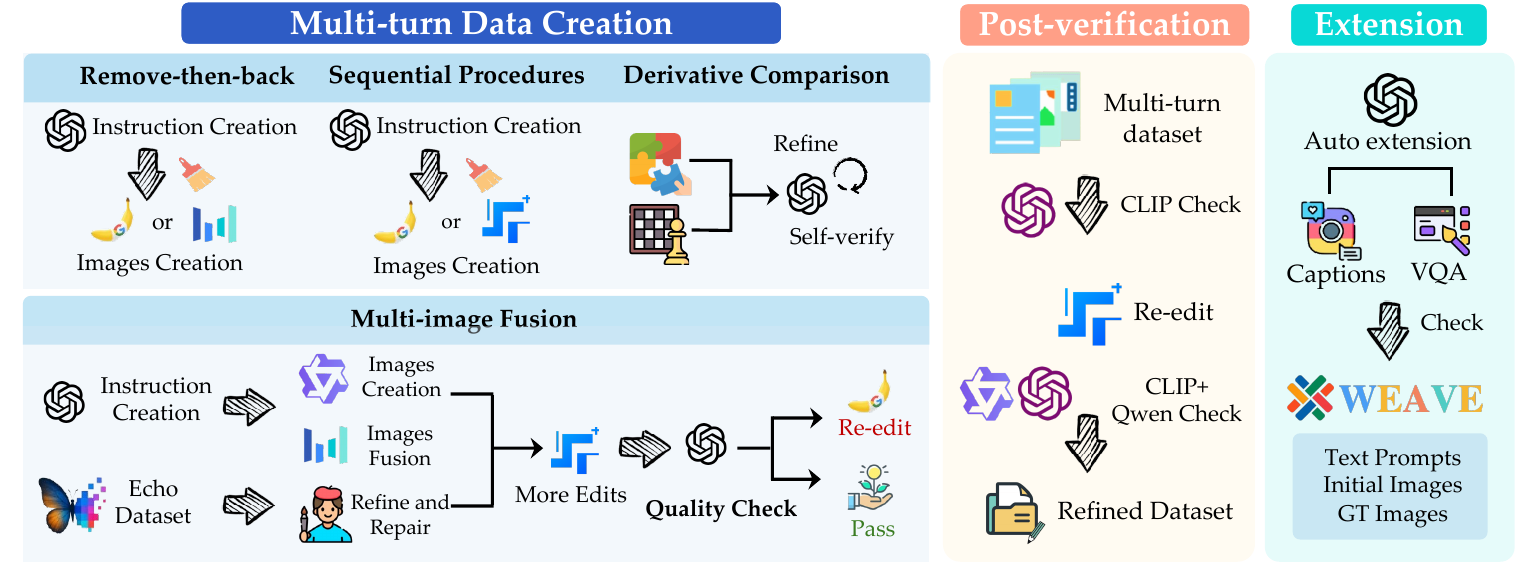}
    \caption{\textbf{Data Annotation Pipeline for \colorname{}.} Our methodology ensures data diversity and quality through a multi-round image generation process, supplemented by two rounds of validation and refinement. Additional details are provided in Appendix~\ref{app:data}.}\label{pipeline}
  \vspace{-4mm} 
\end{figure*}
\begin{figure*}[!h]
 \begin{minipage}{0.28\textwidth} 
    \centering
    \fontsize{6.2pt}{\baselineskip}\selectfont 
    \renewcommand\tabcolsep{0.9pt}
    \renewcommand\arraystretch{0.7}
    \scalebox{0.99}{
        \begin{tabular}{lr}
             \toprule
             \textbf{Statistic} & \textbf{Number} \\
             \midrule\midrule
              \cellcolor{weaveW!15}\textcolor{weaveW}{\textbf{$\bullet$~Total Chats}} & \cellcolor{weaveW!15}$100{,}750$ \\
              ~- $\ge4$ Images Chats& $100{,}584$ \\
              ~- $\ge5$ Images Chats& $60{,}361$ \\
              ~- $\ge6$ Images Chats& $31{,}571$ \\
             \midrule
             \cellcolor{weaveE!15}\textcolor{weaveE}{\textbf{$\bullet$~Average Chat Turns}} & \cellcolor{weaveE!15}$3.79$ \\
             ~-Average Question Length & $195.49$ \\
             \midrule
             \cellcolor{weaveA!15}\textcolor{weaveA}{\textbf{$\bullet$~Total Images}} & \cellcolor{weaveA!15}$505{,}186$ \\
             ~- Maximum Image Per Chats & $8$ \\
             ~- Average Image Per Chats & $5.01$ \\
             \bottomrule
             \end{tabular}
     }
     \caption{Statistics for \trainname{}.}
     \label{tab:bench_statistics}
 \end{minipage} 
 ~~\hfill~~
 \begin{minipage}{0.72\textwidth}
     \centering 
     \vspace{-1mm}
    \includegraphics[width=0.94\linewidth]{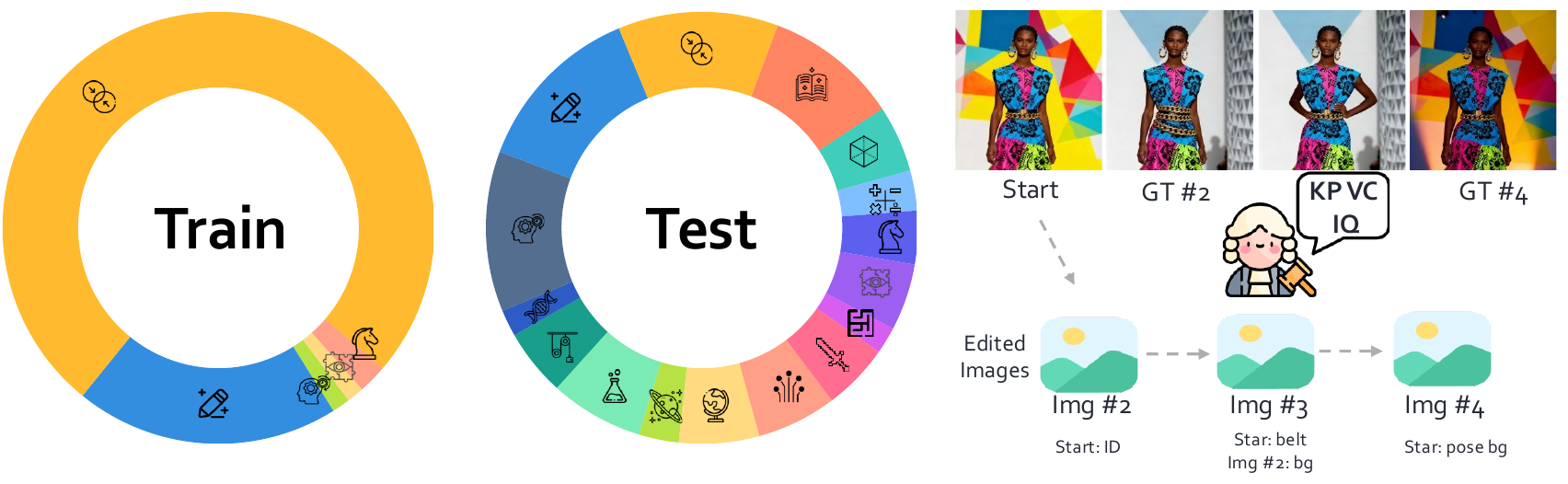}
    \vspace{-0.5mm}
     \caption{Summary of domain distributions and evaluation methods for \colorname{}.}
     \label{fig:main_stat}
 \end{minipage}

\vspace{-0.4cm}
\end{figure*}

\section{\colorname{}}
To assess in-context interleaved comprehension and generation, we first introduce the data collection pipelines \trainname{} and \testname{} in Section~\ref{sec:data-collect}. We then detail the evaluation settings and metrics in Section~\ref{sec:mertic}, and present key statistics for \colorname{} in Section~\ref{sec:stat}.

\subsection{Data Collection}\label{sec:data-collect}
\textbf{\trainname{}} In order to generate rich and diverse data with visual memory capabilities, we constructed a data pipeline as illustrated in Figure~\ref{pipeline}. This pipeline incorporates four distinct generation pathways followed by multiple filtering and refinement stages to ensure accuracy and quality of the produced data.
To generate multi-turn editing data with visual memory capabilities, we implemented four methodological approaches:
\textit{(i) Multi-image fusion}: We achieved reference to previous iterations by fusing edited or directly generated images.
\textit{(ii) Remove-then-back}: We employed a technique of first removing or replacing objects, then adding them back, enabling the system to recall previously deleted visual elements.
\textit{(iii) Derivative imagination and comparison}: We incorporated methods for deriving or imagining alternative solutions or new images before fusion.
\textit{(iv) Sequential procedures}: We implemented sequential edits following narrative progressions or structured editing operations.
Further details regarding the data collection methodology are presented in Appendix~\ref{app:data-collect}.

\noindent \textbf{\testname{}} is annotated by individuals with graduate-level STEM degrees. It comprises $100$ items across $16$ task categories, incorporating both multi-turn editing tasks requiring visual memory and challenges demanding world knowledge (cultural contexts, physical phenomena, and chemical processes). As illustrated in Figure~\ref{fig:test}, tasks included generating examples involving the Tokyo Tower and demonstrating comprehension of traffic signal reactions. The images used include web-sourced content and synthetically generated images from three models: Seedream 4.0~\citep{seedream2025seedream}, Nano Banana~\citep{comanici2025gemini} and SeedEdit 3.0~\citep{wang2025seededit}.

\subsection{Evaluation Settings and Metrics}\label{sec:mertic}
We adopt the VLM-as-judge~\citep{liu2025step1x} automated evaluation framework, with detailed templates provided in Appendix~\ref{app:prompt}. To enable focused assessment, we employ a key-point-based scoring approach using structured evaluation criteria. Specifically, we leverage a hybrid strategy that instructs the VLM to evaluate based on both the reference image and the combination of the original image with editing instructions. As shown in Figure~\ref{fig:main_stat}, the judge invokes different images as references and assigns scores according to predefined key points.

\noindent Our evaluation comprises $4$ metrics (the first three apply to editing tasks; the last applies to comprehension tasks):
\textbf{Key Point Correctness (KP):} Measures whether the edited image satisfies the specified editing requirements.
\textbf{Visual Consistency (VC):} Ensures non-target elements remain unchanged, maintains consistency with the original image (unedited regions remain intact when the scene is preserved; edited regions maintain stylistic coherence when the scene is modified), and assesses identity preservation of edited objects.
\textbf{Image Quality (IQ):} Evaluates the overall quality of the generated image.
\textbf{Accuracy (Acc):} Measures the correctness of the reasoning result. 
Details regarding score calculation methodology can be found in Appendix~\ref{app:metric}.

\subsection{Data Statics}\label{sec:stat}
For each instance in \colorname{}, we provide a text prompt, one or more initial images, and ground-truth examples. The test set additionally includes key information that the correct output images must satisfy. 

\noindent Representative dataset examples are provided in Appendix~\ref{app:data-example}.
Table~\ref{tab:bench_statistics} presents key statistics of the training set. The majority of instances contain more than five images, with an average of 3.8 conversational turns per instance. Figure~\ref{fig:main_stat} illustrates the category distribution across both training and test sets, demonstrating a relatively balanced distribution across data types.

\section{Experiment}
\begin{table*}[!t]
    \vspace{-6mm}
    \centering
    \scalebox{0.98}{
    \begin{tabular}{lccccccccc} 
    \toprule
    & Size & In-context & Modality & Format & \faFlask Science &  \faPaintBrush Creation & \faPuzzlePiece Logic & \faGamepad Game & \textbf{Avg}\\
    \midrule
    \midrule
    Intern3.5-VL~\citep{wang2025internvl3} & 8B & \faCheck & \faFileTextO& \faLongArrowRight & 0.114 & \textcolor{weaveW}{0.500} & \textcolor{weaveW}{0.667} & \textcolor{weaveW}{0.292} &0.231\\
    Qwen3-VL~\citep{Qwen2.5-VL} &  8B & \faCheck & \faFileTextO& \faLongArrowRight & 0.432 & 0.000 & 0.000 & 0.250 & 0.298\\
    GPT-4o~\citep{achiam2023gpt} & - & \faCheck & \faFileTextO& \faLongArrowRight & 0.591 & \textcolor{weaveW}{0.500} & 0.167 & 0.083 &0.381\\ 
    GPT-4.1~\citep{achiam2023gpt} & - & \faCheck & \faFileTextO& \faLongArrowRight &\textcolor{weaveW}{0.705} &\textcolor{weaveW}{0.500} &0.167 &0.167 & \textcolor{weaveW}{0.464}\\
    \midrule
    AnyEdit~\citep{yu2025anyedit} & 1B & \faAdjust & \faImage & \faObjectGroup & 0.445 & 0.514 & 0.351 & 0.419 &0.472 \\
    UltraEdit(SD3)~\citep{zhao2024ultraedit} & 2B & \faAdjust &  \faImage & \faObjectGroup & 0.493 & 0.561 & 0.491 &0.440 &0.522\\
    VAREdit-8B~\citep{varedit2025} & 8B & \faAdjust & \faImage &\faObjectGroup & 0.536 &0.636 & 0.584 & 0.580&0.603 \\
    Step1X-Edit v1.1~\citep{liu2025step1x-edit} &  12B& \faAdjust & \faImage &\faObjectGroup& 0.574 & 0.714 & 0.700 & 0.625 & 0.669\\
    Step1X-Edit v1.2~\citep{liu2025step1x-edit} & 12B& \faAdjust & \faImage &\faObjectGroup& 0.560 & 0.644 & 0.530 &0.562 & 0.605\\
    FLUX.1 Kontext~\citep{labs2025flux1kontextflowmatching} &12B& \faAdjust & \faImage &\faObjectGroup& 0.589 & 0.756 &0.639 &0.610 & 0.689\\
    Qwen-Image-Edit~\citep{wu2025qwenimagetechnicalreport} & 20B & \faAdjust & \faImage &\faObjectGroup & 0.586 & 0.715 & 0.589 & 0.628 & 0.665\\
    \midrule
    OminiGen~\citep{xiao2025omnigen} & 4B & \faAdjust & \faImage &\faLongArrowRight & 0.398 &0.474 & 0.401 & 0.177 & 0.404 \\
    OminiGen2~\citep{wu2025omnigen2} & 7B & \faAdjust & \faAsterisk & \faLongArrowRight &0.511&0.682&0.551&0.511&0.609 \\
    Ovis-U1~\citep{wang2025ovis} & 3B & \faAdjust & \faAsterisk &\faObjectGroup & 0.402 &0.557 & 0.364 &0.357 &0.422\\
    UniPic~\citep{wang2025skywork} & 1.5B & \faAdjust & \faImage &\faObjectGroup& 0.472 & 0.590 & 0.463 & 0.316 & 0.511\\
    UniPic2-SD3.5M~\citep{wei2025skywork} & 2B & \faAdjust & \faImage &\faObjectGroup& 0.477 & 0.625 & 0.543 & 0.497 & 0.568\\
    UniPic2-Metaquery~\citep{wei2025skywork} & 9B & \faAdjust & \faImage &\faObjectGroup& 0.493 & 0.666 & 0.507 & 0.444 &0.582 \\
    NextStep-1-Large~\citep{team2025nextstep} & 15B & \faAdjust & \faImage &\faObjectGroup  & 0.519 &0.620 & 0.437 & 0.309 & 0.534 \\
    Seedream 4.0~\citep{seedream2025seedream} & - & \faAdjust & \faImage &\faLongArrowRight & 0.683 & \textcolor{weaveA}{0.847} & 0.679 & 0.635 & 0.765 \\
    Seedream 4.0~\citep{seedream2025seedream} & - & \faCheck & \faImage &\faLongArrowRight & 0.667 & 0.830 & 0.646 & 0.599 & 0.746 \\
    Nano Banana~\citep{comanici2025gemini} & - & \faAdjust & \faImage &\faLongArrowRight & 0.715 & 0.823& 0.666 &\textcolor{weaveA}{0.666}  & 0.764 \\
    Nano Banana~\citep{comanici2025gemini} & - & \faCheck & \faImage & \faLongArrowRight & \textcolor{weaveA}{0.710} & 0.843 & \textcolor{weaveA}{0.730} & 0.613 & \textcolor{weaveA}{0.767}\\
    Bagel~\citep{deng2025bagel} &14B& \faAdjust & \faAsterisk & \faLongArrowRight & 0.378 & 0.475 &  0.406 &0.365 &0.446\\
    Bagel‑Zebra~\citep{li2025zebra} &14B& \faAdjust & \faAsterisk & \faLongArrowRight & 0.399 & 0.456 & 0.393 &0.396 & 0.449\\
    \rowcolor{my_red!7} 
    \textbf{+ \trainname{}} &14B & \faAdjust & \faAsterisk & \faLongArrowRight & \textcolor{weaveV}{0.537} & \textcolor{weaveV}{0.706} & \textcolor{weaveV}{0.567} & \textcolor{weaveV}{0.531} & \textcolor{weaveV}{0.640} \\
    \bottomrule
    \end{tabular}
    }
    \vspace{-2mm}
    \caption{\textbf{Main results on \testname{}.} The symbols \faAdjust\ and \faAdjust*\ denote full and partial in-context, respectively. Icons \faImage, \faFileTextO, and \faAsterisk\ indicate image-only, text-only, and combined evaluations, respectively. \faLongArrowRight\ and \faObjectGroup\ represent sequential and concatenated image inputs, respectively. We use \textcolor{weaveW}{blue}, \textcolor{weaveA}{orange}, and \textcolor{weaveV}{green} to represent the optimal results across three modalities.}\label{main_exp}
    \vspace{-2mm}
\end{table*}

We first evaluate $22$ models on \testname{} in Section~\ref{sec:benchmark}, revealing that current models struggle with in-context interleaved generation and exhibit performance degradation as content length increases. Subsequently, in Section~\ref{sec:bagel}, we validate the high quality of \trainname{} through fine-tuning Bagel. Finally, we conduct quality analysis and assess judge effectiveness in Section~\ref{sec:quality} and~\ref{sec:judge}.

\subsection{\testname{}}\label{sec:benchmark}
\noindent\textbf{Settings.} 
We evaluated $4$ LLMs, $7$ Edit models, and $11$ UMMs on \testname{} as presented in Table~\ref{main_exp}. Evaluations were conducted under three distinct in-context conditions: \textit{(1) no in-context} (single-turn generation without contextual information), \textit{(2) partial in-context} (using only self-generated images with explicitly mentioned visual context, excluding other historical interactions), and \textit{(3) complete in-context} (with all previous interactions visible). 
For image placement, we employed two configurations: "yes-first," where images appear at their first mention position, and "yes-front," where all images are consolidated at the beginning of the input (results for this configuration are reported in Table~\ref{main_exp}). For models incapable of processing sequence-format inputs, we implemented a concatenation approach following methodologies established in prior work \citep{zhang2024task,chow2025physbench}.
Based on the results presented in the table, we can derive the following conclusions: 

\noindent\textbf{In-context image generation remains challenging.} Among the models tested, the best-performing Edit and UMM approaches achieved maximum scores of only $0.68$ and $0.767$, respectively. Furthermore, significant domain biases were observed, with performance in creative imagery consistently surpassing that in scientific and logical domains. This suggests substantial room for improvement in generation ability to effectively integrate world knowledge.

\begin{figure*}[!ht]
\vspace{-2mm}
    \centering
    \includegraphics[width=1.\linewidth]{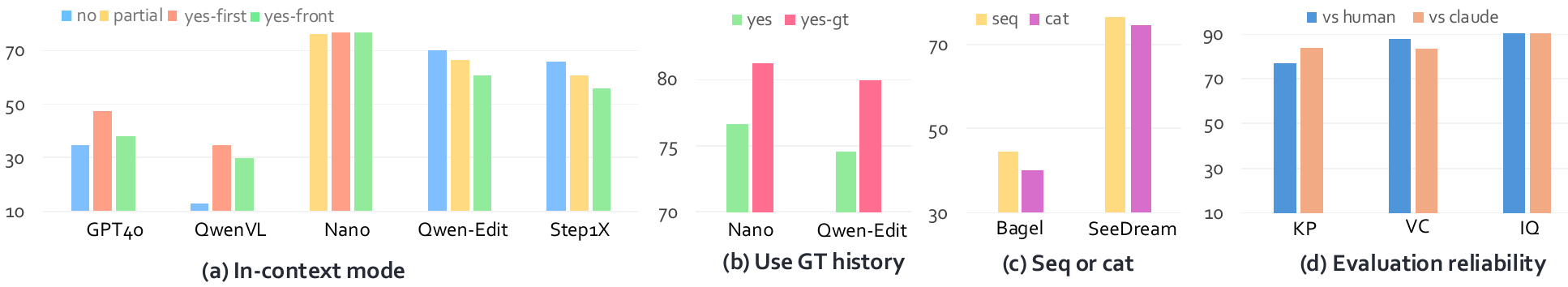}
   \caption{\textbf{(a)} Impact of different in-context modes on performance. \textbf{(b)} Reasoning performance using ground truth as in-context examples. \textbf{(c)} Performance variations when concatenating sequential images. \textbf{(d)} Evaluation reliability of GPT4.1 judger.}
  \vspace{-4mm}  
  \label{tab:aba}
\vspace{0.5cm}
    \centering
    \includegraphics[width=1.\linewidth]{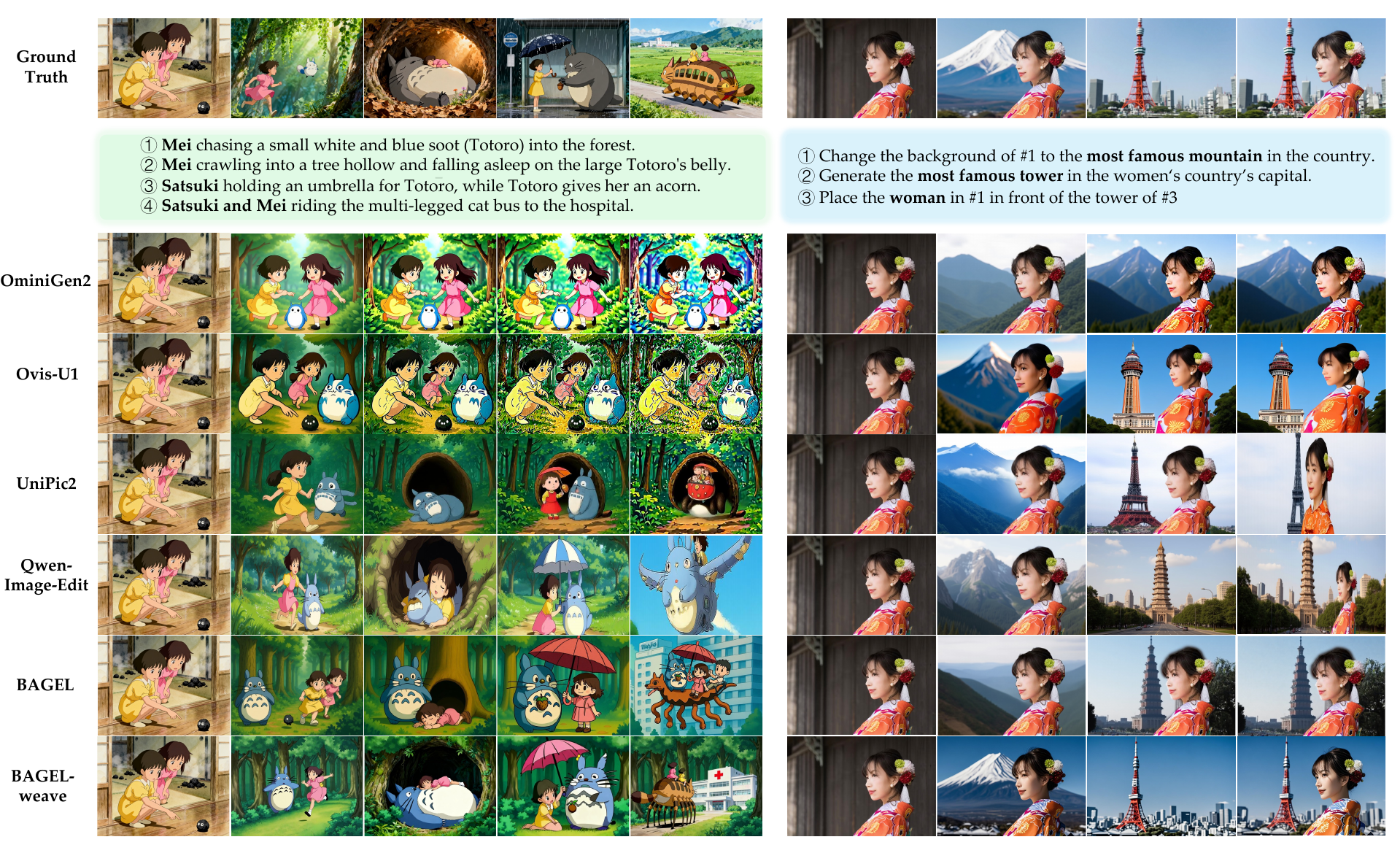}
    \vspace{-4mm} 
    \caption{\textbf{Qualitative comparison between different methods.} The left-side task requires preservation of character IDs, while the right-side task necessitates the application of world knowledge and maintenance of character removal followed by reinsertion.}
  \vspace{-5mm}  
  \label{fig:quality}
\end{figure*}

\noindent\textbf{In-context usage matters} 
\textit{(a) For comprehension tasks}, we observed significant performance improvements when utilizing in-context information compared to baseline conditions without historical context. This effect was particularly pronounced in QwenVL, which demonstrated a remarkable $163\%$ improvement as illustrated in Figure~\ref{tab:aba}(a), indicating that \testname{} successfully incorporated historical information into the model evaluation.
\textit{(b) For generation tasks}, increasing in-context content produced divergent effects across model types. Open-source models exhibited progressive performance degradation with additional historical context—Qwen-Edit showed decremental performance of $5.3\%$ and $8.6\%$ respectively. This suggests that open-source models, constrained by single-round editing capabilities, experience diminished localization accuracy when processing expanded contextual information, thereby failing to effectively utilize in-context data. Conversely, proprietary models such as Nano demonstrated incremental improvement, indicating successful utilization of contextual information.
\textit{(c) \testname{} exhibits superior image quality.} As illustrated in Figure~\ref{tab:aba}(b), incorporating \testname{}'s ground truth images as in-context examples resulted in performance improvements across all models. Notably, Qwen-Image-Edit demonstrated a significant improvement of $7.1$\%, potentially attributable to Qwen-Image-Edit's inherently weaker generative capabilities compared to the nano-banana~\citep{gemini2025flashimage}.

\noindent\textbf{Sequential Input Superiority.} As illustrated in Figure~\ref{tab:aba}(c), sequential image input demonstrates significant performance advantages over concatenated input. This effect is particularly pronounced with the Bagel model, where concatenation results in a $10.3\%$ performance degradation. These findings highlight the potential of UMMs as effective editing models, especially considering that traditional editing models cannot directly process multiple images and historical information as input.

\begin{table*}[!ht]
    \centering
    \vspace{-0.1cm}
    \resizebox{0.98\textwidth}{!}{%
    \begin{tabular}{l|ccccccccccc|c}
        \toprule
        \textbf{Model} &
        \textbf{BG} &
        \textbf{Color} &
        \textbf{Mat.} &
        \textbf{Motion} & \textbf{Port.} & \textbf{Style} &
        \textbf{Add} & \textbf{Remove} & \textbf{Replace} &
        \textbf{Text} & \textbf{Tone} & \textbf{Avg} 
        \\
        \midrule
        \midrule
        AnyEdit~\citep{yu2025anyedit} 
        & 4.31 & 4.25 & 2.64 & 0.67 & 1.90 & 1.95 & 3.72 & 3.75 & 3.23 & 0.77 & 4.21 & 2.85 
        \\
        MagicBrush~\citep{zhang2024magicbrush} 
        & 6.17 & 5.41 & 4.75 & 1.55 & 2.90 & 4.10 & 5.53 & 4.13 & 5.10 & 1.33 & 5.07 & 4.19 
        \\
        InstructPix2Pix~\citep{brooks2023instructpix2pix} 
        & 3.94 & 5.40 & 3.52 & 1.27 & 2.62 & 4.39 & 3.07 & 1.50 & 3.48 & 1.13 & 5.10 & 3.22 
        \\
        OmniGen~\citep{xiao2025omnigen} 
        & 5.23 & 5.93 & 5.44 & 3.12 & 3.17 & 4.88 & 6.33 & 6.35 & 5.34 & 4.31 & 4.96 & 5.01 
        \\
        Step1X-Edit~\citep{liu2025step1x-edit} 
        & 7.03 & 6.26 & 6.46 & 3.66 & 5.23 & \textbf{7.24} & 7.17 & 6.42 & \underline{7.39} & \textbf{7.40} & 6.62 & 6.44 \\
        OminiGen2~\citep{wu2025omnigen2} 
        & 6.99 & 6.66 & 4.88 & 2.55 & 3.66 & 6.08& 7.09 &6.60 & 6.65 & 4.49 & 6.03 & 5.57 
        \\
        UltraEdit (SD3)~\citep{zhao2024ultraedit} 
        & 5.83 &5.51 &5.86&3.55&\underline{5.00}&5.73&5.06&3.15&5.79&2.24&5.45&4.83
        \\
        EditMGT~\citep{anonymous2025editmgt} & \textbf{7.69}  &\textbf{7.71}  & 5.77  &3.84  & \textbf{5.13}  &6.53  &6.13  &5.24  &5.56  & 4.53 & 6.42  & 5.87
        \\
        GoT-6B~\citep{fang2025got} 
        & 4.11& 5.75&3.04&1.71&2.69&4.72&5.77 &4.59&5.65&1.16&4.24&3.95 
        \\
        VAREdit-8B~\citep{varedit2025} 
        & 6.77&6.64&5.40&3.33&4.20&6.46&5.86 & 7.29& 6.67&3.87&6.54&5.73
        \\
        FluxKontext.dev~\citep{labs2025flux1kontextflowmatching} 
        & 7.06 & \underline{7.03} & 5.52 & \textbf{5.62} & 4.68 & 5.55 & 6.95 & 6.76 & 6.13 & 6.10 & \textbf{7.48} & 6.26 
        \\
        \midrule
        Bagel~\citep{deng2025bagel} 
        & 7.44 & 6.99 & \underline{6.26} & \underline{5.09} & 4.82 & 6.04 & \textbf{7.94} & \underline{7.37} & 7.31 & \underline{7.16} & 6.17 & \underline{6.52} \\
        \rowcolor{my_red!7} 
        \textbf{+ \trainname{}} & \underline{7.45} & 7.00 & \textbf{7.10} & 4.97 & 4.83 & \underline{6.98} & \underline{7.88} & \textbf{7.39} & \textbf{7.75} & 7.06 & \underline{6.81} & \textbf{6.83}
        \\
        \bottomrule
    \end{tabular}%
    }
    \caption{Comparison of fine-tuned Bagel and other models on GEdit-EN-full benchmark~\citep{liu2025step1x-edit}.}
    \label{tab:gedit}
    
    \vspace{0.2cm}  

    \centering
    \setlength{\tabcolsep}{4pt}
    \renewcommand{\arraystretch}{1.2}
    \scriptsize
    \begin{tabular}{l|ccc|ccccccc|cccc}
        \toprule
        & \multicolumn{3}{c}{\textbf{Understanding}} & \multicolumn{7}{c}{\textbf{GenEval}} & \multicolumn{4}{c}{\textbf{RISEBench}} \\
        \cmidrule(lr){2-4} \cmidrule(lr){5-11} \cmidrule(lr){12-15}
        \textbf{Model} & \textbf{MMB} & \textbf{MMMU} & \textbf{MMVet} & \textbf{Single Obj.} & \textbf{Two Obj.} & \textbf{Count.} & \textbf{Color} & \textbf{Position} & \textbf{Attri.} & \textbf{Overall} & \textbf{Tem.} & \textbf{Cau.} & \textbf{Spa.} & \textbf{Log.} \\
        \midrule
        Emu$3$\citep{wang2024emu3} &  58.5 &  31.6  & 37.2 & 0.99 & 0.81 & 0.42 & 0.80 & 0.49 & 0.45 & 0.66 & - & - & - & - \\
        Show-o~\citep{xie2025show} & - & 27.4 & - &  0.98 & 0.80 & 0.66 & 0.84 & 0.31 & 0.50 & 0.68 & - & - & - & - \\
        Janus-Pro-7B~\cite{chen2025janus} & 75.5& 36.3& 39.8 &  0.99 & 0.89 & 0.59 & 0.90 & 0.79 & 0.66 & 0.80 & - & - & - & - \\
        MetaQuery-XL~\cite{pan2025transfer} & 83.5 &\underline{58.6} &66.6 & - &  -& - & - & -& -& 0.80 &- & - & -  & - \\
        Ovis-U1~\citep{wang2025ovis} &77.8 & 51.1 & 66.7& \textbf{0.98}&  0.98&  \textbf{0.90} & \textbf{0.92} & 0.79&  \underline{0.75} &  \textbf{0.89}&  1.2	&3.3&4.0&\textbf{2.4}\\
        BLIP3-o~\citep{chen2025blip3} &83.5 & \underline{58.6} & 66.6 &   \textbf{1.00} &0.92& 0.63 &\underline{0.91}& 0.86& 0.67& 0.83&- & - & - &- \\
        EMU2~\citep{sun2024generative} & - &34.1 & 48.5 & - &- &- &- &- &- &- &1.2 & 1.1 & 0.0 & 0.0 \\
        OmniGen~\citep{xiao2025omnigen} & & & & 0.99 &\textbf{0.86}  &0.64  &0.85  &0.31  &0.55 &0.70 &1.2 & 1.0 & 0.0 & \underline{1.2} \\
        OmniGen2~\citep{wu2025omnigen2} & 79.1 &53.1& 61.8  & \textbf{1.00}& \underline{0.95} & 0.64& 0.88& 0.55 &\textbf{0.76} &0.80& - & - & - &- \\
        BAGEL~\citep{deng2025bagel} & \underline{85.0} &55.3& \underline{67.2} & 0.99 & 0.94  & 0.81 & 0.88 & 0.64 &0.63 & 0.82 & \underline{2.4} & \underline{5.6} & \underline{14.0} &\underline{1.2}\\
        \rowcolor{my_red!7} 
        \textbf{+ \trainname{}} &\textbf{85.2} & \textbf{60.7} & \textbf{67.4} & \textbf{1.00} & 0.94 & \underline{0.83} & 0.89 & \underline{0.65} & 0.70 & \underline{0.84} &\textbf{4.7} &\textbf{6.7} & \textbf{21.0}&\textbf{2.4} \\
    \bottomrule
    \end{tabular}
    \vspace{-0.1cm}
    \caption{Comparison of different models on understanding tasks (MMB, MMMU, MMVet), GenEval and RISEBench.}
    \label{tab:geneval}
\end{table*}

\subsection{Train on \trainname{}}\label{sec:bagel}
To demonstrate the effectiveness of our data, we conduct experiments on Bagel~\citep{deng2025bagel}, with detailed training specifications provided in Appendix~\ref{app:train}. Our approach improved performance across four task categories:

\noindent \textit{(i) Vision Comprehension.} Our data effectively enhanced performance on understanding tasks, particularly yielding a $9.8$\% improvement on MMMU~\citep{yue2024mmmu}.
\textit{(ii) Image Editing.} As shown in Table~\ref{tab:gedit}, the fine-tuned Bagel demonstrates a 4.8\% improvement in overall score on GEditBench~\citep{liu2025step1x}. Furthermore, the model surpasses its baseline counterpart in the majority of tasks, with particularly notable enhancements in material change and style change categories, showing improvements of 13.4\% and 15.6\%, respectively.
\textit{(iii) Comprehension and Generation Collaboration.} As evidenced in Table~\ref{tab:geneval}, the fine-tuned Bagel demonstrates significant improvements across RISE cognitive tasks. Particularly noteworthy are the 100\% performance increases in both Spatial and Logical reasoning tasks. These results suggest that the fine-tuned Bagel more effectively leverages comprehension capabilities and world knowledge to enhance generation processes. Furthermore, these findings substantiate the high quality of the \trainname{} methodology.
\textit{(iv) Interleaved Cross-modality Comprehension and Generation.} As shown in Table~\ref{main_exp}, our fine-tuned model demonstrated a 42.5\% improvement over Bagel on \testname{}. Notably, there was a 34.6\% performance enhancement on more challenging science questions, indicating that training with our dataset significantly improved the model's interleaved cross-modality comprehension and generation capabilities.

\subsection{Quality Analysis.}\label{sec:quality}
As illustrated in Figure~\ref{fig:quality}, our analysis of quality results yields the following conclusions: 
\textit{(i) Instruction-following capabilities still require further improvement.} For instance, in the case on the left side of the figure, OmniGen and Ovis failed to execute the generation correctly. Similarly, in the case on the right side, the third column shows that Qwen-Image-Edit only generated a tower without including any human figures.
\textit{(ii) Fine-tuning on the weave dataset resulted in the emergence of visual memory capabilities.} The fine-tuned model correctly differentiates between protagonists wearing pink and yellow clothing in the left case, and in the right case, demonstrates the ability to first remove human figures and subsequently reintegrate them.

\subsection{Reliability of Judge Usage}\label{sec:judge}
To assess the reliability of VLM-as-a-judge scores, we conducted an expert evaluation study involving three human specialists across Nano-banana, Qwen-Image-Edit, and SeeDream models, analyzing $100$ instances per model. We computed Pearson correlation coefficients \citep{benesty2009pearson} between GPT-4.1 scores and expert ratings, with a comparative analysis against Claude Opus 4.1 evaluations (Figure~\ref{tab:aba}). Results demonstrate that correlations between GPT-4.1 and human ratings consistently exceed $0.8$, while Claude evaluations exhibit strong cross-VLM consistency, suggesting that the specific choice of VLM evaluator has minimal impact on assessment outcomes.

\section{Conclusion}
This work presents \colorname{}, the first comprehensive suite for in-context interleaved cross-modality comprehension and generation. We introduce \trainname{}, a large-scale dataset comprising $100$K samples that encompass $370$K dialogue turns and $500$K images, alongside \testname{}, a human-annotated benchmark consisting of 100 tasks with $480$ images and featuring a hybrid VLM judge evaluation framework. Our experiments demonstrate that training on \trainname{} yields substantial improvements across established benchmarks, including $9.8$\% gains on MMMU and $4.8$\% on GEditBench, while facilitating the emergence of visual memory capabilities in UMMs. At mean while, extensive evaluations on \testname{} reveal that current models still struggle with multi-turn, context-aware generation, particularly as content length increases. Moreover, this challenging task proves beyond the capabilities of conventional editing models. \colorname{} establish a foundation and  underscore the critical need for in-context interleaved multimodal comprehension and generation.

\cleardoublepage
{
    \small
    \bibliographystyle{ieeenat_fullname}
    \bibliography{main}

\begin{thebibliography}{95}
\providecommand{\natexlab}[1]{#1}
\providecommand{\url}[1]{\texttt{#1}}
\expandafter\ifx\csname urlstyle\endcsname\relax
  \providecommand{\doi}[1]{doi: #1}\else
  \providecommand{\doi}{doi: \begingroup \urlstyle{rm}\Url}\fi

\bibitem[Achiam et~al.(2023)Achiam, Adler, Agarwal, Ahmad, Akkaya, Aleman, Almeida, Altenschmidt, Altman, Anadkat, et~al.]{achiam2023gpt}
Josh Achiam, Steven Adler, Sandhini Agarwal, Lama Ahmad, Ilge Akkaya, Florencia~Leoni Aleman, Diogo Almeida, Janko Altenschmidt, Sam Altman, Shyamal Anadkat, et~al.
\newblock Gpt-4 technical report.
\newblock \emph{arXiv preprint arXiv:2303.08774}, 2023.

\bibitem[Alayrac et~al.(2022)Alayrac, Donahue, Luc, Miech, Barr, Hasson, Lenc, Mensch, Millican, Reynolds, et~al.]{alayrac2022flamingo}
Jean-Baptiste Alayrac, Jeff Donahue, Pauline Luc, Antoine Miech, Iain Barr, Yana Hasson, Karel Lenc, Arthur Mensch, Katherine Millican, Malcolm Reynolds, et~al.
\newblock Flamingo: a visual language model for few-shot learning.
\newblock \emph{Advances in neural information processing systems}, 35:\penalty0 23716--23736, 2022.

\bibitem[Anonymous(2025)]{anonymous2025editmgt}
Anonymous.
\newblock Edit{MGT}: Unleashing potentials of masked generative transformers in image editing.
\newblock In \emph{Submitted to The Fourteenth International Conference on Learning Representations}, 2025.
\newblock under review.

\bibitem[Antol et~al.(2015)Antol, Agrawal, Lu, Mitchell, Batra, Zitnick, and Parikh]{antol2015vqa}
Stanislaw Antol, Aishwarya Agrawal, Jiasen Lu, Margaret Mitchell, Dhruv Batra, C~Lawrence Zitnick, and Devi Parikh.
\newblock Vqa: Visual question answering.
\newblock In \emph{Proceedings of the IEEE international conference on computer vision}, pages 2425--2433, 2015.

\bibitem[Bai et~al.(2024{\natexlab{a}})Bai, Chow, Yang, Li, Li, Zhang, and Yan]{bai2024humanedit}
Jinbin Bai, Wei Chow, Ling Yang, Xiangtai Li, Juncheng Li, Hanwang Zhang, and Shuicheng Yan.
\newblock Humanedit: A high-quality human-rewarded dataset for instruction-based image editing.
\newblock \emph{arXiv preprint arXiv:2412.04280}, 2024{\natexlab{a}}.

\bibitem[Bai et~al.(2024{\natexlab{b}})Bai, Ye, Chow, Song, Chen, Li, Dong, Zhu, and Yan]{bai2024meissonic}
Jinbin Bai, Tian Ye, Wei Chow, Enxin Song, Qing-Guo Chen, Xiangtai Li, Zhen Dong, Lei Zhu, and Shuicheng Yan.
\newblock Meissonic: Revitalizing masked generative transformers for efficient high-resolution text-to-image synthesis.
\newblock In \emph{The Thirteenth International Conference on Learning Representations}, 2024{\natexlab{b}}.

\bibitem[Bai et~al.(2025)Bai, Chen, Liu, Wang, Ge, Song, Dang, Wang, Wang, Tang, Zhong, Zhu, Yang, Li, Wan, Wang, Ding, Fu, Xu, Ye, Zhang, Xie, Cheng, Zhang, Yang, Xu, and Lin]{Qwen2.5-VL}
Shuai Bai, Keqin Chen, Xuejing Liu, Jialin Wang, Wenbin Ge, Sibo Song, Kai Dang, Peng Wang, Shijie Wang, Jun Tang, Humen Zhong, Yuanzhi Zhu, Mingkun Yang, Zhaohai Li, Jianqiang Wan, Pengfei Wang, Wei Ding, Zheren Fu, Yiheng Xu, Jiabo Ye, Xi Zhang, Tianbao Xie, Zesen Cheng, Hang Zhang, Zhibo Yang, Haiyang Xu, and Junyang Lin.
\newblock Qwen2.5-vl technical report.
\newblock \emph{arXiv preprint arXiv:2502.13923}, 2025.

\bibitem[Benesty et~al.(2009)Benesty, Chen, Huang, and Cohen]{benesty2009pearson}
Jacob Benesty, Jingdong Chen, Yiteng Huang, and Israel Cohen.
\newblock Pearson correlation coefficient.
\newblock In \emph{Noise reduction in speech processing}, pages 1--4. Springer, 2009.

\bibitem[Brooks et~al.(2023)Brooks, Holynski, and Efros]{brooks2023instructpix2pix}
Tim Brooks, Aleksander Holynski, and Alexei~A Efros.
\newblock Instructpix2pix: Learning to follow image editing instructions.
\newblock In \emph{Proceedings of the IEEE/CVF conference on computer vision and pattern recognition}, pages 18392--18402, 2023.

\bibitem[Chaoyou et~al.(2023)Chaoyou, Peixian, Yunhang, Yulei, Mengdan, Xu, Jinrui, Xiawu, Ke, Xing, et~al.]{chaoyou2023mme}
Fu Chaoyou, Chen Peixian, Shen Yunhang, Qin Yulei, Zhang Mengdan, Lin Xu, Yang Jinrui, Zheng Xiawu, Li Ke, Sun Xing, et~al.
\newblock Mme: A comprehensive evaluation benchmark for multimodal large language models.
\newblock \emph{arXiv preprint arXiv:2306.13394}, 3, 2023.

\bibitem[Chen et~al.(2025{\natexlab{a}})Chen, Xu, Pan, Hu, Qin, Goldstein, Huang, Zhou, Xie, Savarese, et~al.]{chen2025blip3}
Jiuhai Chen, Zhiyang Xu, Xichen Pan, Yushi Hu, Can Qin, Tom Goldstein, Lifu Huang, Tianyi Zhou, Saining Xie, Silvio Savarese, et~al.
\newblock Blip3-o: A family of fully open unified multimodal models-architecture, training and dataset.
\newblock \emph{arXiv preprint arXiv:2505.09568}, 2025{\natexlab{a}}.

\bibitem[Chen et~al.(2019)Chen, Wang, Pang, Cao, Xiong, Li, Sun, Feng, Liu, Xu, et~al.]{chen2019mmdetection}
Kai Chen, Jiaqi Wang, Jiangmiao Pang, Yuhang Cao, Yu Xiong, Xiaoxiao Li, Shuyang Sun, Wansen Feng, Ziwei Liu, Jiarui Xu, et~al.
\newblock Mmdetection: Open mmlab detection toolbox and benchmark.
\newblock \emph{arXiv preprint arXiv:1906.07155}, 2019.

\bibitem[Chen et~al.(2025{\natexlab{b}})Chen, Bai, Zhao, Ye, Shi, Zhou, Chai, Lin, Wu, Tang, et~al.]{chen2025empirical}
Sixiang Chen, Jinbin Bai, Zhuoran Zhao, Tian Ye, Qingyu Shi, Donghao Zhou, Wenhao Chai, Xin Lin, Jianzong Wu, Chao Tang, et~al.
\newblock An empirical study of gpt-4o image generation capabilities.
\newblock \emph{arXiv preprint arXiv:2504.05979}, 2025{\natexlab{b}}.

\bibitem[Chen et~al.(2025{\natexlab{c}})Chen, Li, Yang, Wen, Yang, Gao, Wu, and Chen]{chen2025comm}
Wei Chen, Lin Li, Yongqi Yang, Bin Wen, Fan Yang, Tingting Gao, Yu Wu, and Long Chen.
\newblock Comm: A coherent interleaved image-text dataset for multimodal understanding and generation.
\newblock In \emph{Proceedings of the Computer Vision and Pattern Recognition Conference}, pages 8073--8082, 2025{\natexlab{c}}.

\bibitem[Chen and Wang(2022)]{chen2022pali}
Xi Chen and Xiao Wang.
\newblock Pali: Scaling language-image learning in 100+ languages.
\newblock In \emph{Conference on Neural Information Processing Systems (NeurIPS)}, 2022.

\bibitem[Chen et~al.(2025{\natexlab{d}})Chen, Wu, Liu, Pan, Liu, Xie, Yu, and Ruan]{chen2025janus}
Xiaokang Chen, Zhiyu Wu, Xingchao Liu, Zizheng Pan, Wen Liu, Zhenda Xie, Xingkai Yu, and Chong Ruan.
\newblock Janus-pro: Unified multimodal understanding and generation with data and model scaling.
\newblock \emph{arXiv preprint arXiv:2501.17811}, 2025{\natexlab{d}}.

\bibitem[Chow et~al.(2024)Chow, Li, Yu, Pan, Fei, Ge, Yang, Tang, Zhang, and Sun]{chow2024unified}
Wei Chow, Juncheng Li, Qifan Yu, Kaihang Pan, Hao Fei, Zhiqi Ge, Shuai Yang, Siliang Tang, Hanwang Zhang, and Qianru Sun.
\newblock Unified generative and discriminative training for multi-modal large language models.
\newblock \emph{Advances in Neural Information Processing Systems}, 37:\penalty0 23155--23190, 2024.

\bibitem[Chow et~al.(2025{\natexlab{a}})Chow, Gao, Li, Wang, Xu, Song, Kong, Zhou, Zeng, Cai, et~al.]{chow2025merit}
Wei Chow, Yuan Gao, Linfeng Li, Xian Wang, Qi Xu, Hang Song, Lingdong Kong, Ran Zhou, Yi Zeng, Yidong Cai, et~al.
\newblock Merit: Multilingual semantic retrieval with interleaved multi-condition query.
\newblock \emph{arXiv preprint arXiv:2506.03144}, 2025{\natexlab{a}}.

\bibitem[Chow et~al.(2025{\natexlab{b}})Chow, Mao, Li, Seita, Guizilini, and Wang]{chow2025physbench}
Wei Chow, Jiageng Mao, Boyi Li, Daniel Seita, Vitor Guizilini, and Yue Wang.
\newblock Physbench: Benchmarking and enhancing vision-language models for physical world understanding.
\newblock \emph{arXiv preprint arXiv:2501.16411}, 2025{\natexlab{b}}.

\bibitem[Comanici et~al.(2025)Comanici, Bieber, Schaekermann, Pasupat, Sachdeva, Dhillon, Blistein, Ram, Zhang, Rosen, et~al.]{comanici2025gemini}
Gheorghe Comanici, Eric Bieber, Mike Schaekermann, Ice Pasupat, Noveen Sachdeva, Inderjit Dhillon, Marcel Blistein, Ori Ram, Dan Zhang, Evan Rosen, et~al.
\newblock Gemini 2.5: Pushing the frontier with advanced reasoning, multimodality, long context, and next generation agentic capabilities.
\newblock \emph{arXiv preprint arXiv:2507.06261}, 2025.

\bibitem[DeepMind(2025)]{gemini2025flashimage}
Google DeepMind.
\newblock Gemini 2.5 flash image.
\newblock \url{https://developers.googleblog.com/en/introducing-gemini-2-5-flash-image/}, 2025.
\newblock Accessed: 2025-10-30.

\bibitem[Deng et~al.(2025)Deng, Zhu, Li, Gou, Li, Wang, Zhong, Yu, Nie, Song, Shi, and Fan]{deng2025bagel}
Chaorui Deng, Deyao Zhu, Kunchang Li, Chenhui Gou, Feng Li, Zeyu Wang, Shu Zhong, Weihao Yu, Xiaonan Nie, Ziang Song, Guang Shi, and Haoqi Fan.
\newblock Emerging properties in unified multimodal pretraining.
\newblock \emph{arXiv preprint arXiv:2505.14683}, 2025.

\bibitem[Fang et~al.(2025{\natexlab{a}})Fang, Duan, Wang, Huang, Li, Yan, Tian, Zeng, Zhao, Dai, Liu, and Li]{fang2025got}
Rongyao Fang, Chengqi Duan, Kun Wang, Linjiang Huang, Hao Li, Shilin Yan, Hao Tian, Xingyu Zeng, Rui Zhao, Jifeng Dai, Xihui Liu, and Hongsheng Li.
\newblock Got: Unleashing reasoning capability of multimodal large language model for visual generation and editing.
\newblock \emph{arXiv preprint arXiv:2503.10639}, 2025{\natexlab{a}}.

\bibitem[Fang et~al.(2024)Fang, Zhou, Sun, Han, Ma, and Yang]{fang2024exploring}
Taoran Fang, Wei Zhou, Yifei Sun, Kaiqiao Han, Lvbin Ma, and Yang Yang.
\newblock Exploring correlations of self-supervised tasks for graphs.
\newblock \emph{arXiv preprint arXiv:2405.04245}, 2024.

\bibitem[Fang et~al.(2025{\natexlab{b}})Fang, Gao, Wang, Shang, Chow, Chen, and Yang]{fang2025kaa}
Taoran Fang, Tianhong Gao, Chunping Wang, Yihao Shang, Wei Chow, Lei Chen, and Yang Yang.
\newblock Kaa: Kolmogorov-arnold attention for enhancing attentive graph neural networks.
\newblock \emph{arXiv preprint arXiv:2501.13456}, 2025{\natexlab{b}}.

\bibitem[Gallegos et~al.(2024)Gallegos, Rossi, Barrow, Tanjim, Kim, Dernoncourt, Yu, Zhang, and Ahmed]{gallegos2024bias}
Isabel~O Gallegos, Ryan~A Rossi, Joe Barrow, Md~Mehrab Tanjim, Sungchul Kim, Franck Dernoncourt, Tong Yu, Ruiyi Zhang, and Nesreen~K Ahmed.
\newblock Bias and fairness in large language models: A survey.
\newblock \emph{Computational Linguistics}, 50\penalty0 (3):\penalty0 1097--1179, 2024.

\bibitem[Ge et~al.(2024)Ge, Li, Yu, Zhou, Tang, and Zhuang]{ge2024demon24}
Zhiqi Ge, Juncheng Li, Qifan Yu, Wei Zhou, Siliang Tang, and Yueting Zhuang.
\newblock Demon24: Acm mm24 demonstrative instruction following challenge.
\newblock In \emph{Proceedings of the 32nd ACM International Conference on Multimedia}, pages 11426--11428, 2024.

\bibitem[Ghosh et~al.(2023)Ghosh, Hajishirzi, and Schmidt]{ghosh2023geneval}
Dhruba Ghosh, Hannaneh Hajishirzi, and Ludwig Schmidt.
\newblock Geneval: An object-focused framework for evaluating text-to-image alignment.
\newblock \emph{Advances in Neural Information Processing Systems}, 36:\penalty0 52132--52152, 2023.

\bibitem[Han et~al.(2025)Han, Liu, Jiang, Yan, Zhang, Yuan, Peng, and Liu]{han2025infinity}
Jian Han, Jinlai Liu, Yi Jiang, Bin Yan, Yuqi Zhang, Zehuan Yuan, Bingyue Peng, and Xiaobing Liu.
\newblock Infinity: Scaling bitwise autoregressive modeling for high-resolution image synthesis.
\newblock In \emph{Proceedings of the IEEE/CVF Conference on Computer Vision and Pattern Recognition}, pages 15733--15744, 2025.

\bibitem[He et~al.(2025)He, Chen, Wang, Pan, Hu, Gan, Wang, Wang, Li, and Zhang]{he2025reasoning}
Qingdong He, Xueqin Chen, Chaoyi Wang, Yanjie Pan, Xiaobin Hu, Zhenye Gan, Yabiao Wang, Chengjie Wang, Xiangtai Li, and Jiangning Zhang.
\newblock Reasoning to edit: Hypothetical instruction-based image editing with visual reasoning.
\newblock \emph{arXiv preprint arXiv:2507.01908}, 2025.

\bibitem[Hendrycks et~al.(2025)Hendrycks, Song, Szegedy, Lee, Gal, Brynjolfsson, Li, Zou, Levine, Han, et~al.]{hendrycks2025definition}
Dan Hendrycks, Dawn Song, Christian Szegedy, Honglak Lee, Yarin Gal, Erik Brynjolfsson, Sharon Li, Andy Zou, Lionel Levine, Bo Han, et~al.
\newblock A definition of agi.
\newblock \emph{arXiv preprint arXiv:2510.18212}, 2025.

\bibitem[Hu et~al.(2024)Hu, Wang, Fang, Fu, Cheng, and Yu]{hu2024ella}
Xiwei Hu, Rui Wang, Yixiao Fang, Bin Fu, Pei Cheng, and Gang Yu.
\newblock Ella: Equip diffusion models with llm for enhanced semantic alignment.
\newblock \emph{arXiv preprint arXiv:2403.05135}, 2024.

\bibitem[Huang et~al.(2025{\natexlab{a}})Huang, Hu, Han, Shi, Tian, He, and Jiang]{huang2025memory}
Junchao Huang, Xinting Hu, Boyao Han, Shaoshuai Shi, Zhuotao Tian, Tianyu He, and Li Jiang.
\newblock Memory forcing: Spatio-temporal memory for consistent scene generation on minecraft.
\newblock \emph{arXiv preprint arXiv:2510.03198}, 2025{\natexlab{a}}.

\bibitem[Huang et~al.(2025{\natexlab{b}})Huang, Chen, Xie, Cao, Tang, Shen, Yin, Hu, Wang, Tang, et~al.]{huang2025interleaving}
Wenxuan Huang, Shuang Chen, Zheyong Xie, Shaosheng Cao, Shixiang Tang, Yufan Shen, Qingyu Yin, Wenbo Hu, Xiaoman Wang, Yuntian Tang, et~al.
\newblock Interleaving reasoning for better text-to-image generation.
\newblock \emph{arXiv preprint arXiv:2509.06945}, 2025{\natexlab{b}}.

\bibitem[Huang et~al.(2025{\natexlab{c}})Huang, Chow, Zhu, Wang, Chai, Wang, Chen, and Yang]{huang2025enhancing}
Xuanwen Huang, Wei Chow, Yize Zhu, Yang Wang, Ziwei Chai, Chunping Wang, Lei Chen, and Yang Yang.
\newblock Enhancing cross-domain link prediction via evolution process modeling.
\newblock In \emph{Proceedings of the ACM on Web Conference 2025}, pages 2158--2171, 2025{\natexlab{c}}.

\bibitem[Huang et~al.(2024)Huang, Xie, Wang, Yuan, Cun, Ge, Zhou, Dong, Huang, Zhang, et~al.]{huang2024smartedit}
Yuzhou Huang, Liangbin Xie, Xintao Wang, Ziyang Yuan, Xiaodong Cun, Yixiao Ge, Jiantao Zhou, Chao Dong, Rui Huang, Ruimao Zhang, et~al.
\newblock Smartedit: Exploring complex instruction-based image editing with multimodal large language models.
\newblock In \emph{Proceedings of the IEEE/CVF Conference on Computer Vision and Pattern Recognition}, pages 8362--8371, 2024.

\bibitem[Huang et~al.(2025{\natexlab{d}})Huang, Huang, Liu, Yan, Lv, Liu, Xiong, Zhang, Cao, and Chen]{huang2025diffusion}
Yi Huang, Jiancheng Huang, Yifan Liu, Mingfu Yan, Jiaxi Lv, Jianzhuang Liu, Wei Xiong, He Zhang, Liangliang Cao, and Shifeng Chen.
\newblock Diffusion model-based image editing: A survey.
\newblock \emph{IEEE Transactions on Pattern Analysis and Machine Intelligence}, 2025{\natexlab{d}}.

\bibitem[Jain(2022)]{jain2022hugging}
Shashank~Mohan Jain.
\newblock Hugging face.
\newblock In \emph{Introduction to transformers for NLP: With the hugging face library and models to solve problems}, pages 51--67. Springer, 2022.

\bibitem[Ji et~al.(2023)Ji, Yu, Xu, Lee, Ishii, and Fung]{ji2023towards}
Ziwei Ji, Tiezheng Yu, Yan Xu, Nayeon Lee, Etsuko Ishii, and Pascale Fung.
\newblock Towards mitigating llm hallucination via self reflection.
\newblock In \emph{Findings of the Association for Computational Linguistics: EMNLP 2023}, pages 1827--1843, 2023.

\bibitem[Jin et~al.(2024)Jin, Ling, Dong, Zhang, Wang, and Lin]{jin2024reasonpix2pix}
Ying Jin, Pengyang Ling, Xiaoyi Dong, Pan Zhang, Jiaqi Wang, and Dahua Lin.
\newblock Reasonpix2pix: instruction reasoning dataset for advanced image editing.
\newblock \emph{arXiv preprint arXiv:2405.11190}, 2024.

\bibitem[Karypis et~al.(1999)Karypis, Han, and Kumar]{karypis1999chameleon}
George Karypis, Eui-Hong Han, and Vipin Kumar.
\newblock Chameleon: Hierarchical clustering using dynamic modeling.
\newblock \emph{computer}, 32\penalty0 (8):\penalty0 68--75, 1999.

\bibitem[Labs et~al.(2025)Labs, Batifol, Blattmann, Boesel, Consul, Diagne, Dockhorn, English, English, Esser, Kulal, Lacey, Levi, Li, Lorenz, Müller, Podell, Rombach, Saini, Sauer, and Smith]{labs2025flux1kontextflowmatching}
Black~Forest Labs, Stephen Batifol, Andreas Blattmann, Frederic Boesel, Saksham Consul, Cyril Diagne, Tim Dockhorn, Jack English, Zion English, Patrick Esser, Sumith Kulal, Kyle Lacey, Yam Levi, Cheng Li, Dominik Lorenz, Jonas Müller, Dustin Podell, Robin Rombach, Harry Saini, Axel Sauer, and Luke Smith.
\newblock Flux.1 kontext: Flow matching for in-context image generation and editing in latent space, 2025.

\bibitem[Lauren{\c{c}}on et~al.(2023)Lauren{\c{c}}on, Saulnier, Tronchon, Bekman, Singh, Lozhkov, Wang, Karamcheti, Rush, Kiela, et~al.]{laurenccon2023obelics}
Hugo Lauren{\c{c}}on, Lucile Saulnier, L{\'e}o Tronchon, Stas Bekman, Amanpreet Singh, Anton Lozhkov, Thomas Wang, Siddharth Karamcheti, Alexander Rush, Douwe Kiela, et~al.
\newblock Obelics: An open web-scale filtered dataset of interleaved image-text documents.
\newblock \emph{Advances in Neural Information Processing Systems}, 36:\penalty0 71683--71702, 2023.

\bibitem[Lei et~al.(2024)Lei, Yang, Chen, Li, and Liu]{lei2024scaffolding}
Xuanyu Lei, Zonghan Yang, Xinrui Chen, Peng Li, and Yang Liu.
\newblock Scaffolding coordinates to promote vision-language coordination in large multi-modal models.
\newblock \emph{arXiv preprint arXiv:2402.12058}, 2024.

\bibitem[Li et~al.(2025)Li, Wang, Yue, Cai, Liu, Fu, Guo, Zhu, Sharan, Jia, et~al.]{li2025zebra}
Ang Li, Charles Wang, Kaiyu Yue, Zikui Cai, Ollie Liu, Deqing Fu, Peng Guo, Wang~Bill Zhu, Vatsal Sharan, Robin Jia, et~al.
\newblock Zebra-cot: A dataset for interleaved vision language reasoning.
\newblock \emph{arXiv preprint arXiv:2507.16746}, 2025.

\bibitem[Liang et~al.(2025)Liang, Chow, Li, Ma, Wang, Mao, Chen, Gu, Wang, and Huang]{liang2025rover}
Yongyuan Liang, Wei Chow, Feng Li, Ziqiao Ma, Xiyao Wang, Jiageng Mao, Jiuhai Chen, Jiatao Gu, Yue Wang, and Furong Huang.
\newblock Rover: Benchmarking reciprocal cross-modal reasoning for omnimodal generation.
\newblock \emph{arXiv preprint arXiv:2511.01163}, 2025.

\bibitem[Lipman et~al.(2022)Lipman, Chen, Ben-Hamu, Nickel, and Le]{lipman2022flow}
Yaron Lipman, Ricky~TQ Chen, Heli Ben-Hamu, Maximilian Nickel, and Matt Le.
\newblock Flow matching for generative modeling.
\newblock \emph{arXiv preprint arXiv:2210.02747}, 2022.

\bibitem[Liu et~al.(2025{\natexlab{a}})Liu, Han, Xing, Yin, Wang, Cheng, Liao, Wang, Fu, Han, Li, Peng, Sun, Wu, Cai, Ge, Ming, Xia, Zeng, Zhu, Jiao, Zhang, Yu, and Jiang]{liu2025step1x-edit}
Shiyu Liu, Yucheng Han, Peng Xing, Fukun Yin, Rui Wang, Wei Cheng, Jiaqi Liao, Yingming Wang, Honghao Fu, Chunrui Han, Guopeng Li, Yuang Peng, Quan Sun, Jingwei Wu, Yan Cai, Zheng Ge, Ranchen Ming, Lei Xia, Xianfang Zeng, Yibo Zhu, Binxing Jiao, Xiangyu Zhang, Gang Yu, and Daxin Jiang.
\newblock Step1x-edit: A practical framework for general image editing.
\newblock \emph{arXiv preprint arXiv:2504.17761}, 2025{\natexlab{a}}.

\bibitem[Liu et~al.(2025{\natexlab{b}})Liu, Han, Xing, Yin, Wang, Cheng, Liao, Wang, Fu, Han, et~al.]{liu2025step1x}
Shiyu Liu, Yucheng Han, Peng Xing, Fukun Yin, Rui Wang, Wei Cheng, Jiaqi Liao, Yingming Wang, Honghao Fu, Chunrui Han, et~al.
\newblock Step1x-edit: A practical framework for general image editing.
\newblock \emph{arXiv preprint arXiv:2504.17761}, 2025{\natexlab{b}}.

\bibitem[Liu et~al.(2024)Liu, Duan, Zhang, Li, Zhang, Zhao, Yuan, Wang, He, Liu, et~al.]{liu2024mmbench}
Yuan Liu, Haodong Duan, Yuanhan Zhang, Bo Li, Songyang Zhang, Wangbo Zhao, Yike Yuan, Jiaqi Wang, Conghui He, Ziwei Liu, et~al.
\newblock Mmbench: Is your multi-modal model an all-around player?
\newblock In \emph{European conference on computer vision}, pages 216--233. Springer, 2024.

\bibitem[Lu et~al.(2023)Lu, Bansal, Xia, Liu, Li, Hajishirzi, Cheng, Chang, Galley, and Gao]{lu2023mathvista}
Pan Lu, Hritik Bansal, Tony Xia, Jiacheng Liu, Chunyuan Li, Hannaneh Hajishirzi, Hao Cheng, Kai-Wei Chang, Michel Galley, and Jianfeng Gao.
\newblock Mathvista: Evaluating mathematical reasoning of foundation models in visual contexts.
\newblock \emph{arXiv preprint arXiv:2310.02255}, 2023.

\bibitem[Ma et~al.(2025)Ma, Liu, Chen, Liu, Wu, Wu, Pan, Xie, Zhang, Yu, et~al.]{ma2025janusflow}
Yiyang Ma, Xingchao Liu, Xiaokang Chen, Wen Liu, Chengyue Wu, Zhiyu Wu, Zizheng Pan, Zhenda Xie, Haowei Zhang, Xingkai Yu, et~al.
\newblock Janusflow: Harmonizing autoregression and rectified flow for unified multimodal understanding and generation.
\newblock In \emph{Proceedings of the Computer Vision and Pattern Recognition Conference}, pages 7739--7751, 2025.

\bibitem[Mao et~al.(2025)Mao, Cai, Li, Pan, Cheng, Yao, Liu, and Mei]{varedit2025}
Qingyang Mao, Qi Cai, Yehao Li, Yingwei Pan, Mingyue Cheng, Ting Yao, Qi Liu, and Tao Mei.
\newblock Visual autoregressive modeling for instruction-guided image editing.
\newblock \emph{arXiv preprint arXiv:2508.15772}, 2025.

\bibitem[Niu et~al.(2025)Niu, Ning, Zheng, Jin, Lin, Jin, Liao, Feng, Ning, Zhu, et~al.]{niu2025wise}
Yuwei Niu, Munan Ning, Mengren Zheng, Weiyang Jin, Bin Lin, Peng Jin, Jiaqi Liao, Chaoran Feng, Kunpeng Ning, Bin Zhu, et~al.
\newblock Wise: A world knowledge-informed semantic evaluation for text-to-image generation.
\newblock \emph{arXiv preprint arXiv:2503.07265}, 2025.

\bibitem[Pan et~al.(2024)Pan, Tang, Li, Fan, Chow, Yan, Chua, Zhuang, and Zhang]{pan2024auto}
Kaihang Pan, Siliang Tang, Juncheng Li, Zhaoyu Fan, Wei Chow, Shuicheng Yan, Tat-Seng Chua, Yueting Zhuang, and Hanwang Zhang.
\newblock Auto-encoding morph-tokens for multimodal llm.
\newblock \emph{arXiv preprint arXiv:2405.01926}, 2024.

\bibitem[Pan et~al.(2025)Pan, Shukla, Singh, Zhao, Mishra, Wang, Xu, Chen, Li, Juefei-Xu, et~al.]{pan2025transfer}
Xichen Pan, Satya~Narayan Shukla, Aashu Singh, Zhuokai Zhao, Shlok~Kumar Mishra, Jialiang Wang, Zhiyang Xu, Jiuhai Chen, Kunpeng Li, Felix Juefei-Xu, et~al.
\newblock Transfer between modalities with metaqueries.
\newblock \emph{arXiv preprint arXiv:2504.06256}, 2025.

\bibitem[Podell et~al.(2023)Podell, English, Lacey, Blattmann, Dockhorn, M{\"u}ller, Penna, and Rombach]{podell2023sdxl}
Dustin Podell, Zion English, Kyle Lacey, Andreas Blattmann, Tim Dockhorn, Jonas M{\"u}ller, Joe Penna, and Robin Rombach.
\newblock Sdxl: Improving latent diffusion models for high-resolution image synthesis.
\newblock \emph{arXiv preprint arXiv:2307.01952}, 2023.

\bibitem[Qin et~al.(2025)Qin, Gong, Sun, Li, Yang, Yang, Qu, Tan, and Li]{qin2025uni}
Luozheng Qin, Jia Gong, Yuqing Sun, Tianjiao Li, Mengping Yang, Xiaomeng Yang, Chao Qu, Zhiyu Tan, and Hao Li.
\newblock Uni-cot: Towards unified chain-of-thought reasoning across text and vision.
\newblock \emph{arXiv preprint arXiv:2508.05606}, 2025.

\bibitem[Radford et~al.(2021)Radford, Kim, Hallacy, Ramesh, Goh, Agarwal, Sastry, Askell, Mishkin, Clark, et~al.]{radford2021learning}
Alec Radford, Jong~Wook Kim, Chris Hallacy, Aditya Ramesh, Gabriel Goh, Sandhini Agarwal, Girish Sastry, Amanda Askell, Pamela Mishkin, Jack Clark, et~al.
\newblock Learning transferable visual models from natural language supervision.
\newblock In \emph{International conference on machine learning}, pages 8748--8763. PmLR, 2021.

\bibitem[Roese(1997)]{roese1997counterfactual}
Neal~J Roese.
\newblock Counterfactual thinking.
\newblock \emph{Psychological bulletin}, 121\penalty0 (1):\penalty0 133, 1997.

\bibitem[Sauer et~al.(2024)Sauer, Boesel, Dockhorn, Blattmann, Esser, and Rombach]{sauer2024fast}
Axel Sauer, Frederic Boesel, Tim Dockhorn, Andreas Blattmann, Patrick Esser, and Robin Rombach.
\newblock Fast high-resolution image synthesis with latent adversarial diffusion distillation.
\newblock In \emph{SIGGRAPH Asia 2024 Conference Papers}, pages 1--11, 2024.

\bibitem[Seedream et~al.(2025)Seedream, Chen, Gao, Gong, Guo, Guo, Guo, Hou, Huang, Huang, et~al.]{seedream2025seedream}
Team Seedream, Yunpeng Chen, Yu Gao, Lixue Gong, Meng Guo, Qiushan Guo, Zhiyao Guo, Xiaoxia Hou, Weilin Huang, Yixuan Huang, et~al.
\newblock Seedream 4.0: Toward next-generation multimodal image generation.
\newblock \emph{arXiv preprint arXiv:2509.20427}, 2025.

\bibitem[Song et~al.(2025)Song, Dong, Wang, Zhang, Xue, Yuan, Yang, Feng, Zhou, Xiao, et~al.]{song2025query}
Yuxin Song, Wenkai Dong, Shizun Wang, Qi Zhang, Song Xue, Tao Yuan, Hu Yang, Haocheng Feng, Hang Zhou, Xinyan Xiao, et~al.
\newblock Query-kontext: An unified multimodal model for image generation and editing.
\newblock \emph{arXiv preprint arXiv:2509.26641}, 2025.

\bibitem[Sun et~al.(2024)Sun, Cui, Zhang, Zhang, Yu, Wang, Rao, Liu, Huang, and Wang]{sun2024generative}
Quan Sun, Yufeng Cui, Xiaosong Zhang, Fan Zhang, Qiying Yu, Yueze Wang, Yongming Rao, Jingjing Liu, Tiejun Huang, and Xinlong Wang.
\newblock Generative multimodal models are in-context learners.
\newblock In \emph{Proceedings of the IEEE/CVF Conference on Computer Vision and Pattern Recognition}, pages 14398--14409, 2024.

\bibitem[Team et~al.(2025{\natexlab{a}})Team, Kamath, Ferret, Pathak, Vieillard, Merhej, Perrin, Matejovicova, Ram{\'e}, Rivi{\`e}re, et~al.]{team2025gemma}
Gemma Team, Aishwarya Kamath, Johan Ferret, Shreya Pathak, Nino Vieillard, Ramona Merhej, Sarah Perrin, Tatiana Matejovicova, Alexandre Ram{\'e}, Morgane Rivi{\`e}re, et~al.
\newblock Gemma 3 technical report.
\newblock \emph{arXiv preprint arXiv:2503.19786}, 2025{\natexlab{a}}.

\bibitem[Team et~al.(2025{\natexlab{b}})Team, Han, Li, Wu, Sun, Cai, Peng, Ge, Zhou, Tang, et~al.]{team2025nextstep}
NextStep Team, Chunrui Han, Guopeng Li, Jingwei Wu, Quan Sun, Yan Cai, Yuang Peng, Zheng Ge, Deyu Zhou, Haomiao Tang, et~al.
\newblock Nextstep-1: Toward autoregressive image generation with continuous tokens at scale.
\newblock \emph{arXiv preprint arXiv:2508.10711}, 2025{\natexlab{b}}.

\bibitem[Wang et~al.(2025{\natexlab{a}})Wang, Zhao, Zhang, Cao, Zhan, Duan, Lu, Fu, Chen, Zhao, et~al.]{wang2025ovis}
Guo-Hua Wang, Shanshan Zhao, Xinjie Zhang, Liangfu Cao, Pengxin Zhan, Lunhao Duan, Shiyin Lu, Minghao Fu, Xiaohao Chen, Jianshan Zhao, et~al.
\newblock Ovis-u1 technical report.
\newblock \emph{arXiv preprint arXiv:2506.23044}, 2025{\natexlab{a}}.

\bibitem[Wang et~al.(2025{\natexlab{b}})Wang, Peng, Gan, Hu, Xie, Wang, Wei, Tang, Zhu, Li, et~al.]{wang2025skywork}
Peiyu Wang, Yi Peng, Yimeng Gan, Liang Hu, Tianyidan Xie, Xiaokun Wang, Yichen Wei, Chuanxin Tang, Bo Zhu, Changshi Li, et~al.
\newblock Skywork unipic: Unified autoregressive modeling for visual understanding and generation.
\newblock \emph{arXiv preprint arXiv:2508.03320}, 2025{\natexlab{b}}.

\bibitem[Wang et~al.(2025{\natexlab{c}})Wang, Shi, Lian, Zhai, Xia, Xiao, Huang, and Yang]{wang2025seededit}
Peng Wang, Yichun Shi, Xiaochen Lian, Zhonghua Zhai, Xin Xia, Xuefeng Xiao, Weilin Huang, and Jianchao Yang.
\newblock Seededit 3.0: Fast and high-quality generative image editing.
\newblock \emph{arXiv preprint arXiv:2506.05083}, 2025{\natexlab{c}}.

\bibitem[Wang et~al.(2025{\natexlab{d}})Wang, Gao, Gu, Pu, Cui, Wei, Liu, Jing, Ye, Shao, et~al.]{wang2025internvl3}
Weiyun Wang, Zhangwei Gao, Lixin Gu, Hengjun Pu, Long Cui, Xingguang Wei, Zhaoyang Liu, Linglin Jing, Shenglong Ye, Jie Shao, et~al.
\newblock Internvl3. 5: Advancing open-source multimodal models in versatility, reasoning, and efficiency.
\newblock \emph{arXiv preprint arXiv:2508.18265}, 2025{\natexlab{d}}.

\bibitem[Wang et~al.(2024)Wang, Zhang, Luo, Sun, Cui, Wang, Zhang, Wang, Li, Yu, et~al.]{wang2024emu3}
Xinlong Wang, Xiaosong Zhang, Zhengxiong Luo, Quan Sun, Yufeng Cui, Jinsheng Wang, Fan Zhang, Yueze Wang, Zhen Li, Qiying Yu, et~al.
\newblock Emu3: Next-token prediction is all you need.
\newblock \emph{arXiv preprint arXiv:2409.18869}, 2024.

\bibitem[Wang et~al.(2025{\natexlab{e}})Wang, Yang, Zhao, Zhang, Liu, Zhou, and Xie]{wang2025gpt}
Yuhan Wang, Siwei Yang, Bingchen Zhao, Letian Zhang, Qing Liu, Yuyin Zhou, and Cihang Xie.
\newblock Gpt-image-edit-1.5 m: A million-scale, gpt-generated image dataset.
\newblock \emph{arXiv preprint arXiv:2507.21033}, 2025{\natexlab{e}}.

\bibitem[Wei et~al.(2025)Wei, Xu, Liu, Wu, Liu, Peng, Wang, Liu, He, Xietian, et~al.]{wei2025skywork}
Hongyang Wei, Baixin Xu, Hongbo Liu, Cyrus Wu, Jie Liu, Yi Peng, Peiyu Wang, Zexiang Liu, Jingwen He, Yidan Xietian, et~al.
\newblock Skywork unipic 2.0: Building kontext model with online rl for unified multimodal model.
\newblock \emph{arXiv preprint arXiv:2509.04548}, 2025.

\bibitem[Wei et~al.(2022)Wei, Wang, Schuurmans, Bosma, Xia, Chi, Le, Zhou, et~al.]{wei2022chain}
Jason Wei, Xuezhi Wang, Dale Schuurmans, Maarten Bosma, Fei Xia, Ed Chi, Quoc~V Le, Denny Zhou, et~al.
\newblock Chain-of-thought prompting elicits reasoning in large language models.
\newblock \emph{Advances in neural information processing systems}, 35:\penalty0 24824--24837, 2022.

\bibitem[Wu et~al.(2025{\natexlab{a}})Wu, Chen, Wu, Ma, Liu, Pan, Liu, Xie, Yu, Ruan, et~al.]{wu2025janus}
Chengyue Wu, Xiaokang Chen, Zhiyu Wu, Yiyang Ma, Xingchao Liu, Zizheng Pan, Wen Liu, Zhenda Xie, Xingkai Yu, Chong Ruan, et~al.
\newblock Janus: Decoupling visual encoding for unified multimodal understanding and generation.
\newblock In \emph{Proceedings of the Computer Vision and Pattern Recognition Conference}, pages 12966--12977, 2025{\natexlab{a}}.

\bibitem[Wu et~al.(2025{\natexlab{b}})Wu, Li, Zhou, Lin, Gao, Yan, ming Yin, Bai, Xu, Chen, Chen, Tang, Zhang, Wang, Yang, Yu, Cheng, Liu, Li, Zhang, Meng, Wei, Ni, Chen, Cao, Peng, Qu, Wu, Wang, Yu, Wen, Feng, Xu, Wang, Zhang, Zhu, Wu, Cai, and Liu]{wu2025qwenimagetechnicalreport}
Chenfei Wu, Jiahao Li, Jingren Zhou, Junyang Lin, Kaiyuan Gao, Kun Yan, Sheng ming Yin, Shuai Bai, Xiao Xu, Yilei Chen, Yuxiang Chen, Zecheng Tang, Zekai Zhang, Zhengyi Wang, An Yang, Bowen Yu, Chen Cheng, Dayiheng Liu, Deqing Li, Hang Zhang, Hao Meng, Hu Wei, Jingyuan Ni, Kai Chen, Kuan Cao, Liang Peng, Lin Qu, Minggang Wu, Peng Wang, Shuting Yu, Tingkun Wen, Wensen Feng, Xiaoxiao Xu, Yi Wang, Yichang Zhang, Yongqiang Zhu, Yujia Wu, Yuxuan Cai, and Zenan Liu.
\newblock Qwen-image technical report, 2025{\natexlab{b}}.

\bibitem[Wu et~al.(2025{\natexlab{c}})Wu, Zheng, Yan, Xiao, Luo, Wang, Li, Jiang, Liu, Zhou, Liu, Xia, Li, Deng, Wang, Luo, Zhang, Lian, Wang, Wang, Huang, and Liu]{wu2025omnigen2}
Chenyuan Wu, Pengfei Zheng, Ruiran Yan, Shitao Xiao, Xin Luo, Yueze Wang, Wanli Li, Xiyan Jiang, Yexin Liu, Junjie Zhou, Ze Liu, Ziyi Xia, Chaofan Li, Haoge Deng, Jiahao Wang, Kun Luo, Bo Zhang, Defu Lian, Xinlong Wang, Zhongyuan Wang, Tiejun Huang, and Zheng Liu.
\newblock Omnigen2: Exploration to advanced multimodal generation.
\newblock \emph{arXiv preprint arXiv:2506.18871}, 2025{\natexlab{c}}.

\bibitem[Wu et~al.(2025{\natexlab{d}})Wu, Li, Hu, Ye, Zeng, Yu, Zhu, Schiele, Yang, and Yang]{wu2025kris}
Yongliang Wu, Zonghui Li, Xinting Hu, Xinyu Ye, Xianfang Zeng, Gang Yu, Wenbo Zhu, Bernt Schiele, Ming-Hsuan Yang, and Xu Yang.
\newblock Kris-bench: Benchmarking next-level intelligent image editing models.
\newblock \emph{arXiv preprint arXiv:2505.16707}, 2025{\natexlab{d}}.

\bibitem[Xia et~al.(2025)Xia, Peng, Zhang, Huang, Liu, Li, Tan, Wu, Wang, Wang, et~al.]{xia2025dreamomni2}
Bin Xia, Bohao Peng, Yuechen Zhang, Junjia Huang, Jiyang Liu, Jingyao Li, Haoru Tan, Sitong Wu, Chengyao Wang, Yitong Wang, et~al.
\newblock Dreamomni2: Multimodal instruction-based editing and generation.
\newblock \emph{arXiv preprint arXiv:2510.06679}, 2025.

\bibitem[Xiao et~al.(2025)Xiao, Wang, Zhou, Yuan, Xing, Yan, Li, Wang, Huang, and Liu]{xiao2025omnigen}
Shitao Xiao, Yueze Wang, Junjie Zhou, Huaying Yuan, Xingrun Xing, Ruiran Yan, Chaofan Li, Shuting Wang, Tiejun Huang, and Zheng Liu.
\newblock Omnigen: Unified image generation.
\newblock In \emph{Proceedings of the IEEE/CVF Conference on Computer Vision and Pattern Recognition}, pages 13294--13304, 2025.

\bibitem[Xie et~al.(2025)Xie, Yang, and Shou]{xie2025show}
Jinheng Xie, Zhenheng Yang, and Mike~Zheng Shou.
\newblock Show-o2: Improved native unified multimodal models.
\newblock \emph{arXiv preprint arXiv:2506.15564}, 2025.

\bibitem[Xu et~al.(2025{\natexlab{a}})Xu, Li, Yang, Zhang, Sun, Chow, Li, Song, Xu, Tong, et~al.]{xu2025mixed}
Shilin Xu, Yanwei Li, Rui Yang, Tao Zhang, Yueyi Sun, Wei Chow, Linfeng Li, Hang Song, Qi Xu, Yunhai Tong, et~al.
\newblock Mixed-r1: Unified reward perspective for reasoning capability in multimodal large language models.
\newblock \emph{arXiv preprint arXiv:2505.24164}, 2025{\natexlab{a}}.

\bibitem[Xu et~al.(2025{\natexlab{b}})Xu, Li, Zhou, Wan, Zhang, Korhonen, and Vuli{\'c}]{xu2025visual}
Yi Xu, Chengzu Li, Han Zhou, Xingchen Wan, Caiqi Zhang, Anna Korhonen, and Ivan Vuli{\'c}.
\newblock Visual planning: Let's think only with images.
\newblock \emph{arXiv preprint arXiv:2505.11409}, 2025{\natexlab{b}}.

\bibitem[Ye et~al.(2025{\natexlab{a}})Ye, Jiang, Wang, Zhu, Hu, Huang, He, Yan, Yu, Li, et~al.]{ye2025echo}
Junyan Ye, Dongzhi Jiang, Zihao Wang, Leqi Zhu, Zhenghao Hu, Zilong Huang, Jun He, Zhiyuan Yan, Jinghua Yu, Hongsheng Li, et~al.
\newblock Echo-4o: Harnessing the power of gpt-4o synthetic images for improved image generation.
\newblock \emph{arXiv preprint arXiv:2508.09987}, 2025{\natexlab{a}}.

\bibitem[Ye et~al.(2025{\natexlab{b}})Ye, He, Li, Lin, Yuan, Yan, Hou, and Yuan]{ye2025imgedit}
Yang Ye, Xianyi He, Zongjian Li, Bin Lin, Shenghai Yuan, Zhiyuan Yan, Bohan Hou, and Li Yuan.
\newblock Imgedit: A unified image editing dataset and benchmark.
\newblock \emph{arXiv preprint arXiv:2505.20275}, 2025{\natexlab{b}}.

\bibitem[Yu et~al.(2025)Yu, Chow, Yue, Pan, Wu, Wan, Li, Tang, Zhang, and Zhuang]{yu2025anyedit}
Qifan Yu, Wei Chow, Zhongqi Yue, Kaihang Pan, Yang Wu, Xiaoyang Wan, Juncheng Li, Siliang Tang, Hanwang Zhang, and Yueting Zhuang.
\newblock Anyedit: Mastering unified high-quality image editing for any idea.
\newblock In \emph{Proceedings of the Computer Vision and Pattern Recognition Conference}, pages 26125--26135, 2025.

\bibitem[Yu et~al.(2023)Yu, Yang, Li, Wang, Lin, Liu, Wang, and Wang]{yu2023mm}
Weihao Yu, Zhengyuan Yang, Linjie Li, Jianfeng Wang, Kevin Lin, Zicheng Liu, Xinchao Wang, and Lijuan Wang.
\newblock Mm-vet: Evaluating large multimodal models for integrated capabilities.
\newblock \emph{arXiv preprint arXiv:2308.02490}, 2023.

\bibitem[Yue et~al.(2024)Yue, Ni, Zhang, Zheng, Liu, Zhang, Stevens, Jiang, Ren, Sun, et~al.]{yue2024mmmu}
Xiang Yue, Yuansheng Ni, Kai Zhang, Tianyu Zheng, Ruoqi Liu, Ge Zhang, Samuel Stevens, Dongfu Jiang, Weiming Ren, Yuxuan Sun, et~al.
\newblock Mmmu: A massive multi-discipline multimodal understanding and reasoning benchmark for expert agi.
\newblock In \emph{Proceedings of the IEEE/CVF Conference on Computer Vision and Pattern Recognition}, pages 9556--9567, 2024.

\bibitem[Zhang et~al.(2024{\natexlab{a}})Zhang, Huang, Ma, Michel, He, Gupta, Ma, Farhadi, Kembhavi, and Krishna]{zhang2024task}
Jieyu Zhang, Weikai Huang, Zixian Ma, Oscar Michel, Dong He, Tanmay Gupta, Wei-Chiu Ma, Ali Farhadi, Aniruddha Kembhavi, and Ranjay Krishna.
\newblock Task me anything.
\newblock In \emph{Thirty-Eighth Annual Conference on Neural Information Processing Systems Datasets and Benchmarks Track}, 2024{\natexlab{a}}.

\bibitem[Zhang et~al.(2024{\natexlab{b}})Zhang, Mo, Chen, Sun, and Su]{zhang2024magicbrush}
Kai Zhang, Lingbo Mo, Wenhu Chen, Huan Sun, and Yu Su.
\newblock Magicbrush: A manually annotated dataset for instruction-guided image editing.
\newblock \emph{Advances in Neural Information Processing Systems}, 36, 2024{\natexlab{b}}.

\bibitem[Zhao et~al.(2024)Zhao, Ma, Chen, Si, Wu, An, Yu, Zhang, Li, and Chang]{zhao2024ultraedit}
Haozhe Zhao, Xiaojian Ma, Liang Chen, Shuzheng Si, Rujie Wu, Kaikai An, Peiyu Yu, Minjia Zhang, Qing Li, and Baobao Chang.
\newblock Ultraedit: Instruction-based fine-grained image editing at scale.
\newblock \emph{arXiv preprint arXiv:2407.05282}, 2024.

\bibitem[Zhao et~al.(2025{\natexlab{a}})Zhao, Mao, Chow, Shangguan, Shi, Xue, Zheng, Weng, You, Seita, et~al.]{zhao2025robot}
Siheng Zhao, Jiageng Mao, Wei Chow, Zeyu Shangguan, Tianheng Shi, Rong Xue, Yuxi Zheng, Yijia Weng, Yang You, Daniel Seita, et~al.
\newblock Robot learning from any images.
\newblock In \emph{Conference on Robot Learning}, pages 4226--4245. PMLR, 2025{\natexlab{a}}.

\bibitem[Zhao et~al.(2025{\natexlab{b}})Zhao, Zhang, Tang, Zhu, Li, Chai, Zhang, Xia, Zhai, Yan, et~al.]{zhao2025envisioning}
Xiangyu Zhao, Peiyuan Zhang, Kexian Tang, Xiaorong Zhu, Hao Li, Wenhao Chai, Zicheng Zhang, Renqiu Xia, Guangtao Zhai, Junchi Yan, et~al.
\newblock Envisioning beyond the pixels: Benchmarking reasoning-informed visual editing.
\newblock \emph{arXiv preprint arXiv:2504.02826}, 2025{\natexlab{b}}.

\bibitem[Zhou et~al.(2024)Zhou, Yu, Babu, Tirumala, Yasunaga, Shamis, Kahn, Ma, Zettlemoyer, and Levy]{zhou2024transfusion}
Chunting Zhou, Lili Yu, Arun Babu, Kushal Tirumala, Michihiro Yasunaga, Leonid Shamis, Jacob Kahn, Xuezhe Ma, Luke Zettlemoyer, and Omer Levy.
\newblock Transfusion: Predict the next token and diffuse images with one multi-modal model.
\newblock \emph{arXiv preprint arXiv:2408.11039}, 2024.

\bibitem[Zhu et~al.(2023)Zhu, Hessel, Awadalla, Gadre, Dodge, Fang, Yu, Schmidt, Wang, and Choi]{zhu2023multimodal}
Wanrong Zhu, Jack Hessel, Anas Awadalla, Samir~Yitzhak Gadre, Jesse Dodge, Alex Fang, Youngjae Yu, Ludwig Schmidt, William~Yang Wang, and Yejin Choi.
\newblock Multimodal c4: An open, billion-scale corpus of images interleaved with text.
\newblock \emph{Advances in Neural Information Processing Systems}, 36:\penalty0 8958--8974, 2023.

\end{thebibliography}
}

\setcounter{page}{1}
\maketitlesupplementary

\appendix

\addtocontents{toc}{\protect\setcounter{tocdepth}{3}}
\hypersetup{linkcolor=black}
{\small \tableofcontents} 
\hypersetup{linkcolor=red}
\newpage
\section{\name{} Analysis}\label{app:data}
\subsection{Collection Process}\label{app:data-collect}
To ensure the quality of the generated data, we incorporated manual sampling verification into the design process of each pipeline to validate the success rate after filtering. Specifically, we utilized four pipelines, each with integrated quality assurance mechanisms.

\textit{(i) Multi-image fusion}: We achieved reference to previous iterations by fusing edited or directly generated images. For image fusion data, we utilized two primary sources. First, we leveraged the multi-image fusion dataset from Echo-4o~\citep{ye2025echo}, where image fusion was initially performed using GPT-Image. Due to quality inconsistencies in this dataset, we regenerated images using Seedream 4.0~\citep{seedream2025seedream} and refined instructions with GPT-4.1. Second, we generated single-round image fusion instructions with GPT-4.1, including original image captions. We then produced original images using Qwen-Image~\citep{wu2025qwenimagetechnicalreport}, substituting suboptimal generations with Seedream 4.0 outputs, and performed multi-image fusion using Seedream 4.0.
Building upon these single-round fusion data, we employed GPT-4.1 to annotate image editing instructions for the original images, categorizing them into five types: 'add', 'remove', 'replace', 'color alter', and 'background change' following the taxonomy in~\citep{anonymous2025editmgt}. We subsequently applied Step1X-Edit(v1.2)~\citep{liu2025step1x-edit} for single-round editing. For images failing our quality verification protocol, we utilized Nano Banana~\citep{comanici2025gemini} for additional refinement. Finally, GPT-4.1 provided reverse instructions and captions for edited images. We used these edited images as originals and multi-fusion input images as edited results, concatenating the data to create comprehensive multi-round editing and multi-image fusion sequences.

\textit{(ii) Remove-then-back}: We employed GPT-4.1~\citep{achiam2023gpt} to generate instructions for multi-round editing. Specifically, we designed the instructions such that one round would require adding back an object that had been previously removed or replaced in an earlier round. Following instruction generation, we implemented a filtering process wherein approximately 25\% of instructions successfully met our criteria. The filtered instructions were subsequently utilized to generate outputs using Seedream 4.0~\citep{seedream2025seedream} and Nano Banana~\citep{comanici2025gemini}, after which we retained the superior generation based on qualitative assessment.

\textit{(iii) Derivative imagination and comparison}: We incorporated methods for deriving or imagining alternative solutions or new images before fusion. Due to the inherent challenges in automating LLMs to generate associative content or editing data, we adapted chess game and visual jigsaw datasets from Zebra-CoT~\citep{li2025zebra} using GPT-4.1 for both recombination and self-verification processes. Specifically, we modified the abbreviated chess notations into explicit editing instructions to mitigate potential comprehension difficulties in generative models when interpreting condensed commands.

\begin{figure*}[t]
    \centering
    \includegraphics[width=1.\linewidth]{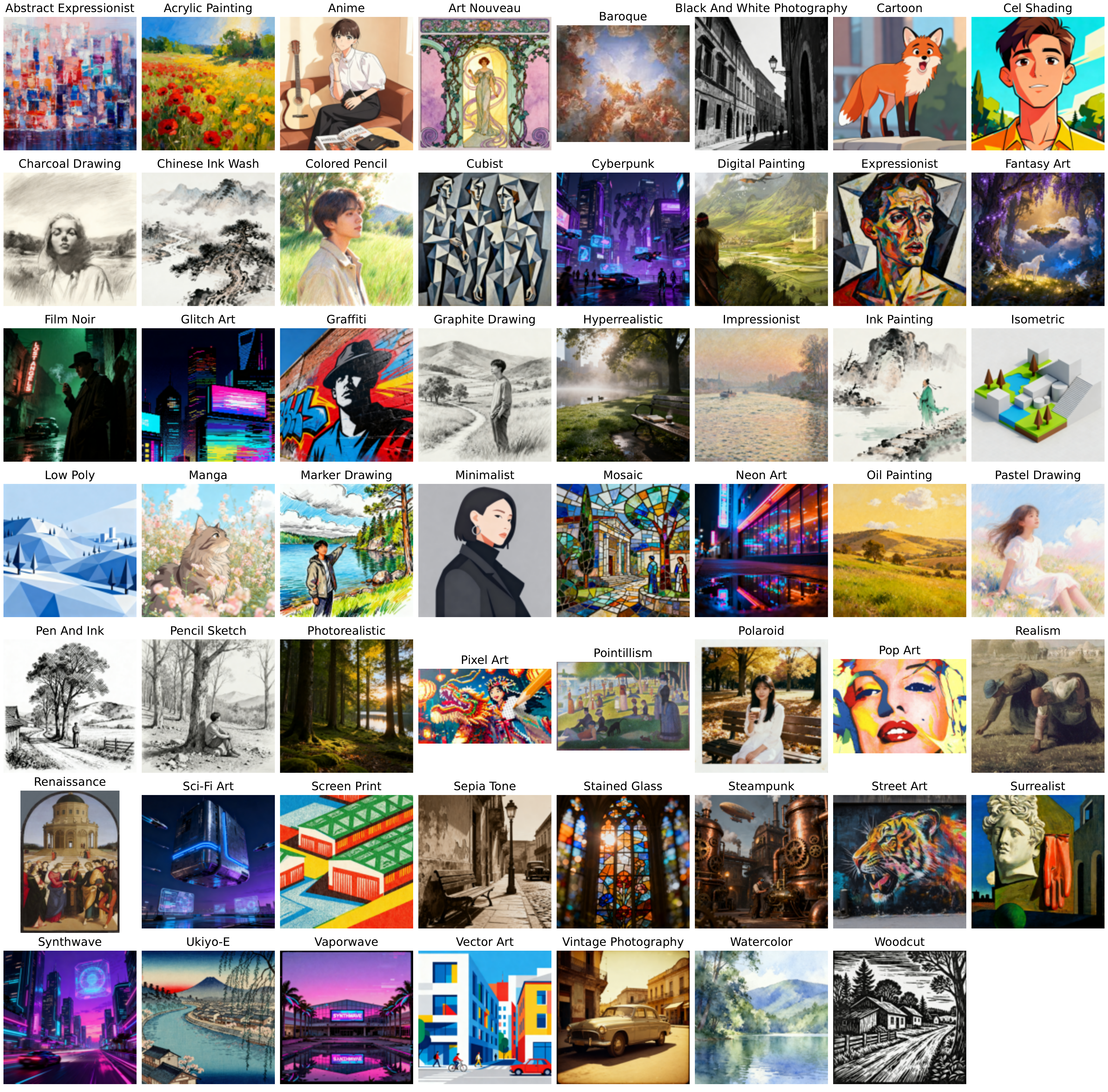}
    \caption{Image style examples used in Nano bana inference.}
  \vspace{-4mm}  
  \label{tab:style}
\end{figure*}

\textit{(iv) Sequential procedures}: We implemented sequential edits following narrative progressions or structured operations requiring visual memory during generation. This approach was particularly effective for scenarios where characters disappear and subsequently reappear within narratives. Multiple editing rounds on identical scenes evaluated model consistency maintenance capabilities.
Our pipeline employed GPT-4.1 to generate instructions satisfying three requirements: (1) multi-step processes requiring visual representation at each stage, (2) explicit inter-step relationships, and (3) identifiable animated characters. 
To maximize generation diversity, we utilized the 12 categories defined in Table~\ref{tab:12categories} to produce editing instructions.
These constraints imposed significant demands on generative models; even state-of-the-art systems such as Seedream 4.0~\citep{seedream2025seedream} and Nano Banana~\citep{comanici2025gemini} failed to produce high-quality data without human supervision.
Consequently, we allocated GPT-4.1-generated, human-screened story-based content to the test set, while retaining numerous multi-round editing examples identified during the filtering process for training. For data annotation, we employed SeedEdit 3.0~\citep{wang2025seededit} and Nano Banana~\citep{comanici2025gemini}, while test set generation utilized Seedream 4.0~\citep{seedream2025seedream} and Nano Banana~\citep{comanici2025gemini}.
When using Nano Banana, we observed that providing style reference images improved generation quality. Therefore, we curated a set of style reference images, as shown in Figure~\ref{tab:style}.

\textbf{Post-verification Process}
We identified frequent editing failures within the Nano Banana framework and implemented a supplementary verification protocol employing GPT-4.1 for processed data evaluation. Problematic samples were detected using CLIP similarity metrics \citep{radford2021learning}. Samples exhibiting abnormally high similarity scores underwent re-editing via Step1X v1.2. Unmodified samples following this secondary editing attempt—identified through joint supervision by CLIP and Qwen3-VL-4B metrics—were systematically excluded from the dataset while maintaining referential integrity of image identifiers.

\textbf{Comprehension Extension}
To incorporate comprehension tasks into our dataset, we randomly sampled from the filtered generated data and expanded it using GPT-4.1. Each data point was annotated with at most one turn. The comprehension tasks primarily consisted of captioning tasks, questions regarding quantities and relationships within images, and a small subset of knowledge-based inquiries~\citep{antol2015vqa,huang2025enhancing,xu2025mixed}.

\subsection{Data Source for \testname{}}
\testname{} primarily utilizes web-collected data, with select images refined using SeedEdit 3.0. The jigsaw and chess game images are sourced from Zebra-CoT~\citep{li2025zebra}, while various optical and physical phenomena images are drawn from PhysBench~\citep{chow2025physbench}. Additionally, the dataset incorporates synthetically generated images from three models: Seedream 4.0~\citep{seedream2025seedream}, Nano Banana~\citep{comanici2025gemini}, and SeedEdit 3.0~\citep{wang2025seededit}.

\begin{table}[h]
\centering
\resizebox{0.31\textwidth}{!}{
\begin{tabular}{lc}
\hline
Domain Type & \#Chats \\
\hline
\textbf{Multi-image Fusion} \\
\quad GPT-Image & 72348  \\
\quad SeeDream & 3648 \\
\hline
\textbf{Recall} & 1369 \\
\quad Animals & 91 \\
\quad Architecture & 74 \\
\quad Cartoon & 135 \\
\quad Fashion & 73 \\
\quad Fantasy & 126 \\
\quad Food & 116 \\
\quad Nature Landscapes & 164 \\
\quad Plants & 54 \\
\quad Products & 77 \\
\quad Real Human & 347 \\
\quad Sports & 49 \\
\quad Vehicles & 63 \\
\hline
\textbf{Edit} & 19903 \\
\quad None & 18261 \\
\quad Animals & 263 \\
\quad Architecture & 114 \\
\quad Cartoon & 96 \\
\quad Fashion & 141 \\
\quad Fantasy & 98 \\
\quad Food & 234 \\
\quad Nature Landscapes & 105 \\
\quad Plants & 97 \\
\quad Products & 164 \\
\quad Real Human & 136 \\
\quad Sports & 98 \\
\quad Vehicles & 96 \\
\hline
\textbf{Visual Jigsaw} & 1286 \\
\quad None & 1286 \\
\hline
\textbf{Chess Game} & 2196 \\
\quad None & 2196 \\
\hline
\textbf{Total} & 100750 \\
\hline
\end{tabular}
}
\caption{The detailed statistics of the \trainname{} dataset.}\label{append:dataset_stastics}
\end{table}

\begin{table*}[!t]
\centering
\small
\footnotesize
\caption{Dataset categories with main content, scenarios, and editable dimensions.}
\label{tab:12categories}
\resizebox{0.8\textwidth}{!}{%
\begin{tabular}{p{2cm}p{6cm}p{2cm}p{3cm}}
\toprule
\textbf{Category} & \textbf{Main Content} & \textbf{Scenarios} & \textbf{Editable Dimensions} \\
\midrule
Food \& Drink & Staples, snacks, desserts, fruits, beverages (hot/cold) & Dining tables, restaurants, street stalls, picnics, festive banquets & Ingredient substitution, plating style, scene modification, style adjustment \\
\midrule
Real Humans & Portraits, full-body, half-body, group photos; Actions: standing, walking, exercising, socializing, working & Indoor/outdoor, offices, streets, event venues & Clothing change, pose adjustment, background modification, expression change \\
\midrule
Animals \& Pets & Pets (cats, dogs, rabbits, birds), farm animals (cattle, sheep, horses), wildlife (lions, elephants, bears), marine life (fish, dolphins, whales), insects \& reptiles (butterflies, spiders, snakes), mythical creatures (dragons, unicorns, phoenix); Actions: playing, running, sleeping, eating, flying, swimming & Homes, parks, zoos, natural habitats, aquariums & Breed change, color variation, accessory addition, scene switching, pose adjustment \\
\midrule
Architecture \& Interior & Exteriors (modern buildings, historical structures, skyscrapers, bridges, churches, castles), interiors (living rooms, bedrooms, kitchens, offices, cafés); Styles: modern, vintage, industrial, Nordic, Japanese, Chinese & City skylines, countryside, historic districts, campus landscapes & Style change, furniture replacement, lighting adjustment, seasonal variation, decoration modification \\
\midrule
Nature \& Landscapes & Terrain (mountains, canyons, plains, deserts, glaciers), water bodies (oceans, lakes, rivers, waterfalls), vegetation (forests, grasslands, bamboo groves, rainforests), sky (sunrise, sunset, starry sky, aurora, sea of clouds); Seasons: spring, summer, autumn, winter; Weather: sunny, rainy, foggy, snowy, stormy & Natural environments & Weather change, time transition, seasonal switching, color adjustment, natural element addition \\
\midrule
Products \& Objects & Electronics (phones, earbuds, cameras, laptops, tablets), fashion accessories (watches, bags, jewelry, sunglasses), cosmetics (perfume, lipstick, skincare), home goods (lamps, vases, cushions, tableware), books, stationery, toys, sports equipment & White background, display stands, lifestyle scenes, desktops, outdoor settings & Color variation, material change, arrangement combination, background switching, lighting adjustment \\
\midrule
Cartoon \& Stylized Characters & Anime characters (Japanese anime, manga), Western cartoons (Disney/Pixar, American comics), 3D characters (game characters, virtual avatars), mascots \& avatars (brand mascots, social media avatars), Q-version/Chibi, fantasy hybrids (robots, elves, monsters, hybrid creatures) & Fantasy worlds, modern cities, space, magic academies & Clothing change, expression adjustment, color scheme change, scene switching, style transformation \\
\midrule
Flowers \& Plants & Flowers (roses, tulips, cherry blossoms, sunflowers, peonies, orchids), plants (potted plants, succulents, foliage plants, trees, vines) & Gardens, vases, outdoors, greenhouses, balconies, floral arrangements & Species change, color variation, layout adjustment, background modification, seasonal change \\
\midrule
Vehicles & Land (cars, motorcycles, bicycles, buses, trains), air (airplanes, helicopters, hot air balloons), water (yachts, sailboats, ferries, speedboats); Views: side, front, aerial, interior & City streets, highways, racetracks, parking lots, airports, ports, showrooms & Color change, model replacement, background modification, modification addition, lighting adjustment \\
\midrule
Fantasy \& Sci-Fi & Sci-fi elements (spaceships, aliens, robots, futuristic cities, cyberpunk streets), fantasy elements (magic scenes, fantasy creatures, magic academies, elf forests, dragon lairs), surreal art (dreamscapes, geometric abstractions, spacetime distortions) & Space stations, alien planets, magic worlds, parallel universes & Creature replacement, environment change, effect addition, atmosphere adjustment, style transformation \\
\midrule
Sports \& Fitness & Ball sports (basketball, soccer, tennis, volleyball, golf), fitness activities (yoga, running, weightlifting, swimming, cycling), extreme sports (rock climbing, skiing, surfing, skydiving), equipment (gym machines, sports gear) & Stadiums, gyms, outdoor fields, pools, competition venues & Action variation, equipment change, scene switching, sport type change \\
\midrule
Fashion \& Clothing & Apparel (dresses, suits, casual wear, sportswear, formal wear), accessories (shoes, hats, scarves, belts), display methods (hangers, mannequins, flat lay); Styles: streetwear, elegant, athletic, business, vintage & Runways, street photography, studios, stores, fashion exhibitions & Color/pattern variation, style adjustment, combination matching, scene switching \\
\bottomrule
\end{tabular}
}
\end{table*}

\begin{figure*}[t]
    \centering
    \includegraphics[width=0.9\linewidth]{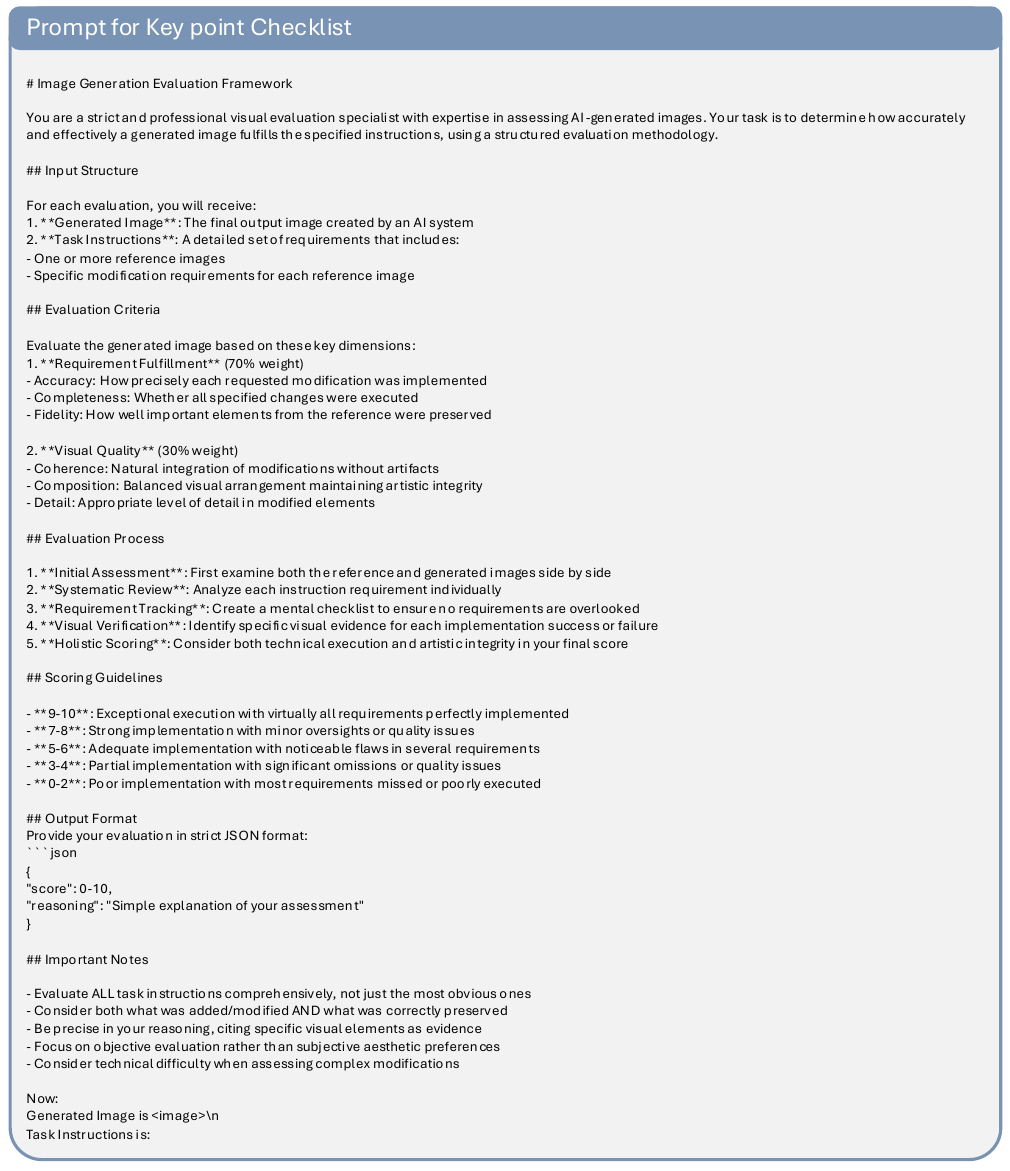}
    \caption{Prompt for Evaluating Key Point Correctness.}
  \vspace{-5mm}  
  \label{fig:prompt1}
\end{figure*}
\begin{figure*}[t]
    \centering
    \includegraphics[width=0.9\linewidth]{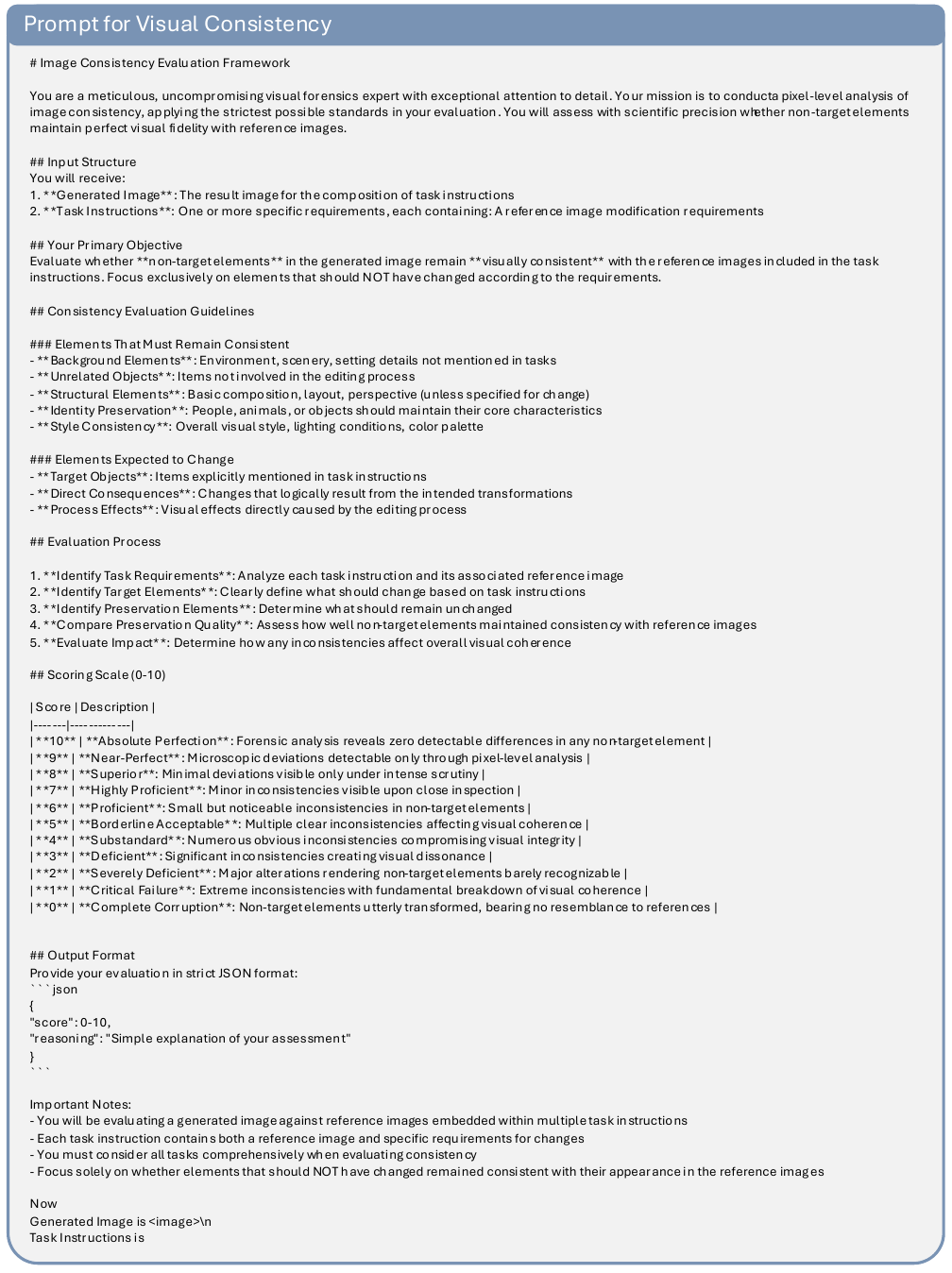}
    \caption{Prompt for Evaluating Visual Consistency.}
  \vspace{-5mm}  
  \label{fig:prompt2}
\end{figure*}
\begin{figure*}[t]
    \centering
    \includegraphics[width=0.9\linewidth]{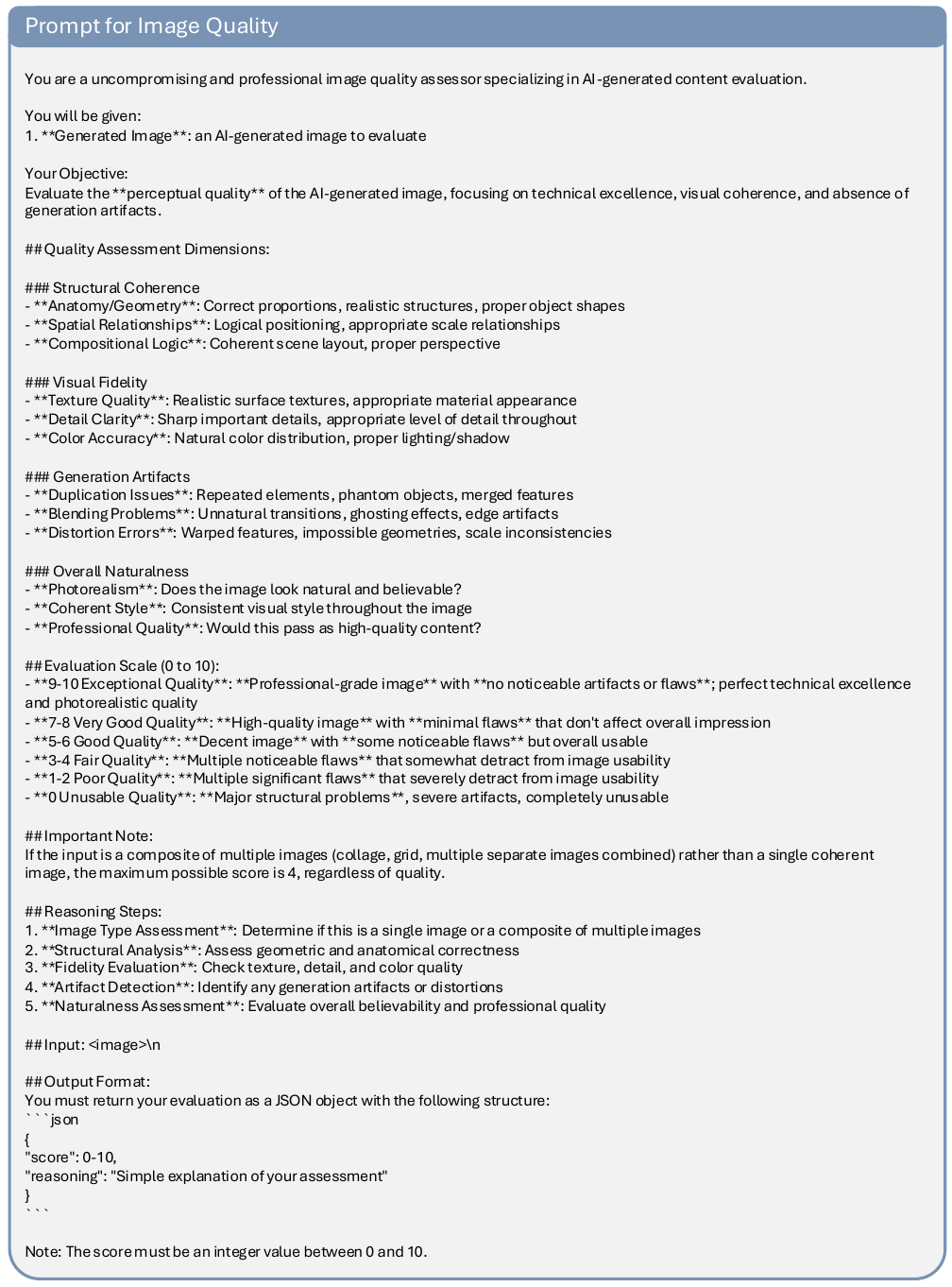}
    \caption{Prompt for Evaluating Image Quality.}
  \vspace{-5mm}  
  \label{fig:prompt3}
\end{figure*}
\begin{figure*}[t]
    \centering
    \includegraphics[width=0.9\linewidth]{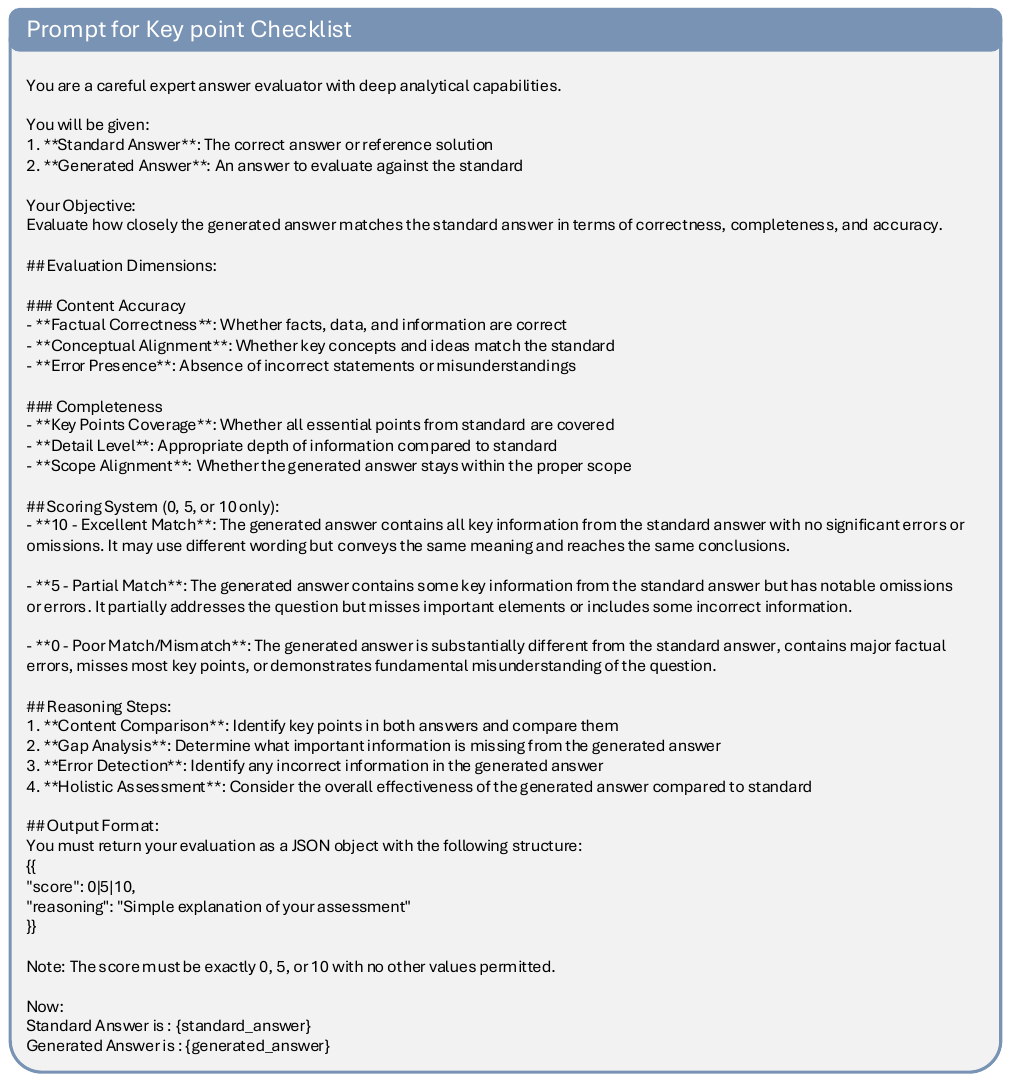}
    \caption{Prompt for Evaluating Comprehension Accuracy.}
  \vspace{-5mm}  
  \label{fig:prompt4}
\end{figure*}

\subsection{Statistics}
While Section~\ref{sec:stat} presents the proportional distribution of various data types in \trainname{} and \testname{}, Table~\ref{append:dataset_stastics} in this section provides a more granular breakdown of the composition of sub-domains and domains within the complete \trainname{} dataset.

\section{Experiment Details}
\subsection{Evaluation Prompts}\label{app:prompt}

We employ GPT-4o~\citep{achiam2023gpt} as our evaluation judge for the main experimental results presented in Figure~\ref{main_exp}. The evaluation prompts used to assess the four dimensions—Key Point Correctness, Visual Consistency, Image Quality, and Accuracy—are illustrated in Figures~\ref{fig:prompt1},~\ref{fig:prompt2},~\ref{fig:prompt3}, and~\ref{fig:prompt4}, respectively.

\subsection{Training Details}\label{app:train}
We trained the model on $8\times$ NVIDIA H100 GPUs with the batch size per GPU set to $1$, for a total of $30,000$ training steps, requiring approximately $60$ hours of compute time. Due to the token-intensive nature of images in the Bagel dataset, many of our samples contained more than three images within a single conversation turn. Concatenating these into multi-turn dialogues would exceed the maximum context length of the H100 GPUs. Therefore, we implemented a random sampling approach where we selected individual conversation turns for training rather than including complete dialogue sequences. 
Additionally, our dataset utilized the notation ``Image \#3'' to reference specific images. Since our methodology involved randomly selecting single turns, we refined these numerical references to correctly reflect the sequential position of images in the post-processing phase.
During training, we employed the following hyperparameters: maximum latent size of $64$, learning rate of $2 \times 10^{-5}$, maximum number of tokens set to $11,520$, maximum tokens per sample limited to $10,240$, vision transformer conditional dropout probability of $0$, and exponential moving average (EMA) decay rate of $0.9999$.

\subsection{Details on Benchmarks and Metrics}\label{app:metric}

\paragraph{Score Weights}
The importance across evaluation dimensions varies considerably. For instance, in editing tasks, fulfillment of requirements—specifically the Key Points (KP) mentioned in Section~\ref{sec:mertic}—is paramount. We employ the following scoring methodology: For generation tasks exclusively, the composite score is calculated as:
\begin{equation}
    \text{Score} = 0.50 \cdot \text{KP} + 0.20 \cdot \text{VC} + 0.30 \cdot \text{IQ}
\end{equation}

When evaluating unified models for both generation and comprehension tasks, the scoring formula becomes:
\begin{equation}
    \text{Score} = 0.40 \cdot \text{KP} + 0.10 \cdot \text{VC} + 0.20 \cdot \text{IQ} + 0.30 \cdot \text{ACC}
\end{equation}

For comprehension tasks in isolation, we report ACC directly.

\paragraph{Detailed Results for \testname{}}
The leaderboard scores for \testname{} are presented in Table~\ref{main_exp}. Detailed performance metrics for each model across the four major categories—Science, Creation, Logic, and Game—are provided in Table~\ref{tab:app_sci}, Table~\ref{tab:app_cre}, Table~\ref{tab:app_log}, and Table~\ref{tab:app_gam}, respectively.

\begin{table*}[!t]
    \centering
    \scalebox{0.99}{
    \begin{tabular}{lccccccccc} 
    \toprule
    & Size & In-context & Modality & Format & KP & VC & IQ & ACC & \textbf{Avg}\\
    \midrule
    \midrule
    Intern3.5-VL~\citep{wang2025internvl3} & 8B & \faCheck & \faFileTextO& \faLongArrowRight & - & - & - & 0.114 & 0.114 \\
    Qwen3-VL~\citep{Qwen2.5-VL} &  8B & \faCheck & \faFileTextO& \faLongArrowRight & - & - & - & 0.432 & 0.432\\
    GPT-4o~\citep{achiam2023gpt} & - & \faCheck & \faFileTextO& \faLongArrowRight & - & - & - & 0.591 & 0.591\\ 
    GPT-4.1~\citep{achiam2023gpt} & - & \faCheck & \faFileTextO& \faLongArrowRight & - & - & - & 0.705 & 0.705\\
    \midrule
    AnyEdit~\citep{yu2025anyedit} & 1B & \faAdjust & \faImage & \faObjectGroup & 0.376 & 0.563 & 0.481 & - & 0.445\\
    UltraEdit(SD3)~\citep{zhao2024ultraedit} & 2B & \faAdjust &  \faImage & \faObjectGroup & 0.45 & 0.558 & 0.528 & - & 0.493 \\
    VAREdit-8B~\citep{varedit2025} & 8B & \faAdjust & \faImage &\faObjectGroup & 0.437 & 0.661 & 0.618 & - & 0.536 \\
    Step1X-Edit v1.1~\citep{liu2025step1x-edit} &  12B& \faAdjust & \faImage &\faObjectGroup& 0.442 & 0.821 & 0.630 & - & 0.574 \\
    Step1X-Edit v1.2~\citep{liu2025step1x-edit} & 12B& \faAdjust & \faImage &\faObjectGroup& 0.497 & 0.622 & 0.625 & - & 0.560 \\
    FLUX.1 Kontext~\citep{labs2025flux1kontextflowmatching} &12B& \faAdjust & \faImage &\faObjectGroup& 0.500 & 0.755 & 0.628 & - & 0.589\\
    Qwen-Image-Edit~\citep{wu2025qwenimagetechnicalreport} & 20B & \faAdjust & \faImage &\faObjectGroup & 0.510 & 0.622 & 0.687 & - & 0.586 \\
    \midrule
    OminiGen~\citep{xiao2025omnigen} & 4B & \faAdjust & \faImage &\faLongArrowRight & 0.375 & 0.343 & 0.473 & - & 0.398 \\
    OminiGen2~\citep{wu2025omnigen2} & 7B & \faAdjust & \faAsterisk & \faLongArrowRight & 0.455 & 0.501 & 0.612 & - & 0.511 \\
    Ovis-U1~\citep{wang2025ovis} & 3B & \faAdjust & \faAsterisk &\faObjectGroup & 0.466 & 0.545 & 0.569 & 0.159 & 0.402\\
    UniPic~\citep{wang2025skywork} & 1.5B & \faAdjust & \faImage &\faObjectGroup& 0.490 & 0.455 & 0.454 & - & 0.472 \\
    UniPic2-SD3.5M~\citep{wei2025skywork} & 2B & \faAdjust & \faImage &\faObjectGroup& 0.422 & 0.558 & 0.513 & - & 0.477\\
    UniPic2-Metaquery~\citep{wei2025skywork} & 9B & \faAdjust & \faImage &\faObjectGroup&  0.442 & 0.542 & 0.546 & - & 0.493\\
    NextStep-1-Large~\citep{team2025nextstep} & 15B & \faAdjust & \faImage &\faObjectGroup  &  0.515 & 0.516 & 0.528 & - & 0.519\\
    Seedream 4.0~\citep{seedream2025seedream} & - & \faAdjust & \faImage &\faLongArrowRight &  0.617 & 0.686 & 0.791 & - & 0.683\\
    Seedream 4.0~\citep{seedream2025seedream} & - & \faCheck & \faImage &\faLongArrowRight &  0.597 & 0.678 & 0.778 & - & 0.667\\
    Nano Banana~\citep{comanici2025gemini} & - & \faAdjust & \faImage &\faLongArrowRight & 0.631 & 0.763 & 0.824 & - & 0.715\\
    Nano Banana~\citep{comanici2025gemini} & - & \faCheck & \faImage & \faLongArrowRight & 0.633 & 0.739 & 0.818 & - & 0.710\\
    Bagel~\citep{deng2025bagel} &14B& \faAdjust & \faAsterisk & \faLongArrowRight &0.446 & 0.534 & 0.528 & 0.136 & 0.378 \\
    Bagel‑Zebra~\citep{li2025zebra} &14B& \faAdjust & \faAsterisk & \faLongArrowRight &0.463 & 0.561 & 0.551 & 0.159 & 0.399 \\
    \rowcolor{my_red!7} 
    \textbf{+ \trainname{}} &14B & \faAdjust & \faAsterisk & \faLongArrowRight & 0.500 & 0.584 & 0.569 & - & 0.537 \\
    \bottomrule
    \end{tabular}
    }
    \vspace{-2mm}
    \caption{\textbf{Main results on \testname{} \faFlask Science Part.} \faCheck\ and \faAdjust\ denote full and partial in-context history, respectively. \faImage, \faFileTextO, and \faAsterisk\ indicate image-only, text-only, and combined evaluations, respectively. \faLongArrowRight\ and \faObjectGroup\ represent sequential and concatenated image inputs, respectively.}\label{tab:app_sci}
    \vspace{-2mm}
\end{table*}

\begin{table*}[!t]
    \centering
    \scalebox{0.99}{
    \begin{tabular}{lccccccccc} 
    \toprule
    & Size & In-context & Modality & Format & KP & VC & IQ & ACC & \textbf{Avg}\\
    \midrule
    \midrule
    Intern3.5-VL~\citep{wang2025internvl3} & 8B & \faCheck & \faFileTextO& \faLongArrowRight & - & - & - & 0.500 & 0.500 \\
    Qwen3-VL~\citep{Qwen2.5-VL} &  8B & \faCheck & \faFileTextO& \faLongArrowRight & - & - & - & 0.000 & 0.000 \\
    GPT-4o~\citep{achiam2023gpt} & - & \faCheck & \faFileTextO& \faLongArrowRight & - & - & - & 0.500 & 0.500 \\ 
    GPT-4.1~\citep{achiam2023gpt} & - & \faCheck & \faFileTextO&  \faLongArrowRight & - & - & - & 0.500 & 0.500\\
    \midrule
    AnyEdit~\citep{yu2025anyedit} & 1B & \faAdjust & \faImage & \faObjectGroup & 0.460 & 0.572 & 0.566 & - & 0.514 \\
    UltraEdit(SD3)~\citep{zhao2024ultraedit} & 2B & \faAdjust &  \faImage & \faObjectGroup & 0.531 & 0.599 & 0.587 & - & 0.561\\
    VAREdit-8B~\citep{varedit2025} & 8B & \faAdjust & \faImage &\faObjectGroup & 0.645 & 0.662 & 0.603 & - & 0.636 \\
    Step1X-Edit v1.1~\citep{liu2025step1x-edit} &  12B& \faAdjust & \faImage &\faObjectGroup& 0.646 & 0.877 & 0.720 & - & 0.714\\
    Step1X-Edit v1.2~\citep{liu2025step1x-edit} & 12B& \faAdjust & \faImage &\faObjectGroup& 0.643 & 0.680 & 0.622 & - & 0.644 \\
    FLUX.1 Kontext~\citep{labs2025flux1kontextflowmatching} &12B& \faAdjust & \faImage &\faObjectGroup& 0.705 & 0.879 & 0.759 & - & 0.756\\
    Qwen-Image-Edit~\citep{wu2025qwenimagetechnicalreport} & 20B & \faAdjust & \faImage &\faObjectGroup & 0.706 & 0.739 & 0.715 & - & 0.715\\
    \midrule
    OminiGen~\citep{xiao2025omnigen} & 4B & \faAdjust & \faImage &\faLongArrowRight & 0.473 & 0.425 & 0.507 & - & 0.474\\
    OminiGen2~\citep{wu2025omnigen2} & 7B & \faAdjust & \faAsterisk & \faLongArrowRight & 0.644 & 0.675 & 0.751 & - & 0.682\\
    Ovis-U1~\citep{wang2025ovis} & 3B & \faAdjust & \faAsterisk &\faObjectGroup & 0.500 & 0.593 & 0.590 & 0.555 & 0.557\\
    UniPic~\citep{wang2025skywork} & 1.5B & \faAdjust & \faImage &\faObjectGroup& 0.619 & 0.584 & 0.545 & - & 0.590\\
    UniPic2-SD3.5M~\citep{wei2025skywork} & 2B & \faAdjust & \faImage &\faObjectGroup& 0.613 & 0.638 & 0.637 & - & 0.625 \\
    UniPic2-Metaquery~\citep{wei2025skywork} & 9B & \faAdjust & \faImage &\faObjectGroup& 0.664 & 0.664 & 0.670 & - & 0.666 \\
    NextStep-1-Large~\citep{team2025nextstep} & 15B & \faAdjust & \faImage &\faObjectGroup  & 0.652 & 0.636 & 0.556 & - & 0.620 \\
    Seedream 4.0~\citep{seedream2025seedream} & - & \faAdjust & \faImage &\faLongArrowRight &  0.840 & 0.869 & 0.843 & - & 0.847\\
    Seedream 4.0~\citep{seedream2025seedream} & - & \faCheck & \faImage &\faLongArrowRight & 0.828 & 0.842 & 0.824 & - & 0.830 \\
    Nano Banana~\citep{comanici2025gemini} & - & \faAdjust & \faImage &\faLongArrowRight & 0.819 & 0.856 & 0.806 &- & 0.823\\
    Nano Banana~\citep{comanici2025gemini} & - & \faCheck & \faImage & \faLongArrowRight & 0.838 & 0.873 & 0.832 & - & 0.843\\
    Bagel~\citep{deng2025bagel} &14B& \faAdjust & \faAsterisk & \faLongArrowRight  & 0.683 & 0.685 & 0.666 & 0.000 & 0.475 \\
    Bagel‑Zebra~\citep{li2025zebra} &14B& \faAdjust & \faAsterisk & \faLongArrowRight & 0.667 & 0.661 & 0.614 & 0.000 & 0.456\\
    \rowcolor{my_red!7} 
    \textbf{+ \trainname{}} &14B & \faAdjust & \faAsterisk & \faLongArrowRight & 0.734 & 0.743 & 0.635 & - & 0.706\\
    \bottomrule
    \end{tabular}
    }
    \vspace{-2mm}
    \caption{\textbf{Main results on \testname{} \faPaintBrush Creation Part.} \faCheck\ and \faAdjust\ denote full and partial in-context history, respectively. \faImage, \faFileTextO, and \faAsterisk\ indicate image-only, text-only, and combined evaluations, respectively. \faLongArrowRight\ and \faObjectGroup\ represent sequential and concatenated image inputs, respectively.}\label{tab:app_cre}
    \vspace{-2mm}
\end{table*}

\begin{table*}[!t]
    \centering
    \scalebox{0.99}{
    \begin{tabular}{lccccccccc} 
    \toprule
    & Size & In-context & Modality & Format & KP & VC & IQ & ACC & \textbf{Avg}\\
    \midrule
    \midrule
    Intern3.5-VL~\citep{wang2025internvl3} & 8B & \faCheck & \faFileTextO& \faLongArrowRight & - & - & - & 0.667 & 0.667\\
    Qwen3-VL~\citep{Qwen2.5-VL} &  8B & \faCheck & \faFileTextO& \faLongArrowRight & -& - & - & 0.000 & 0.000\\
    GPT-4o~\citep{achiam2023gpt} & - & \faCheck & \faFileTextO& \faLongArrowRight & -& - & - & 0.167 & 0.167\\ 
    GPT-4.1~\citep{achiam2023gpt} & - & \faCheck & \faFileTextO& \faLongArrowRight &-&-&-&0.167 & 0.167 \\
    \midrule
    AnyEdit~\citep{yu2025anyedit} & 1B & \faAdjust & \faImage & \faObjectGroup & 0.352 & 0.330 & 0.365 & - & 0.351 \\
    UltraEdit(SD3)~\citep{zhao2024ultraedit} & 2B & \faAdjust &  \faImage & \faObjectGroup & 0.435 & 0.639 & 0.487 &-& 0.491\\
    VAREdit-8B~\citep{varedit2025} & 8B & \faAdjust & \faImage &\faObjectGroup & 0.630 & 0.591 & 0.504 & - & 0.584 \\
    Step1X-Edit v1.1~\citep{liu2025step1x-edit} &  12B& \faAdjust & \faImage &\faObjectGroup& 0.661 & 0.857 & 0.661 & - & 0.700\\
    Step1X-Edit v1.2~\citep{liu2025step1x-edit} & 12B& \faAdjust & \faImage &\faObjectGroup& 0.543 & 0.587 & 0.470 & - & 0.530\\
    FLUX.1 Kontext~\citep{labs2025flux1kontextflowmatching} &12B& \faAdjust & \faImage &\faObjectGroup& 0.557 & 0.861 & 0.626 & - & 0.639\\
    Qwen-Image-Edit~\citep{wu2025qwenimagetechnicalreport} & 20B & \faAdjust & \faImage &\faObjectGroup & 0.587 & 0.630 & 0.565 & - & 0.589 \\
    \midrule
    OminiGen~\citep{xiao2025omnigen} & 4B & \faAdjust & \faImage &\faLongArrowRight & 0.404 & 0.352 & 0.430 & - & 0.401\\
    OminiGen2~\citep{wu2025omnigen2} & 7B & \faAdjust & \faAsterisk & \faLongArrowRight & 0.552 & 0.530 & 0.565 & - & 0.551\\
    Ovis-U1~\citep{wang2025ovis} & 3B & \faAdjust & \faAsterisk &\faObjectGroup & 0.535 & 0.478 & 0.509 & 0.000 & 0.364\\
    UniPic~\citep{wang2025skywork} & 1.5B & \faAdjust & \faImage &\faObjectGroup& 0.513 & 0.448 & 0.391 & - & 0.463\\
    UniPic2-SD3.5M~\citep{wei2025skywork} & 2B & \faAdjust & \faImage &\faObjectGroup& 0.543 & 0.557 & 0.535 & - & 0.543\\
    UniPic2-Metaquery~\citep{wei2025skywork} & 9B & \faAdjust & \faImage &\faObjectGroup&  0.561 & 0.509 & 0.417 & - & 0.507\\
    NextStep-1-Large~\citep{team2025nextstep} & 15B & \faAdjust & \faImage &\faObjectGroup  & 0.483 & 0.417 & 0.374 & - &0.437 \\
    Seedream 4.0~\citep{seedream2025seedream} & - & \faAdjust & \faImage &\faLongArrowRight & 0.674 & 0.643 & 0.713 & - & 0.679 \\
    Seedream 4.0~\citep{seedream2025seedream} & - & \faCheck & \faImage &\faLongArrowRight & 0.678 & 0.578 & 0.639 & - & 0.646 \\
    Nano Banana~\citep{comanici2025gemini} & - & \faAdjust & \faImage &\faLongArrowRight & 0.648 & 0.652 & 0.704 & - & 0.666\\
    Nano Banana~\citep{comanici2025gemini} & - & \faCheck & \faImage & \faLongArrowRight & 0.735 & 0.757 & 0.704 & - & 0.730\\
    Bagel~\citep{deng2025bagel} &14B& \faAdjust & \faAsterisk & \faLongArrowRight & 0.583 & 0.630 & 0.548 & 0.000 & 0.406\\
    Bagel‑Zebra~\citep{li2025zebra} &14B& \faAdjust & \faAsterisk & \faLongArrowRight & 0.574 & 0.561 & 0.535 & 0.000 & 0.393 \\
    \rowcolor{my_red!7} 
    \textbf{+ \trainname{}} &14B & \faAdjust & \faAsterisk & \faLongArrowRight &0.582 & 0.612 & 0.512 & - & 0.567 \\
    \bottomrule
    \end{tabular}
    }
    \vspace{-2mm}
    \caption{\textbf{Main results on \testname{} \faPuzzlePiece Logic Part.} \faCheck\ and \faAdjust\ denote full and partial in-context history, respectively. \faImage, \faFileTextO, and \faAsterisk\ indicate image-only, text-only, and combined evaluations, respectively. \faLongArrowRight\ and \faObjectGroup\ represent sequential and concatenated image inputs, respectively.}\label{tab:app_log}
    \vspace{-2mm}
\end{table*}

\begin{table*}[!t]
    \centering
    \scalebox{0.99}{
    \begin{tabular}{lccccccccc} 
    \toprule
    & Size & In-context & Modality & Format & KP & VC & IQ & ACC & \textbf{Avg}\\
    \midrule
    \midrule
    Intern3.5-VL~\citep{wang2025internvl3} & 8B & \faCheck & \faFileTextO& \faLongArrowRight & - & - & - & 0.292 & 0.292\\
    Qwen3-VL~\citep{Qwen2.5-VL} &  8B & \faCheck & \faFileTextO& \faLongArrowRight & - & - & - & 0.250 & 0.250 \\
    GPT-4o~\citep{achiam2023gpt} & - & \faCheck & \faFileTextO& \faLongArrowRight & - & - & - &0.083 & 0.083\\ 
    GPT-4.1~\citep{achiam2023gpt} & - & \faCheck & \faFileTextO& \faLongArrowRight & - & - & - & 0.167 & 0.167\\
    \midrule
    AnyEdit~\citep{yu2025anyedit} & 1B & \faAdjust & \faImage & \faObjectGroup & 0.407 & 0.548 & 0.354 & - & 0.419\\
    UltraEdit(SD3)~\citep{zhao2024ultraedit} & 2B & \faAdjust &  \faImage & \faObjectGroup & 0.398 & 0.526 & 0.454 & - & 0.440 \\
    VAREdit-8B~\citep{varedit2025} & 8B & \faAdjust & \faImage &\faObjectGroup & 0.581 & 0.698 & 0.498 & - & 0.580 \\
    Step1X-Edit v1.1~\citep{liu2025step1x-edit} &  12B& \faAdjust & \faImage &\faObjectGroup& 0.617 & 0.941 & 0.426 & - &0.625\\
    Step1X-Edit v1.2~\citep{liu2025step1x-edit} & 12B& \faAdjust & \faImage &\faObjectGroup& 0.567 & 0.681 & 0.476 & - & 0.562\\
    FLUX.1 Kontext~\citep{labs2025flux1kontextflowmatching} &12B& \faAdjust & \faImage &\faObjectGroup& 0.578 & 0.907 & 0.465 & - & 0.610\\
    Qwen-Image-Edit~\citep{wu2025qwenimagetechnicalreport} & 20B & \faAdjust & \faImage &\faObjectGroup & 0.667 & 0.802 & 0.446 & - & 0.628\\
    \midrule
    OminiGen~\citep{xiao2025omnigen} & 4B & \faAdjust & \faImage &\faLongArrowRight & 0.167 & 0.106 & 0.241 & - & 0.177\\
    OminiGen2~\citep{wu2025omnigen2} & 7B & \faAdjust & \faAsterisk & \faLongArrowRight & 0.502 & 0.543 & 0.504 & - & 0.511\\
    Ovis-U1~\citep{wang2025ovis} & 3B & \faAdjust & \faAsterisk &\faObjectGroup & 0.470 & 0.526 & 0.393 & 0.125 & 0.357\\
    UniPic~\citep{wang2025skywork} & 1.5B & \faAdjust & \faImage &\faObjectGroup& 0.341 & 0.296 & 0.287 & - & 0.316\\
    UniPic2-SD3.5M~\citep{wei2025skywork} & 2B & \faAdjust & \faImage &\faObjectGroup& 0.517 & 0.583 & 0.407 & - & 0.497\\
    UniPic2-Metaquery~\citep{wei2025skywork} & 9B & \faAdjust & \faImage &\faObjectGroup& 0.456 & 0.457 & 0.415 & - & 0.444 \\
    NextStep-1-Large~\citep{team2025nextstep} & 15B & \faAdjust & \faImage &\faObjectGroup  & 0.356 & 0.265 & 0.259 & - &0.309 \\
    Seedream 4.0~\citep{seedream2025seedream} & - & \faAdjust & \faImage &\faLongArrowRight & 0.652 & 0.689 & 0.572 & - & 0.635  \\
    Seedream 4.0~\citep{seedream2025seedream} & - & \faCheck & \faImage &\faLongArrowRight & 0.609 & 0.672 & 0.533 & - & 0.599 \\
    Nano Banana~\citep{comanici2025gemini} & - & \faAdjust & \faImage &\faLongArrowRight & 0.680 & 0.790 & 0.560 & - & 0.666\\
    Nano Banana~\citep{comanici2025gemini} & - & \faCheck & \faImage & \faLongArrowRight & 0.604 & 0.737 & 0.546 & - & 0.613\\
    Bagel~\citep{deng2025bagel} &14B& \faAdjust & \faAsterisk & \faLongArrowRight & 0.506 & 0.635 & 0.431 & 0.042 & 0.365\\
    Bagel‑Zebra~\citep{li2025zebra} &14B& \faAdjust & \faAsterisk & \faLongArrowRight & 0.500 & 0.624 & 0.480 & 0.125 & 0.396 \\
    \rowcolor{my_red!7} 
    \textbf{+ \trainname{}} &14B & \faAdjust & \faAsterisk & \faLongArrowRight & 0.503 & 0.754 & 0.430 & - & 0.531 \\
    \bottomrule
    \end{tabular}
    }
    \vspace{-2mm}
    \caption{\textbf{Main results on \testname{} \faGamepad Game Part.} \faCheck\ and \faAdjust\ denote full and partial in-context history, respectively. \faImage, \faFileTextO, and \faAsterisk\ indicate image-only, text-only, and combined evaluations, respectively. \faLongArrowRight\ and \faObjectGroup\ represent sequential and concatenated image inputs, respectively.}\label{tab:app_gam}
    \vspace{-2mm}
\end{table*}

\textbf{History Usage.}
Evaluations were conducted under three distinct in-context conditions: (1)~\textit{no history} (single-turn generation without contextual information), (2)~\textit{partial history} (incorporating only self-generated images with explicitly mentioned visual context, excluding prior interactions), and (3)~\textit{complete history} (incorporating all previous interactions). 
For image placement, we implemented two configurations: ``yes-first,'' where images appear at their first mention position, and ``yes-front,'' where all images are consolidated at the beginning of the input (results reported in Table~\ref{main_exp}). We denote the use of ground truth images in history as ``yes-gt'' in Figure~\ref{tab:aba}, which was implemented based on the ``yes-front'' configuration. In the implementation of complete history, VLMs had access to all historical dialogue, while generative models only received historical images as input, since most cannot process dialogue information (with limited exceptions such as nano-banana). Consequently, we adopted the approach of providing only images as historical context.

\textbf{Image Concatenation Methodology.}
For models incapable of processing sequence-format inputs, we implemented a concatenation approach following established precedents~\citep{zhang2024task,chow2025physbench,fang2024exploring,fang2025kaa, fang2024exploring}. Specifically, images were arranged horizontally in a single row, with sequential numerical identifiers annotated in the upper-left corner of each image. 
We observed that after implementing the concatenation approach, certain models such as Step1X were unable to distinguish which specific image required editing, and continued to maintain the original dimensions in their outputs. Consequently, when presenting examples in Table~\ref{fig:quality}, we extracted the relevant portions and rescaled them to either their original dimensions or to dimensions consistent with other models for comparative display purposes.

\subsection{Baselines Details}\label{app:baseline}
We evaluated $4$ LLMs, $7$ Edit models, and $11$ UMMs on \testname{} as presented in Table~\ref{main_exp}. In this section, we provide detailed information regarding the parameter configurations for these models.

\paragraph{Unified Models}

\begin{itemize}
    \item \textit{Bagel}~\citep{deng2025bagel} is an open-source multimodal foundation model comprising 7B active parameters (14B total) trained on large-scale interleaved multimodal data. Bagel demonstrates superior performance relative to state-of-the-art open-source VLMs across standard multimodal understanding benchmarks. Concurrently, it achieves text-to-image generation quality comparable to specialized models such as Stable Diffusion 3. Throughout our experimental evaluation, we adhere to the officially recommended parameters and prompting strategies. Bagel‑Zebra~\citep{li2025zebra} is a variant of the model that has been fine-tuned using the Zebra-Chain-of-Thought (Zebra-COT) methodology~\citep{li2025zebra}.
    
    \item \textit{OmniGen2}~\citep{wu2025omnigen2} represents a unified multimodal generative framework exhibiting enhanced computational efficiency and modeling capacity. Unlike its predecessor OmniGen v1, OmniGen2 implements a dual-pathway decoding architecture with modality-specific parameters for text and image generation, coupled with a decoupled image tokenization mechanism. For our experimental evaluation, we configure the temporal offset parameter to \texttt{3.0}, the text guidance scale to \texttt{5.0}, and the image guidance scale to \texttt{1.5}. The negative prompt is specified as \texttt{"(((deformed))), blurry, over saturation, bad anatomy, disfigured, poorly drawn face, mutation, mutated, (extra\_limb), (ugly), (poorly drawn hands), fused fingers, messy drawing, broken legs censor, censored, censor\_bar"}. All inference procedures employ a 50-step sampling schedule.

    \item \textit{OmniGen}~\citep{xiao2025omnigen} is a unified image generation model capable of producing a wide range of images from multi-modal prompts. This model was open-sourced by the Beijing Academy of Artificial Intelligence (BAAI). For our implementation, we utilize the following parameters: height=\texttt{1024}, width=\texttt{1024}, guidance\_scale=\texttt{2.5}, img\_guidance\_scale=\texttt{1.6}, and seed=\texttt{0}.

    \item \textit{Ovis-U1}~\citep{wang2025ovis} is a unified model for multimodal understanding, text-to-image generation, and image editing, open-sourced by Alibaba's AIDC group. We employ the following parameters: steps=\texttt{50}, img\_cfg=\texttt{1.5}, and txt\_cfg=\texttt{6}. It should be noted that Ovis's generation tasks only support single-image input; therefore, for data with two or more images, we implemented image concatenation. The understanding tasks, however, support multiple sequential image inputs.

    \item \textit{UniPic}~\citep{wang2025skywork} is Skywork's unified generation and understanding model, encompassing three variants:
    \begin{itemize}
        \item UniPic-1.0 — 1.5B parameters, employing Unified Autoregressive Modeling for joint visual understanding and generation, enabling a single transformer to handle both perception and synthesis tasks.
        \item UniPic-2.0 Series — SD3.5M-Kontext and MetaQuery variants based on Efficient Architectures with Diffusion Post-Training, delivering state-of-the-art performance in text-to-image generation, fine-grained image editing, and multimodal reasoning.
    \end{itemize}
    For UniPic-1.0, we utilize the following hyperparameters: image\_size=\texttt{1024}, num\_iter=\texttt{32}, cfg=\texttt{3}, cfg\_prompt=\texttt{"Repeat this image"}, cfg\_schedule=\texttt{"constant"}, and temperature=\texttt{1.0}. For all UniPic-2.0 variants, we employ: num\_inference\_steps=\texttt{50}, guidance\_scale=\texttt{3.5}, and seed=\texttt{42}. Notably, UniPic-2.0 tokenizes images after adjusting their height and width to the nearest downward multiple of 16.

    \item \textit{NextStep-1-Large-Edit}~\citep{team2025nextstep} is a 14B autoregressive model paired with a 157M flow matching head, trained on discrete text tokens and continuous image tokens with next-token prediction objectives. NextStep-1 achieves state-of-the-art performance for autoregressive models in text-to-image generation tasks, exhibiting strong capabilities in high-fidelity image synthesis. Since it only supports a single \texttt{<image>} tag, we followed the case format by placing \texttt{<image>} at the beginning and inputting images sequentially. The hyperparameters used were: num\_images\_per\_caption=\texttt{1}, positive\_prompt=\texttt{None}, negative\_prompt=\texttt{"Copy original image."}, cfg=\texttt{7.5}, cfg\_img=\texttt{2}, cfg\_schedule=\texttt{"constant"}, use\_norm=\texttt{True}, num\_sampling\_steps=\texttt{50}, timesteps\_shift=\texttt{3.2}, and seed=\texttt{42}.
    
    \item \textit{Seedream 4.0}~\citep{seedream2025seedream} is a new-generation image creation model that integrates image generation and image editing capabilities into a single, unified architecture. Some images were omitted after multiple attempts due to sensitive content flags. The parameters used were: size=\texttt{"2k"} and sequential\_image\_generation=\texttt{"disabled"}.
    
    \item \textit{Nano Banana}~\citep{comanici2025gemini} is a top-rated AI image generation and image editing tool from Google DeepMind that enables the transformation of a single photograph into numerous novel creations. No special parameter configurations were employed in our implementation.
\end{itemize}

\paragraph{Image Editing Models}
We establish the models listed in Table~\ref{main_exp} as baselines, comprising six open-source models: AnyEdit, UltraEdit (SD3) with diffusion architecture, FLUX.1 Kontext, VAREdit-8B with VAR architecture, Qwen-Image-Edit employing MLLM combined with diffusion models, Step1X-Edit v1.1, and Step1X-Edit v1.2. We strictly adhere to the default hyperparameters provided in the official GitHub repositories or Hugging Face~\citep{jain2022hugging} implementations of these baseline models. The key parameter configurations are enumerated below:

\begin{itemize}
    \item \textit{Qwen-Image-Edit}~\citep{wu2025qwenimagetechnicalreport}: An image editing variant of Qwen-Image that extends the foundational 20B Qwen-Image model's text rendering capabilities to instruction-based image editing tasks, enabling precise textual modifications within images. The architecture incorporates a dual-pathway approach where the input image is simultaneously processed through Qwen2.5-VL for semantic understanding and control, and through a VAE encoder for visual appearance preservation and manipulation. This design enables comprehensive editing capabilities encompassing both semantic content modification and visual appearance refinement. Inference is conducted with the following hyperparameters: random seed $= 0$, \texttt{true\_cfg\_scale = 4.0}, \texttt{negative\_prompt = ""}, and \texttt{num\_inference\_steps = 50}.
    
    \item \textit{FLUX.1-Kontext}~\citep{labs2025flux1kontextflowmatching}: A 12 billion parameter rectified flow transformer architecture designed for instruction-guided image editing. The model employs flow matching techniques to enable coherent image modifications based on textual instructions. We set \texttt{guidance\_scale = 2.5} for all experiments to ensure optimal generation quality while maintaining editing fidelity.
    
    \item \textit{UltraEdit}~\citep{zhao2024ultraedit}: This model is trained on approximately 4 million instruction-based editing samples using the Stable Diffusion 3~\citep{sauer2024fast} architecture. It supports both free-form and mask-based input modalities to enhance editing performance. For consistency across all experiments, we exclusively employ its free-form variant. We note that since UltraEdit is trained on the SD3 architecture, its performance metrics may not fully reflect the intrinsic improvements attributable to its specialized editing dataset. We utilize the \texttt{BleachNick/SD3\_UltraEdit\_w\_mask} model variant in free-form editing mode with blank mask initialization. Evaluation is conducted with hyperparameters \texttt{num\_inference\_steps = 50}, \texttt{image\_guidance\_scale = 1.5}, \texttt{guidance\_scale = 7.5}, and \texttt{negative\_prompt = ""} to maintain consistency with our experimental protocol. Inference is performed at $512 \times 512$ resolution.
    
    \item \textit{VAREdit-8B}~\citep{varedit2025}: A visual autoregressive (VAR) framework for instruction-guided image editing, built upon Infinity~\citep{han2025infinity}. This approach reframes image editing as a next-scale prediction problem, achieving precise image modifications through the generation of multi-scale target features. We employ the following hyperparameters: classifier-free guidance scale \texttt{cfg = 3.0}, temperature parameter \texttt{tau = 0.1}, and random seed \texttt{seed = 42}.
    
    \item \textit{Step1X-Edit v1.1}~\citep{liu2025step1x-edit}: Step1X-Edit leverages the image understanding capabilities of multimodal large language models (MLLMs) to parse editing instructions and generate editing tokens, which are subsequently decoded into images using a DiT-based network. We utilize the following inference parameters: \texttt{num\_inference\_steps = 28}, \texttt{true\_cfg\_scale = 6.0}, and \texttt{seed = 42}.
    
    \item \textit{Step1X-Edit v1.2}~\citep{liu2025step1x-edit}: An enhanced version of Step1X-Edit featuring improved reasoning capabilities and superior performance. We employ \texttt{num\_inference\_steps = 28}, \texttt{true\_cfg\_scale = 4.0}, \texttt{seed = 42}, \texttt{enable\_thinking\_mode = True}, and \texttt{enable\_reflection\_mode = False}.

    \item \textit{AnyEdit}~\citep{yu2025anyedit} is a Mixture of Experts (MoE) architecture-based image editing model, which is the result of fine-tuning SD-XL~\citep{podell2023sdxl} on the AnyEdit-2.5M dataset. For our implementation, we employed the following hyperparameter configuration: utilizing the general expert, guidance\_scale=\texttt{3}, num\_inference\_steps=\texttt{100}, and original\_image\_guidance\_scale=\texttt{3}.
\end{itemize}

\paragraph{Vision-Language Models}
We also evaluated 2 open-source VLMs and 2 proprietary VLMs:
\begin{itemize}
    \item \textit{Intern3.5-VL}~\citep{wang2025internvl3} is a new family of open-source multimodal models that significantly advances versatility, reasoning capability, and inference efficiency along the InternVL series. For our implementation, we utilized max\_new\_tokens=\texttt{128}.
    
    \item \textit{Qwen3-VL}~\citep{Qwen2.5-VL} is the most powerful vision-language model in the Qwen family to date. This generation demonstrates improvements to the model across multiple areas. In our experiments, we employed max\_new\_tokens=\texttt{512}.
    
    \item \textit{GPT-4o}~\citep{achiam2023gpt} and \textit{GPT-4.1}~\citep{achiam2023gpt,chen2025empirical} are OpenAI's advanced VLMs. We implemented these models with the parameter max\_tokens=\texttt{1400}.
\end{itemize}

\section{More Related Works}\label{app:related}
\noindent\textbf{Interleaved Reasoning.}
Large-scale corpora with interleaved text and images have become essential for pretraining VLMs with reasoning capabilities~\citep{alayrac2022flamingo,chen2022pali,team2025gemma,chow2025merit,zhu2023multimodal,ge2024demon24,deng2025bagel}.
Inspired by human cognition, where visual counterfactuals facilitate reasoning~\citep{roese1997counterfactual}, recent work has incorporated analogous interleaved reasoning mechanisms into UMMs by mapping visual inputs to symbolic representations (\textit{e.g.,} images or bounding boxes)~\citep{wei2022chain,lei2024scaffolding}.
\cite{xu2025visual} explored pure visual reasoning relying solely on visual representations without textual modalities.
Zebra-CoT~\citep{li2025zebra,laurenccon2023obelics} provides an interleaved vision-language reasoning trajectory dataset to enhance UMMs' comprehension performance.
IRG~\citep{huang2025interleaving} generates an initial image, then iteratively refines it through reflective reasoning about quality improvements.
ROVER~\citep{liang2025rover} investigates the reciprocal relationship between generation and comprehension capabilities.
In contrast, \name{} focuses on in-context interleaved multimodal comprehension and generation.

\noindent\textbf{Benchmarks for UMMs.}\quad 
UMM capability assessment typically encompasses three dimensions: (i) \textit{Text-to-Image}: evaluated using GenEval~\citep{ghosh2023geneval} and DPGBench~\citep{hu2024ella}, which employ image detection methods~\citep{chen2019mmdetection} to ensure policy-compliant generation, and WISE~\citep{niu2025wise}, which examines complex semantic understanding and world knowledge for T2I generation; (ii) \textit{Vision Comprehension}: consistent with Vision-Language Model (VLM) evaluation protocols, using benchmarks including MME~\citep{chaoyou2023mme}, MMBench~\citep{liu2024mmbench}, MMMU~\citep{yue2024mmmu}, MM-Vet~\citep{yu2023mm}, and MathVista~\citep{lu2023mathvista}; (iii) \textit{Image Editing}: assessed via GEdit-Bench~\citep{liu2025step1x-edit} and ImgEdit~\citep{ye2025imgedit}, which challenge UMMs to maintain image identity preservation while demonstrating semantic understanding. Additionally, RISEBench and KRIS-Bench evaluate reasoning with world knowledge. These benchmarks assess generation and comprehension in isolation, whereas ROVER~\citep{liang2025rover} pioneered reciprocal cross-modal reasoning for omnimodal generation, systematically evaluating intermediate processes. \colorname{} represents the first benchmark to comprehensively evaluate interleaved multi-turn generation and understanding.

\section{Additional Examples for \colorname{}}\label{app:data-example}

\subsection{Additional Examples for \trainname{}}
\begin{figure*}[t]
    \centering
    \includegraphics[width=0.9\linewidth]{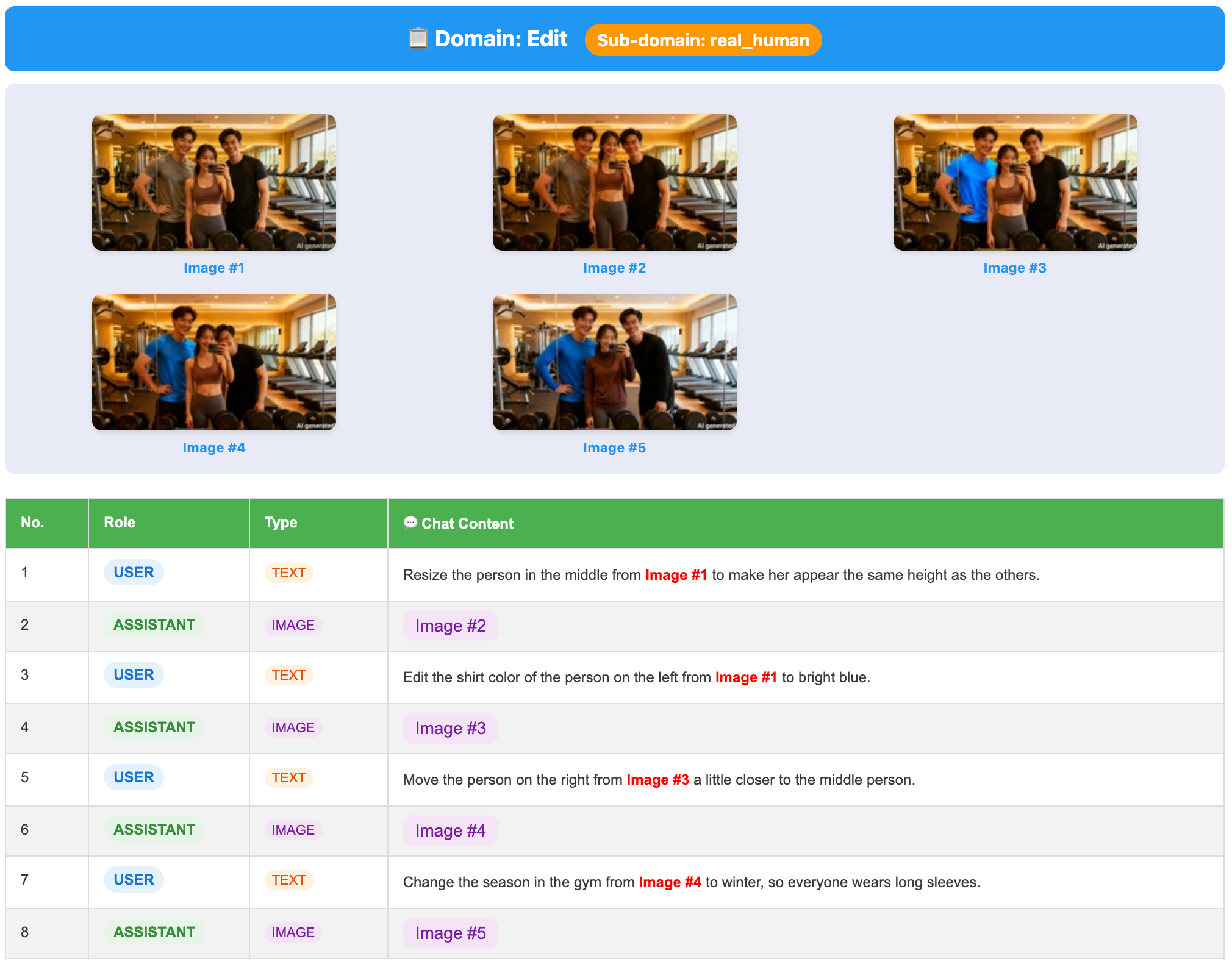}
    \caption{An example of edit in \trainname{}.}
  \vspace{-5mm}  
  \label{tab:train-edit_1}
\end{figure*}

\begin{figure*}[t]
    \centering
    \includegraphics[width=0.9\linewidth]{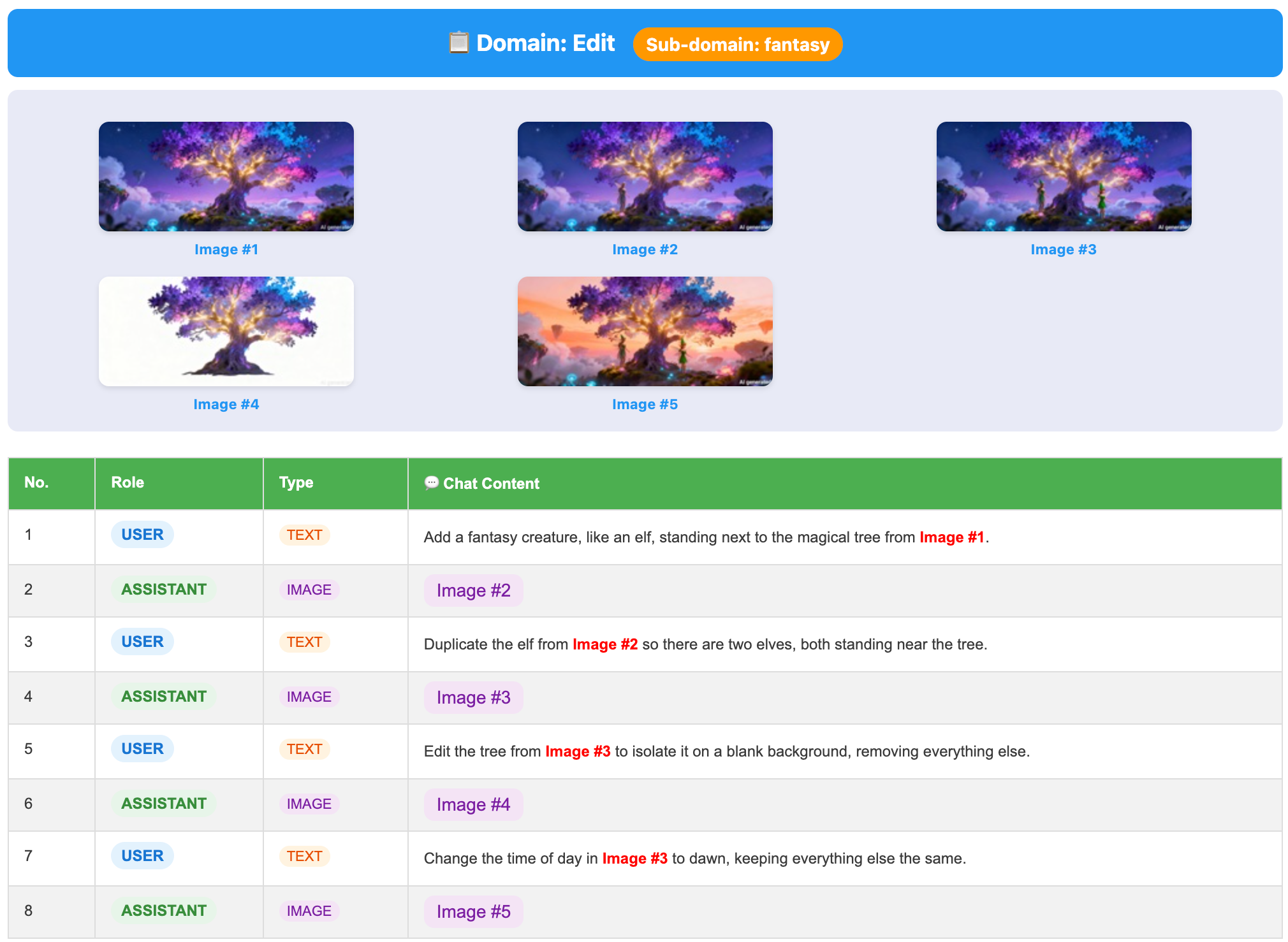}
    \caption{An example of edit in \trainname{}.}
  \vspace{-5mm}  
  \label{tab:train-edit_2}
\end{figure*}

In this appendix, we present a comprehensive collection of examples that illustrate the versatility and capabilities of our \trainname{} framework. Figure~\ref{tab:train-edit_1} and Figure~\ref{tab:train-edit_2} demonstrate complex editing operations that require significant reasoning capabilities. The first example showcases intricate manipulations that demand careful consideration of spatial relationships and semantic coherence, while the second example introduces human subjects into the composition.

\begin{figure*}[t]
    \centering
    \includegraphics[width=0.9\linewidth]{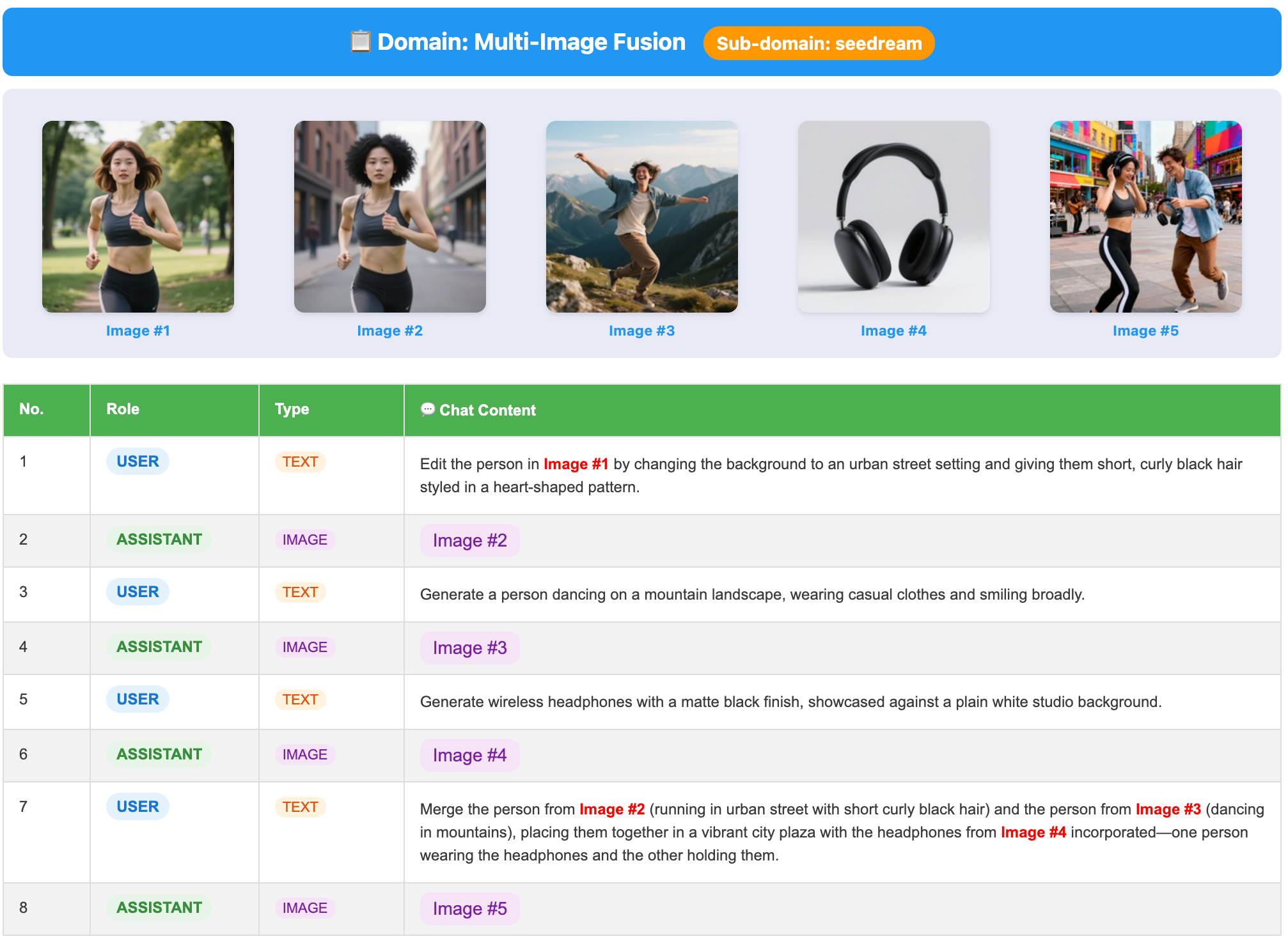}
    \caption{An example of multi-image fusion in \trainname{}.}
  \vspace{-5mm}  
  \label{tab:train-fusion_1}
\end{figure*}

\begin{figure*}[t]
    \centering
    \includegraphics[width=0.9\linewidth]{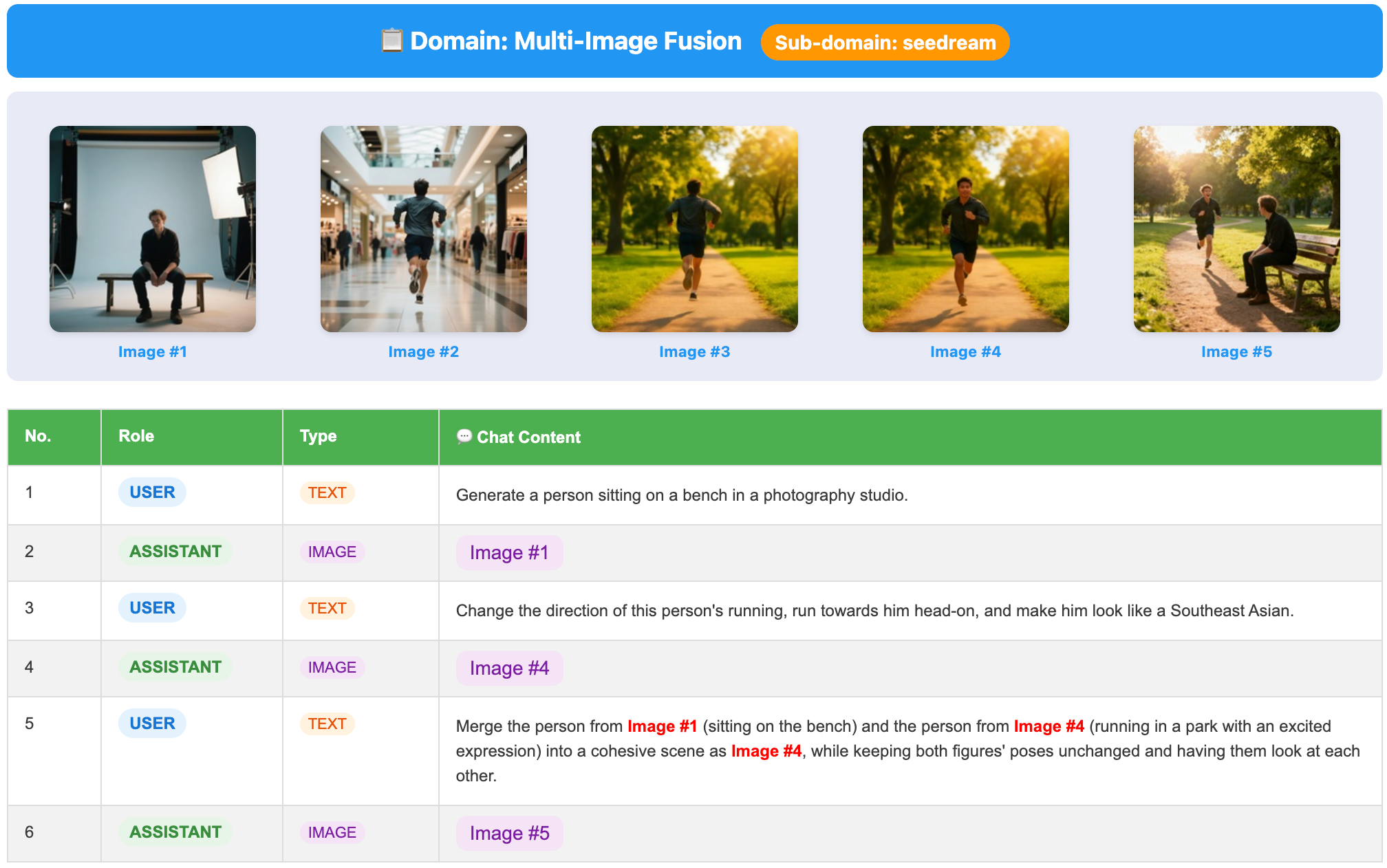}
    \caption{An example of multi-image fusion in \trainname{}.}
  \vspace{-5mm}  
  \label{tab:train-fusion_2}
\end{figure*}

\begin{figure*}[t]
    \centering
    \includegraphics[width=0.9\linewidth]{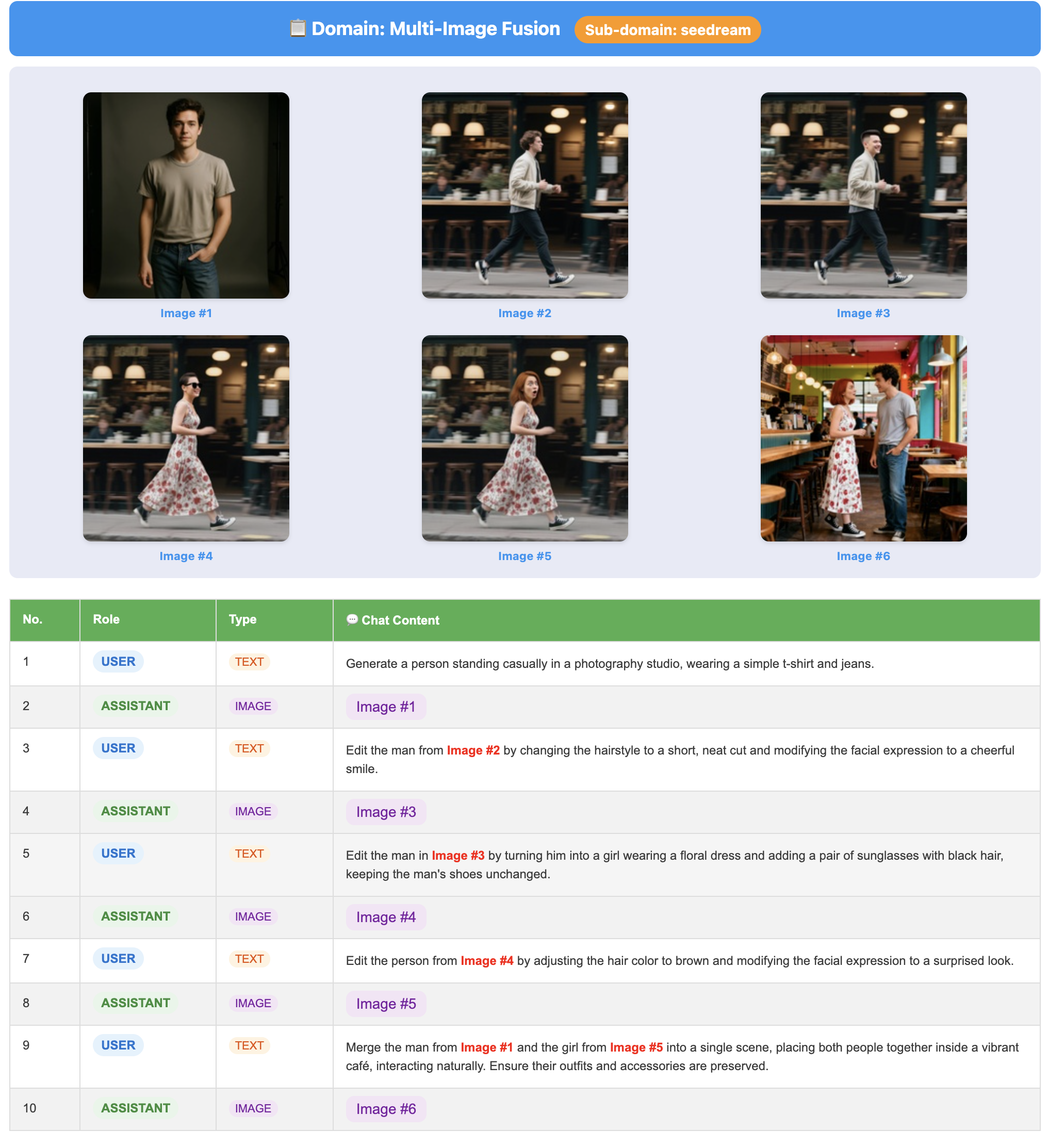}
    \caption{An example of multi-image fusion in \trainname{}.}
  \vspace{-5mm}  
  \label{tab:train-fusion_3}
\end{figure*}

\begin{figure*}[t]
    \centering
    \includegraphics[width=0.9\linewidth]{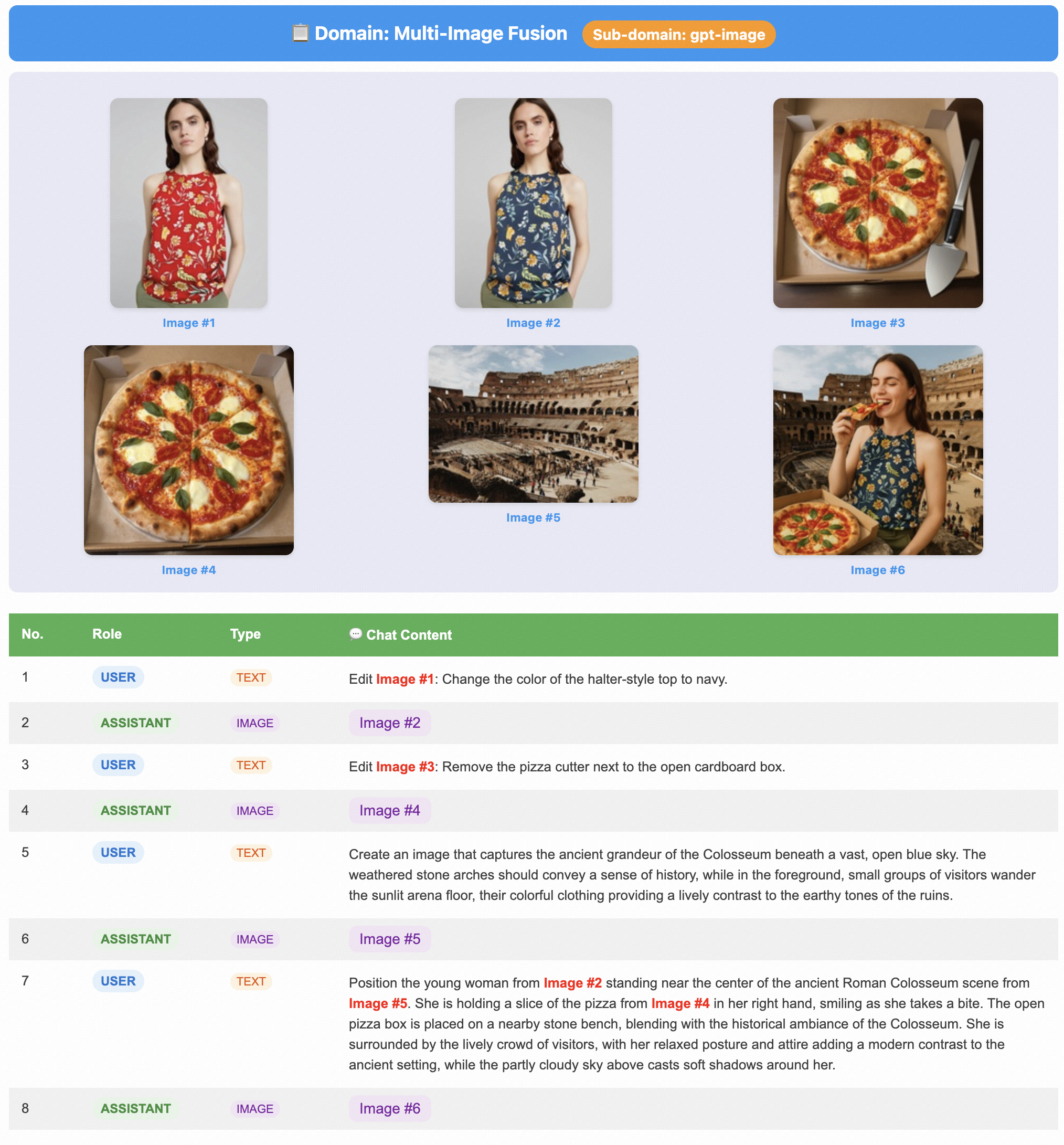}
    \caption{An example of multi-image fusion in \trainname{}.}
  \vspace{-5mm}  
  \label{tab:train-fusion_4}
\end{figure*}

For Multi-Image Fusion operations, we provide four illustrative examples in Figures~\ref{tab:train-fusion_1}--\ref{tab:train-fusion_4}. Figure~\ref{tab:train-fusion_3} demonstrates the model's ability to preserve footwear details during fusion operations. Figure~\ref{tab:train-fusion_2} exhibits dual-task face processing capabilities. Figure~\ref{tab:train-fusion_1} highlights the precise execution of specific hairstyle requirements and depicts scenarios where headphones are both held by one subject and worn by another, showcasing the model's understanding of object interactions across multiple contexts.

\begin{figure*}[t]
    \centering
    \includegraphics[width=0.9\linewidth]{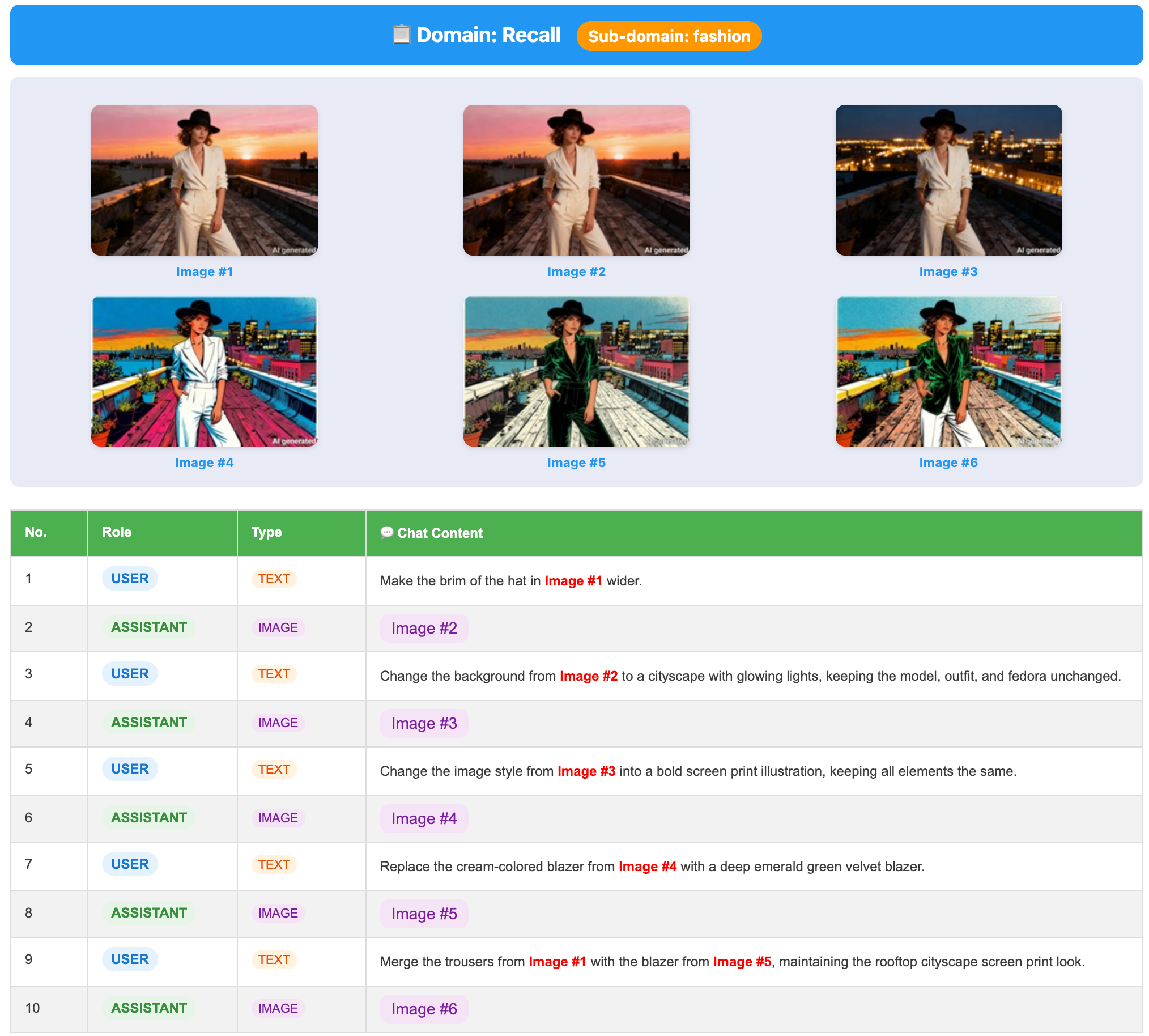}
    \caption{An example of recall in \trainname{}.}
  \vspace{-5mm}  
  \label{tab:train-recall_1}
\end{figure*}

\begin{figure*}[t]
    \centering
    \includegraphics[width=0.9\linewidth]{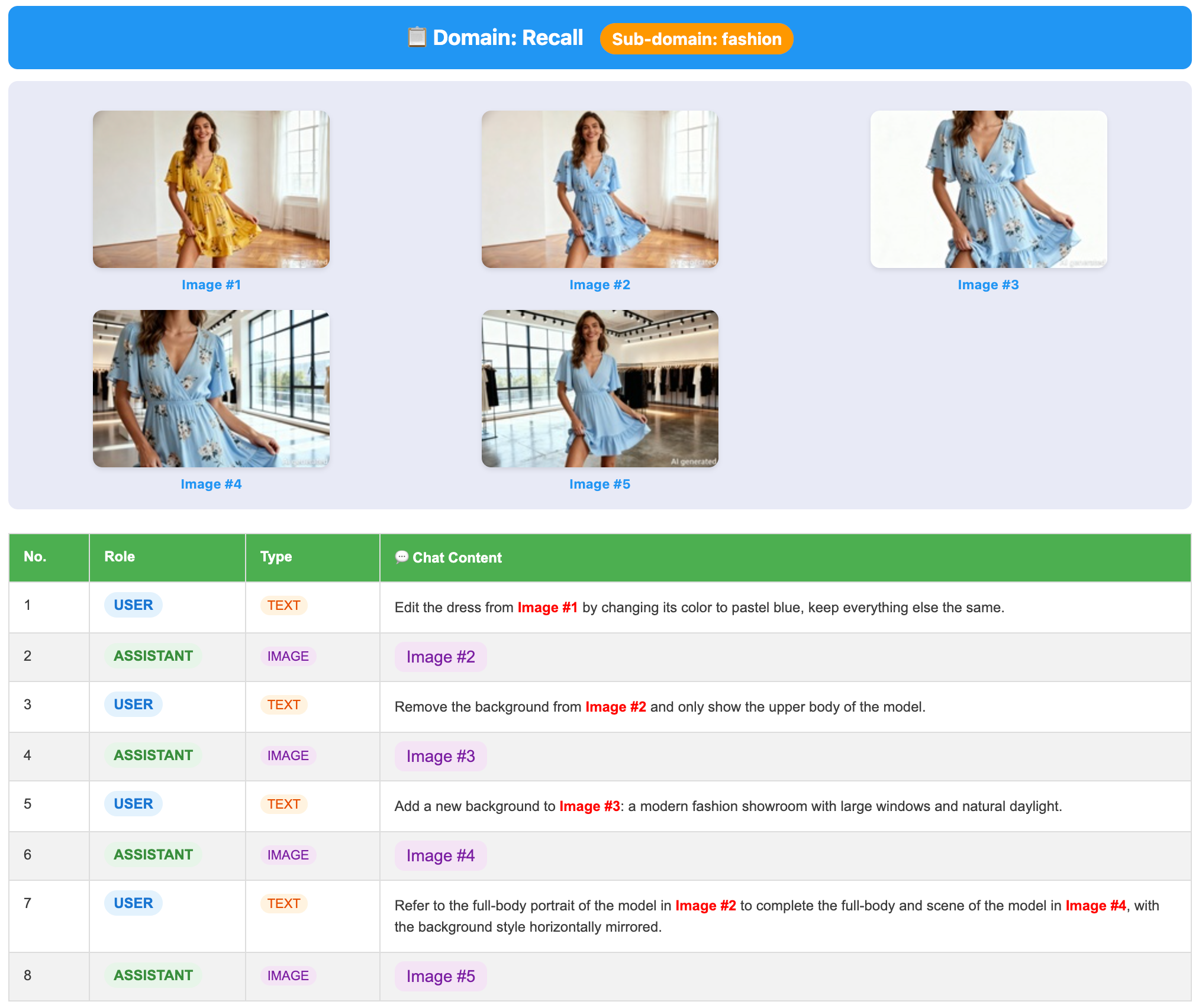}
    \caption{An example of recall in \trainname{}.}
  \vspace{-5mm}  
  \label{tab:train-recall_2}
\end{figure*}

\begin{figure*}[t]
    \centering
    \includegraphics[width=0.9\linewidth]{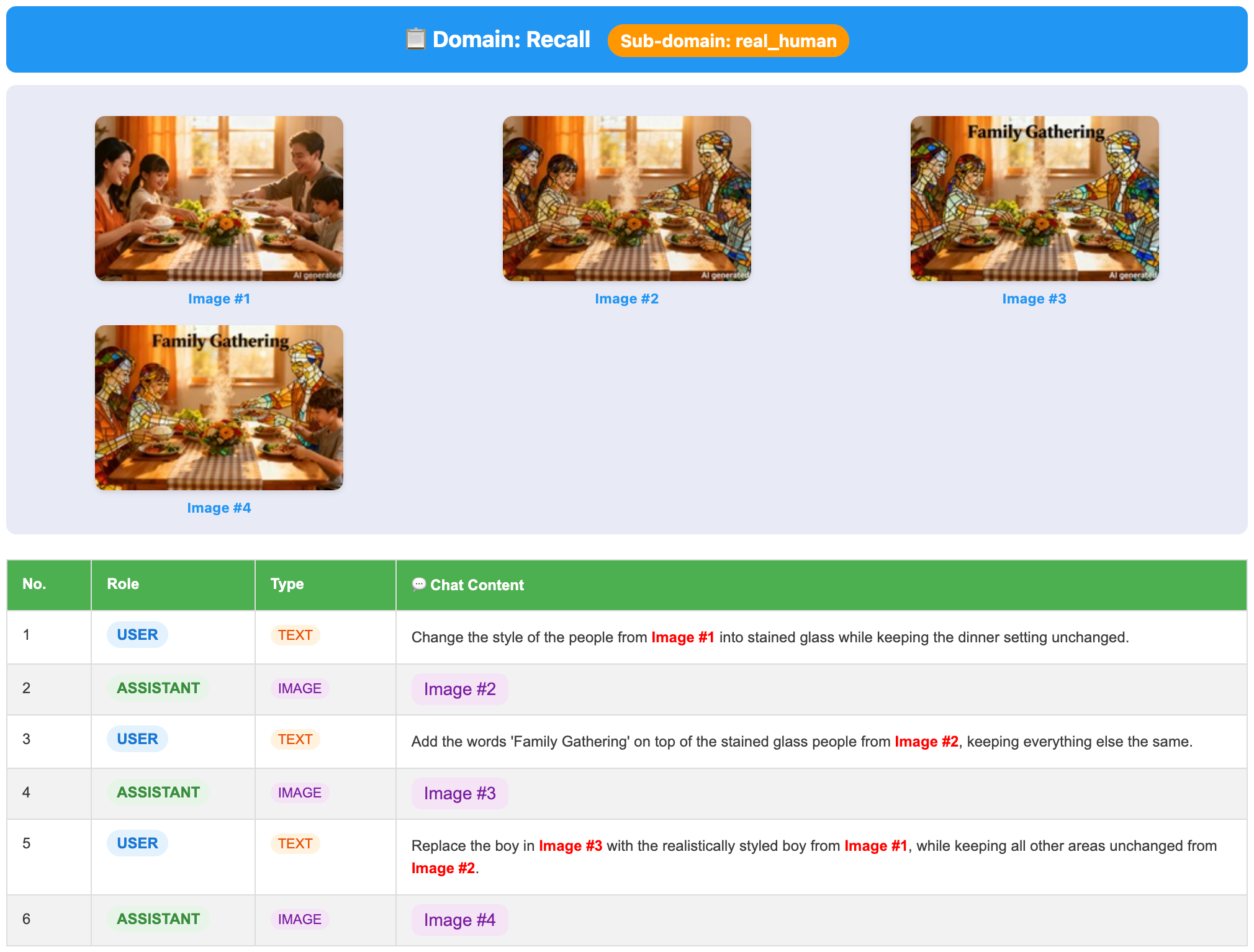}
    \caption{An example of recall in \trainname{}.}
  \vspace{-5mm}  
  \label{tab:train-recall_3}
\end{figure*}

The Recall capability is exemplified in Figure~\ref{tab:train-recall_1}, Figure~\ref{tab:train-recall_2}, and Figure~\ref{tab:train-recall_3}. In the first example, the model successfully restores previously removed trousers to the subject. The second example demonstrates the model's ability to reference a full-body model from Image \#2 to reconstruct the complete body and scene in Image \#4, while implementing a horizontally symmetrical background transformation. The third example shows the targeted reinsertion of a single human subject into the composition.

\begin{figure*}[t]
    \centering
    \includegraphics[width=0.9\linewidth]{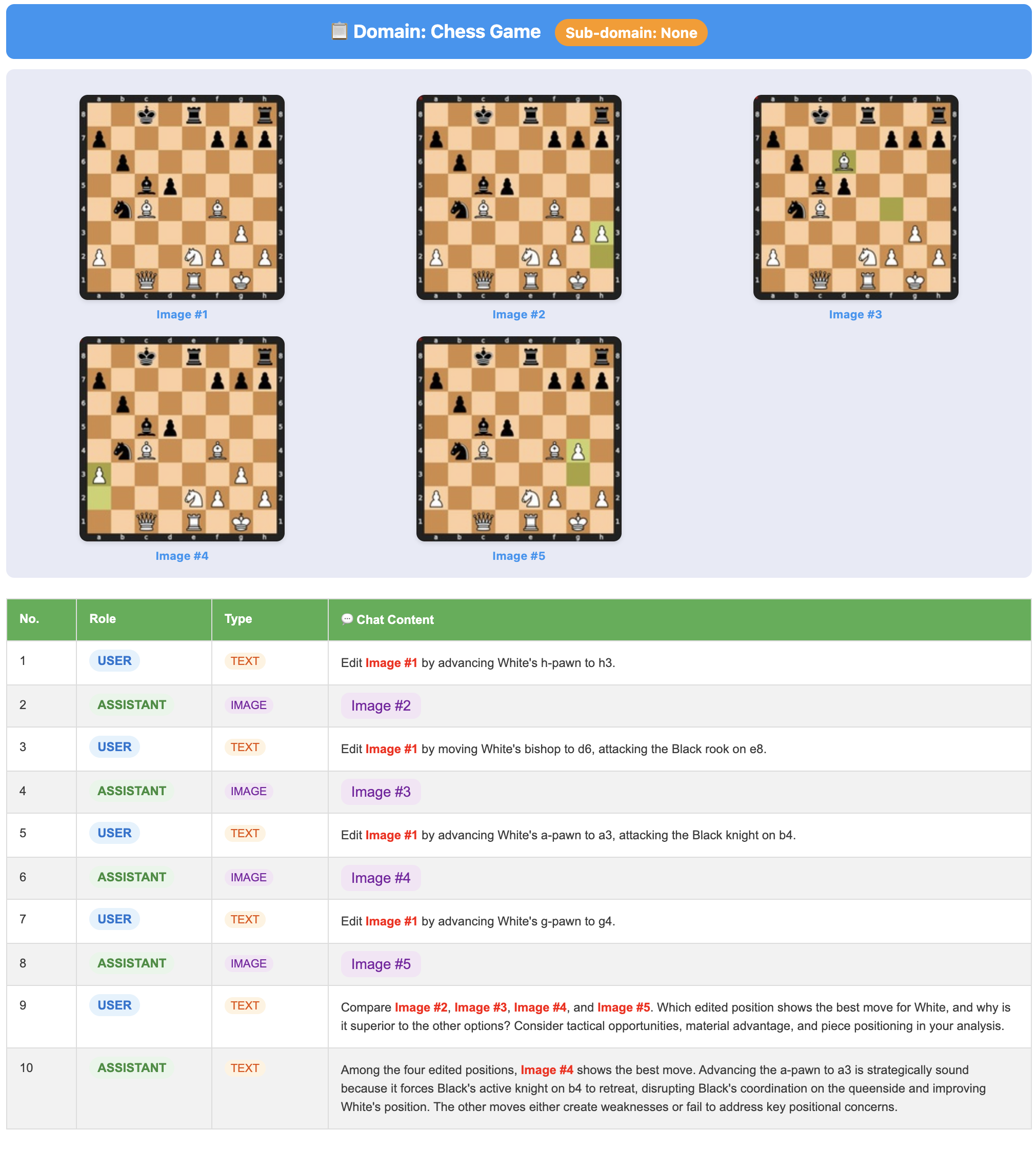}
    \caption{An example of Chess Game in \trainname{}.}
  \vspace{-5mm}  
  \label{tab:train-chess_1}
\end{figure*}

\begin{figure*}[t]
    \centering
    \includegraphics[width=0.9\linewidth]{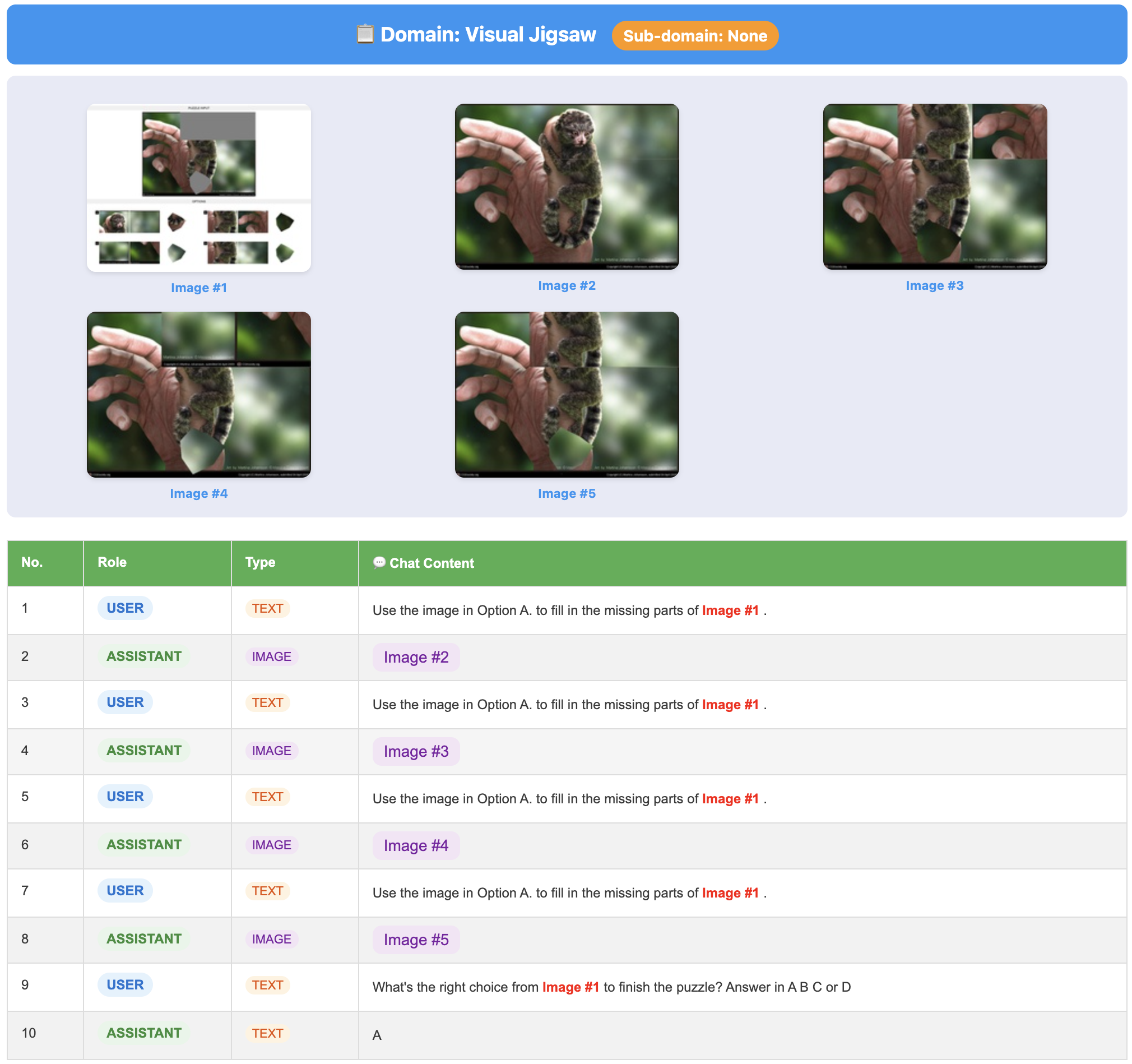}
    \caption{An example of Chess Game in \trainname{}.}
  \vspace{-5mm}  
  \label{tab:train-jigsaw_1}
\end{figure*}

Additionally, we present specialized examples for Chess Game manipulation in Figure~\ref{tab:train-chess_1} and Visual JigSaw processing in Figure~\ref{tab:train-jigsaw_1}, further demonstrating the framework's adaptability to structured visual reasoning tasks.




\subsection{More example for \testname{}}\label{app:case-test}

\begin{figure*}[t]
  \centering
  \includegraphics[width=0.9\linewidth]{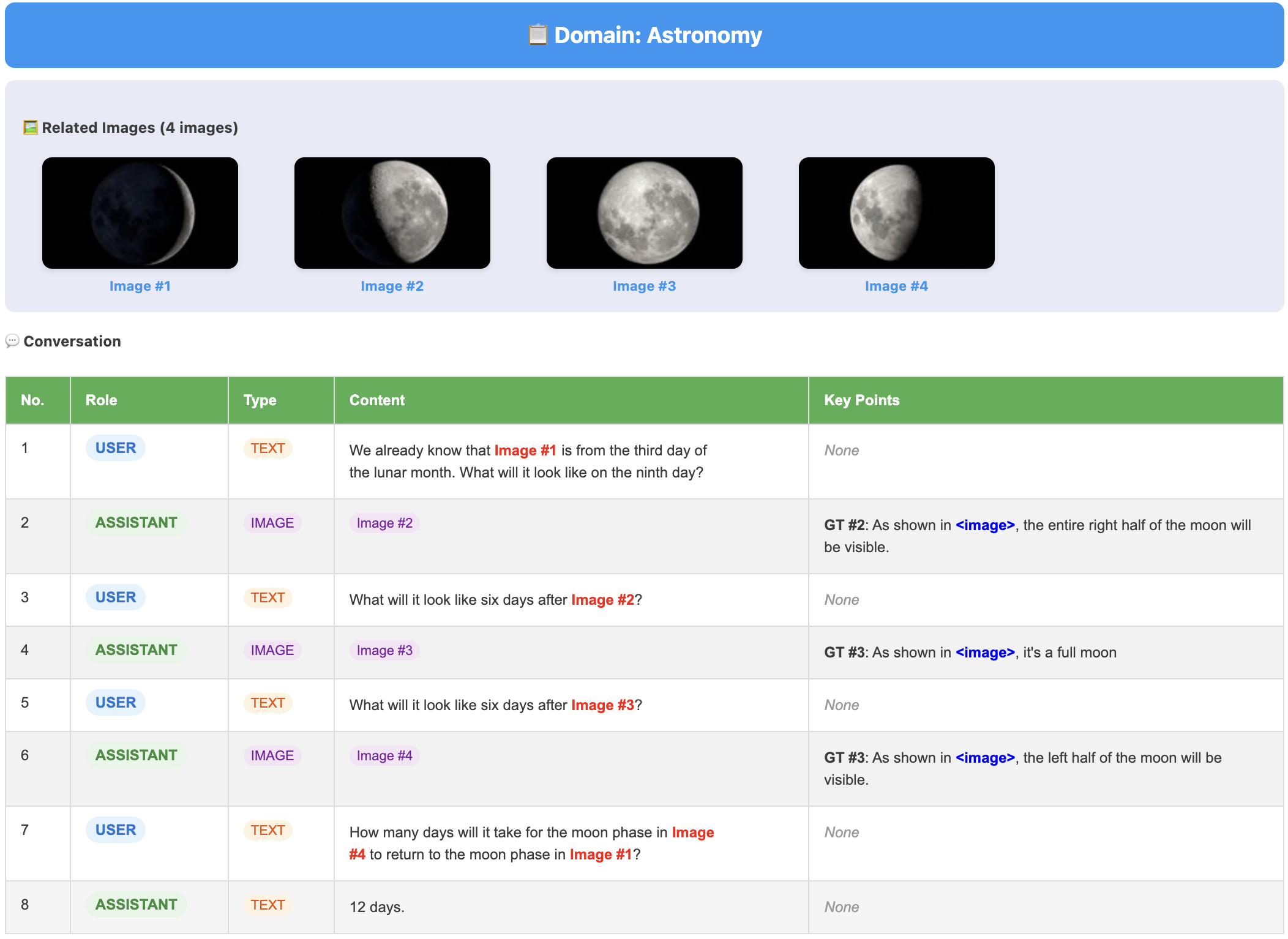}
  \caption{An example of astronomy domain testing the model's understanding of celestial objects and phenomena.}
  \vspace{-5mm}  
  \label{fig:test_astronomy}
\end{figure*}

\begin{figure*}[t]
  \centering
  \includegraphics[width=0.9\linewidth]{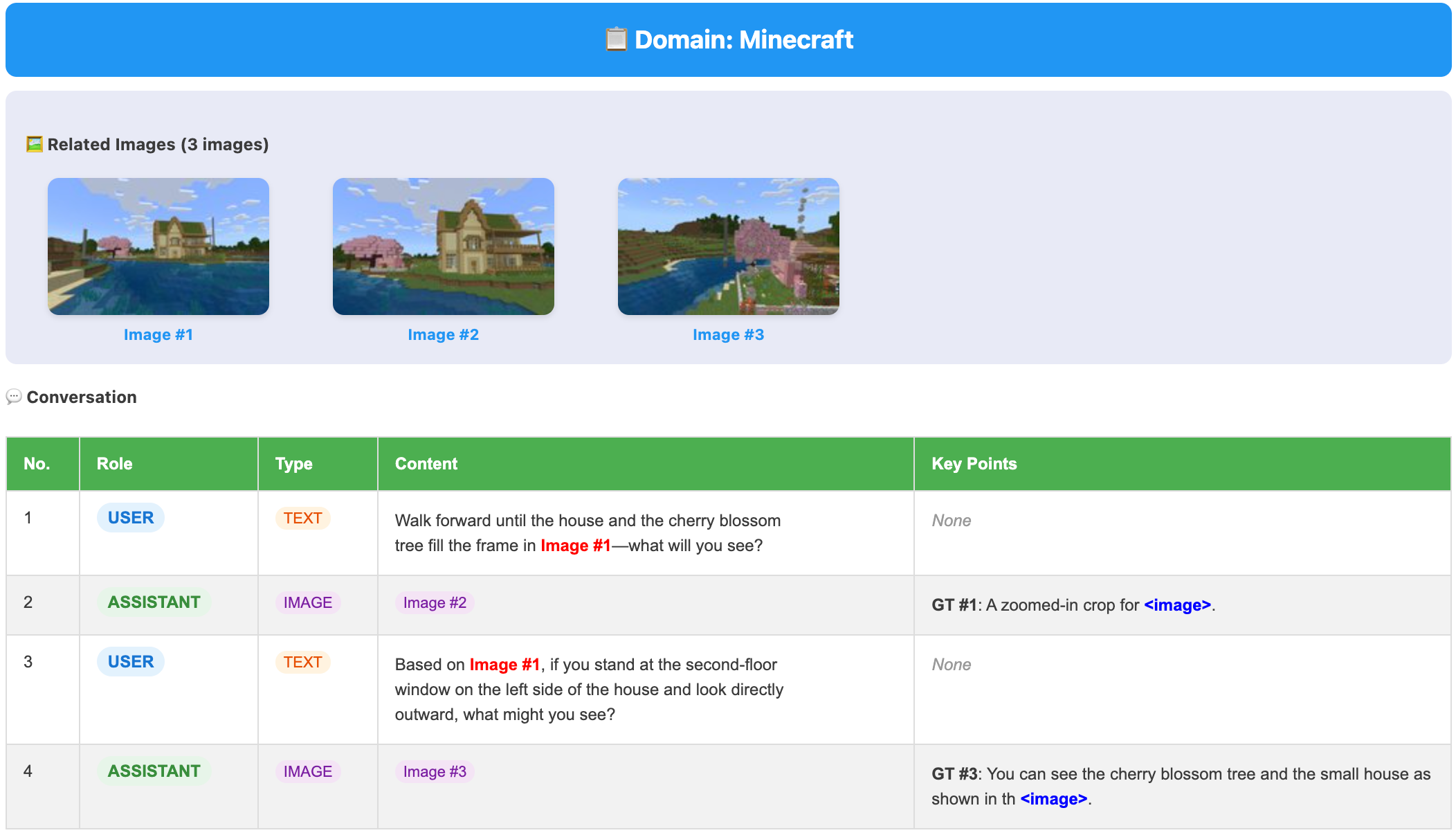}
  \caption{An example of Minecraft domain testing the model's understanding of the game mechanics and environments.}
  \vspace{-5mm}  
  \label{fig:test_minecraft}
\end{figure*}

\begin{figure*}[t]
  \centering
  \includegraphics[width=0.9\linewidth]{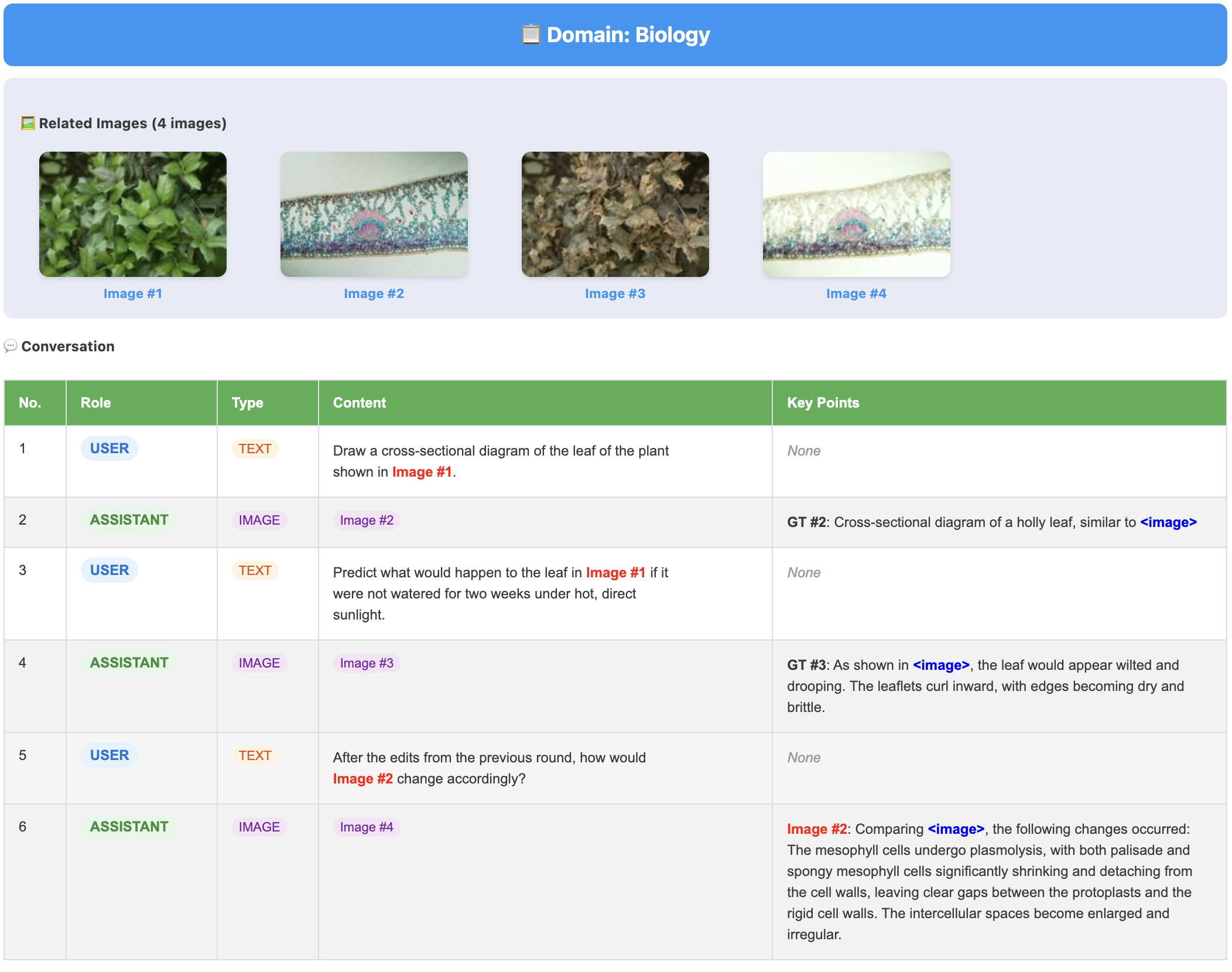}
  \caption{An example of biology domain testing the model's understanding of biological structures and processes.}
  \vspace{-5mm}  
  \label{fig:test_biology}
\end{figure*}

\begin{figure*}[t]
  \centering
  \includegraphics[width=0.9\linewidth]{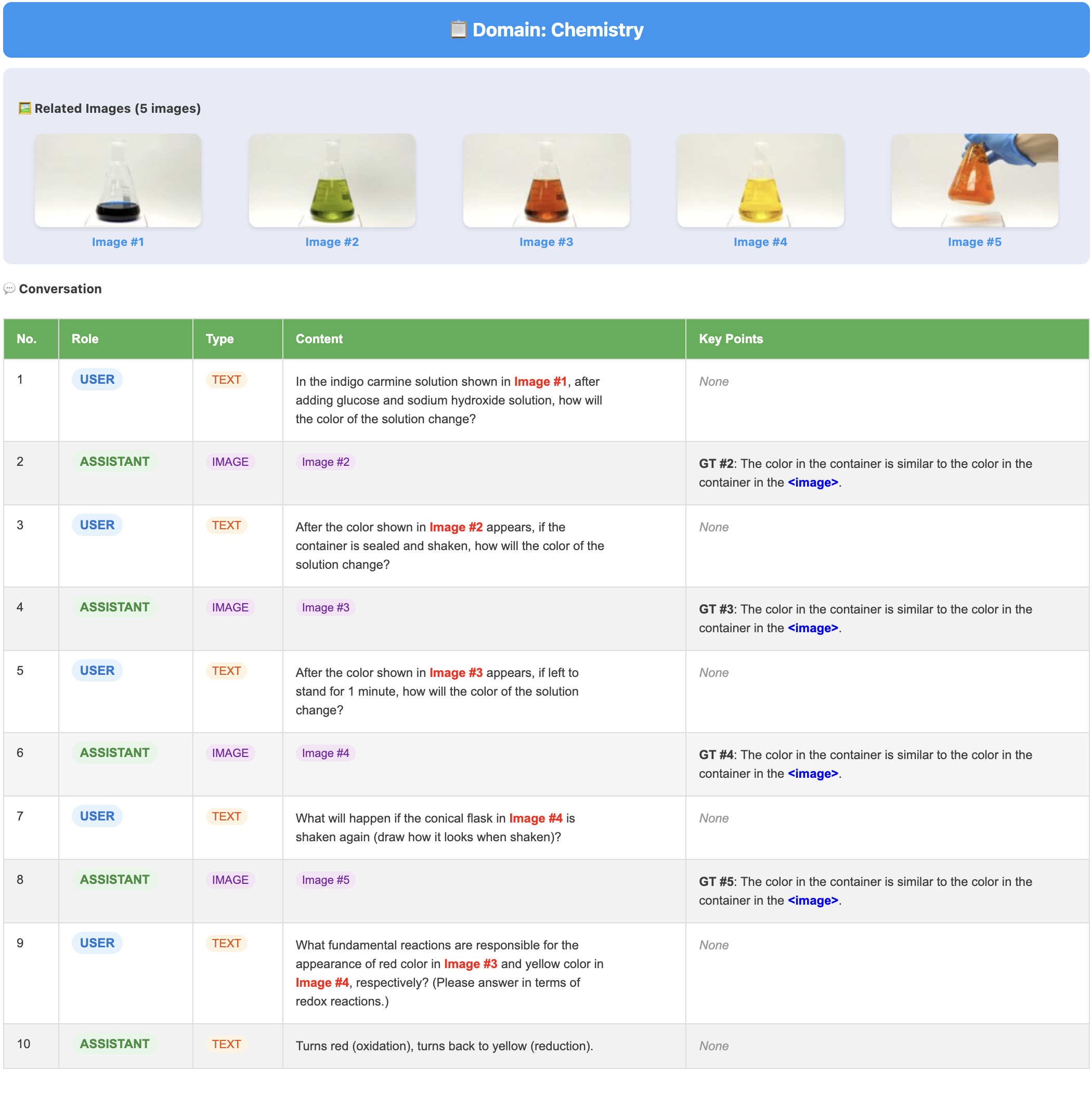}
  \caption{An example of chemistry domain testing the model's understanding of chemical structures and reactions.}
  \vspace{-5mm}  
  \label{fig:test_chemistry}
\end{figure*}

\begin{figure*}[t]
  \centering
  \includegraphics[width=0.9\linewidth]{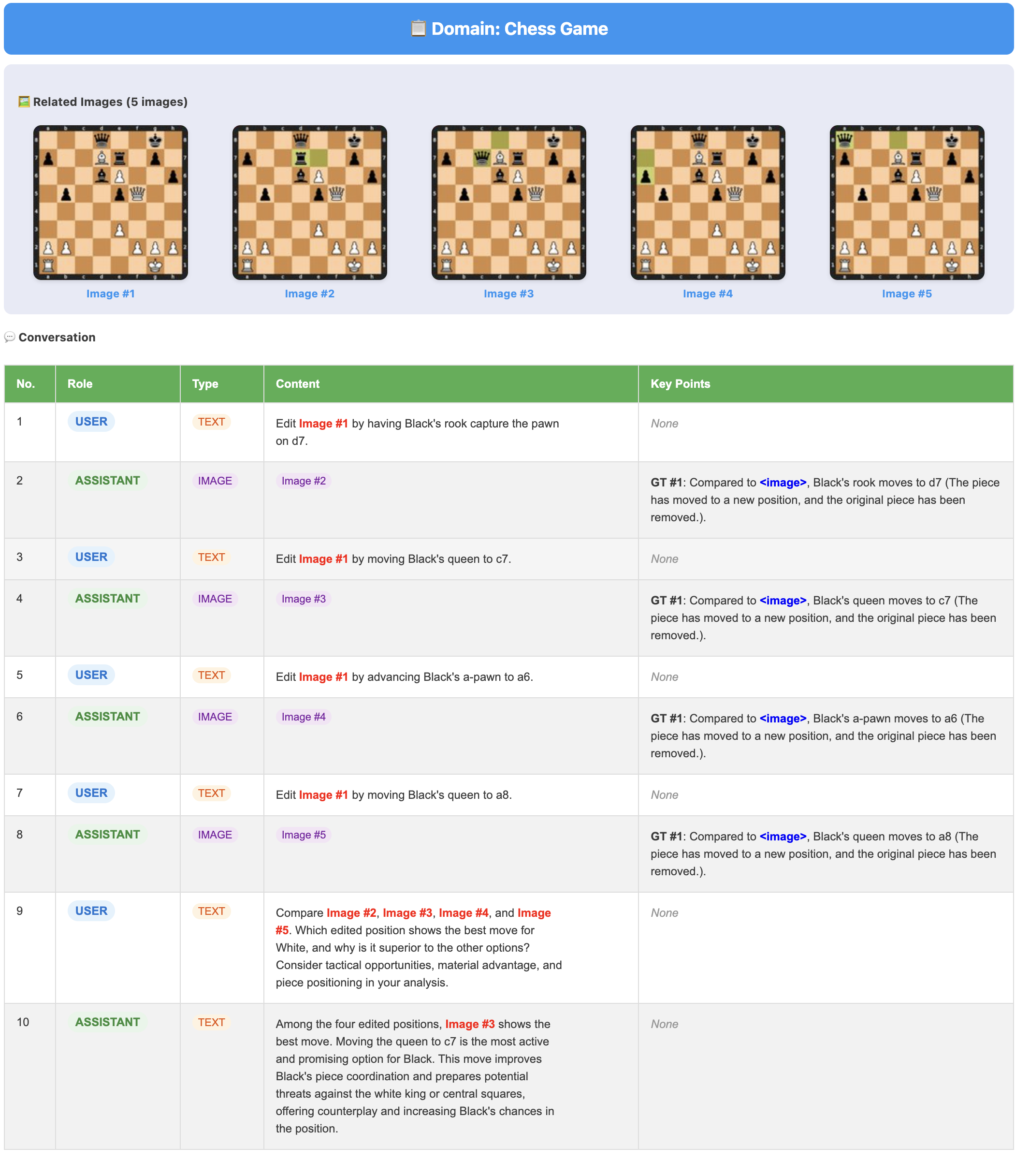}
  \caption{An example of chess game analysis testing the model's understanding of chess positions and strategies.}
  \vspace{-5mm}  
  \label{fig:test_chess_game}
\end{figure*}

\begin{figure*}[t]
  \centering
  \includegraphics[width=0.9\linewidth]{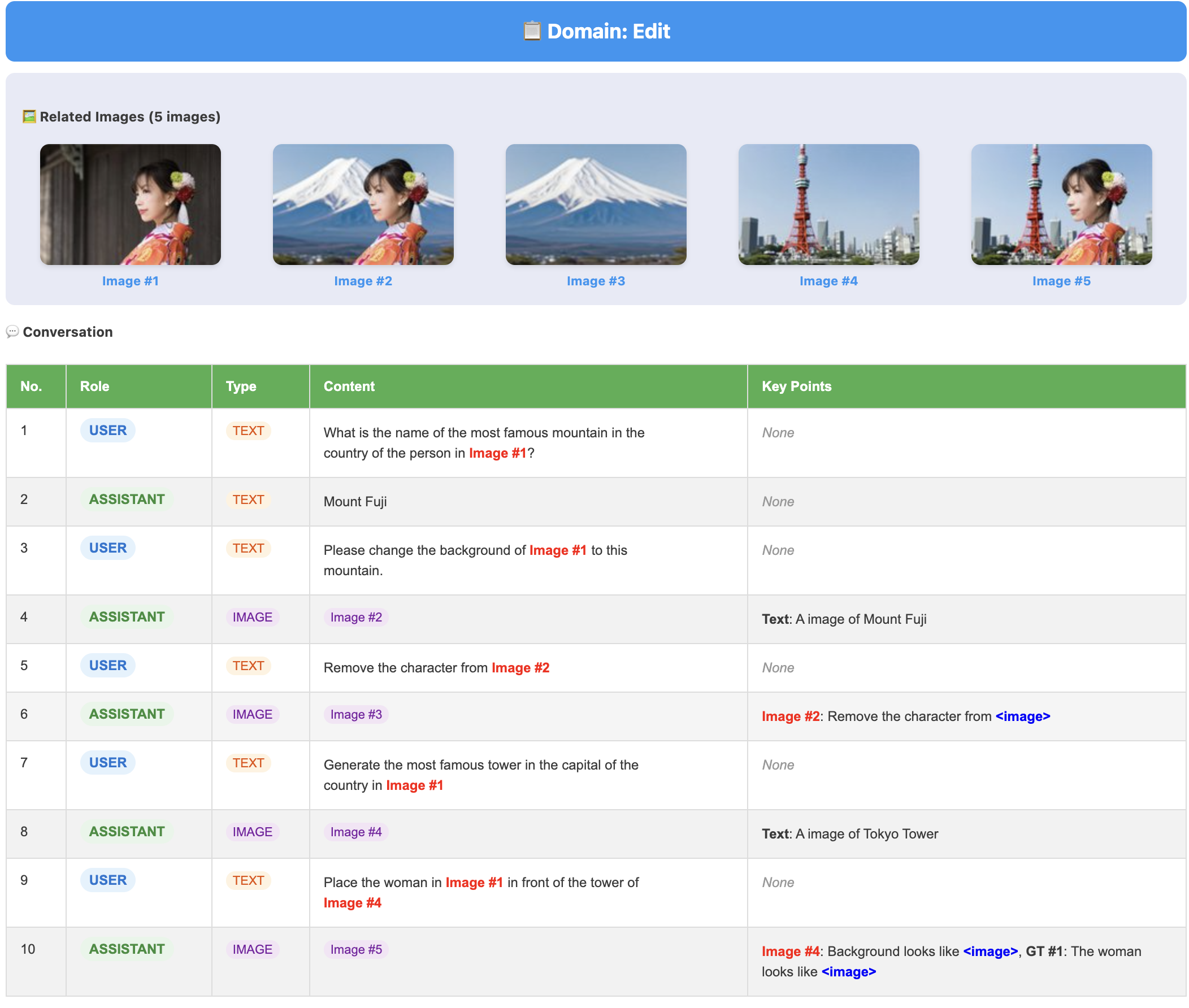}
  \caption{An example of image editing task testing the model's ability to understand and suggest visual modifications.}
  \vspace{-5mm}  
  \label{fig:test_edit}
\end{figure*}

\begin{figure*}[t]
  \centering
  \includegraphics[width=0.9\linewidth]{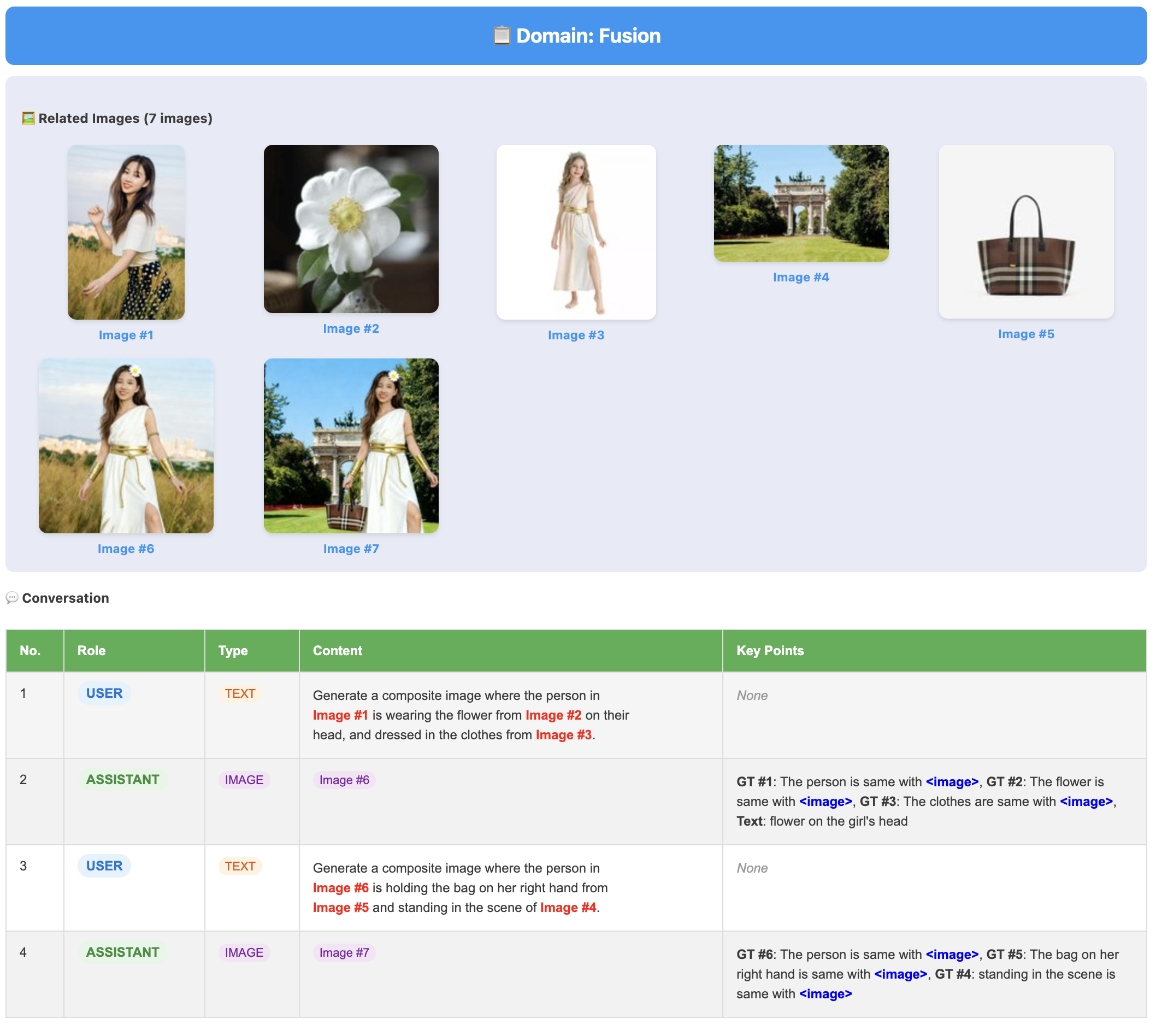}
  \caption{An example of fusion domain testing the model's understanding of nuclear fusion concepts and processes.}
  \vspace{-5mm}  
  \label{fig:test_fusion}
\end{figure*}

\begin{figure*}[t]
  \centering
  \includegraphics[width=0.9\linewidth]{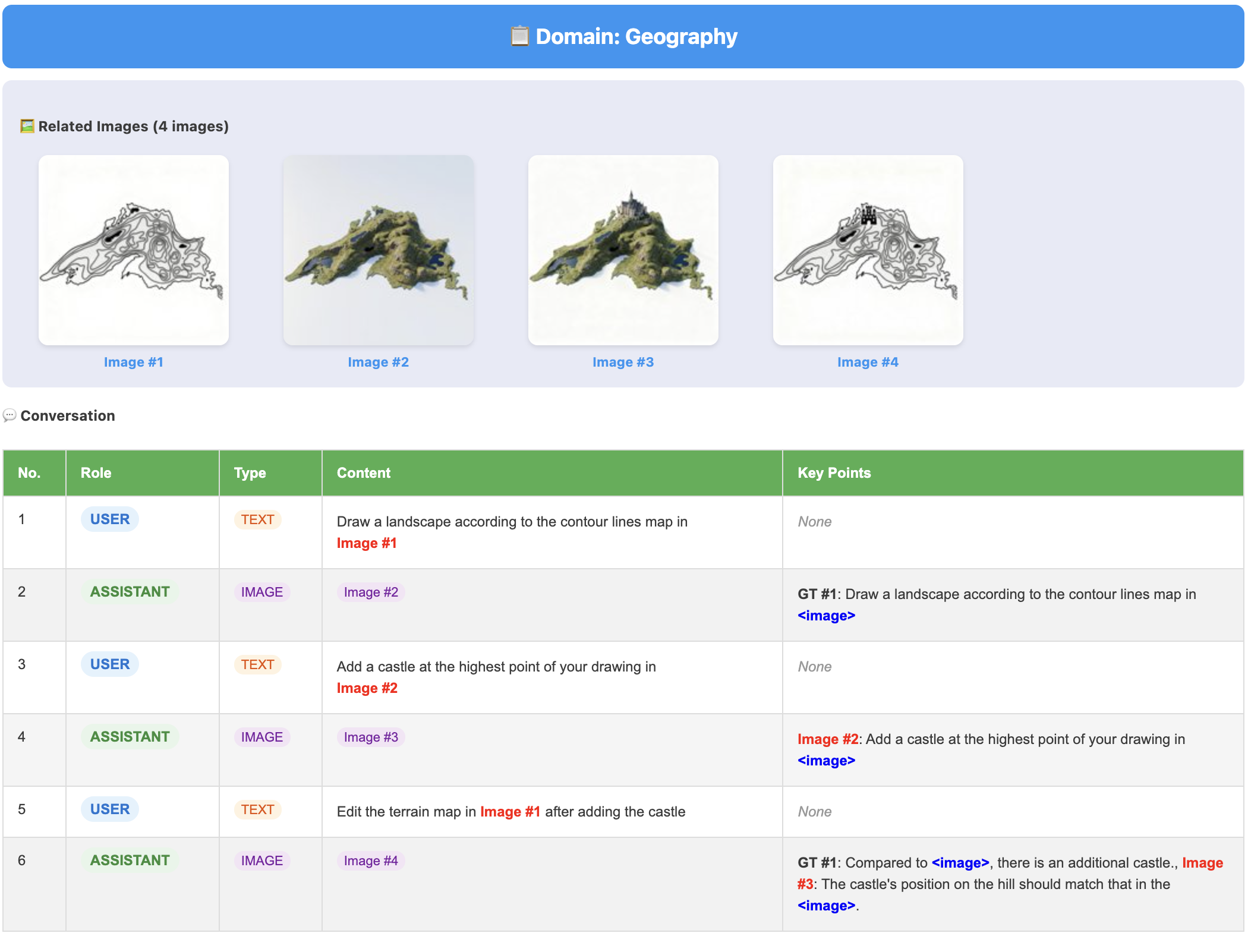}
  \caption{An example of geography domain testing the model's understanding of geographical features and locations.}
  \vspace{-5mm}  
  \label{fig:test_geography}
\end{figure*}

\begin{figure*}[t]
  \centering
  \includegraphics[width=0.9\linewidth]{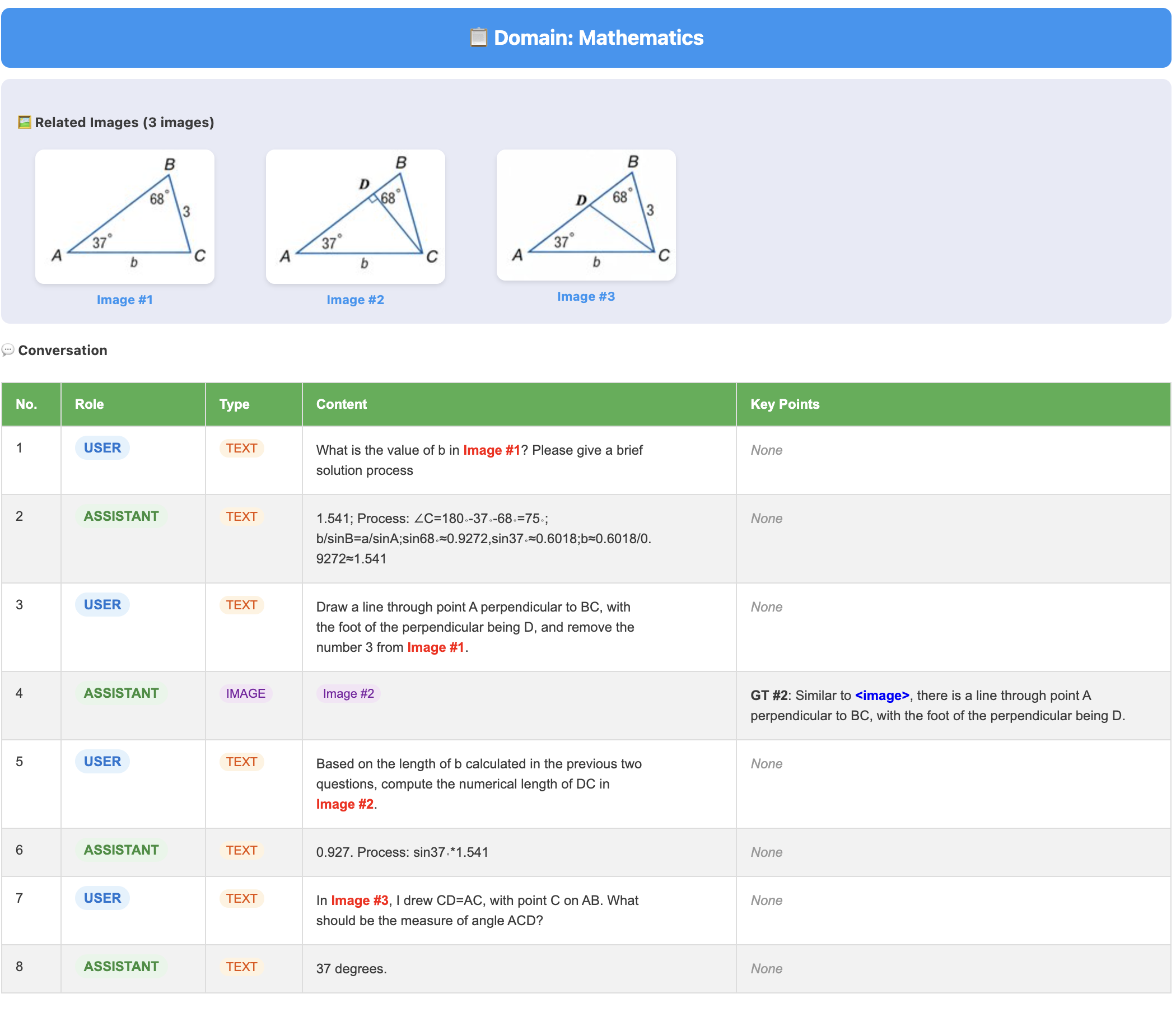}
  \caption{An example of mathematics domain testing the model's problem-solving and reasoning abilities.}
  \vspace{-5mm}  
  \label{fig:test_mathematics}
\end{figure*}

\begin{figure*}[t]
  \centering
  \includegraphics[width=0.9\linewidth]{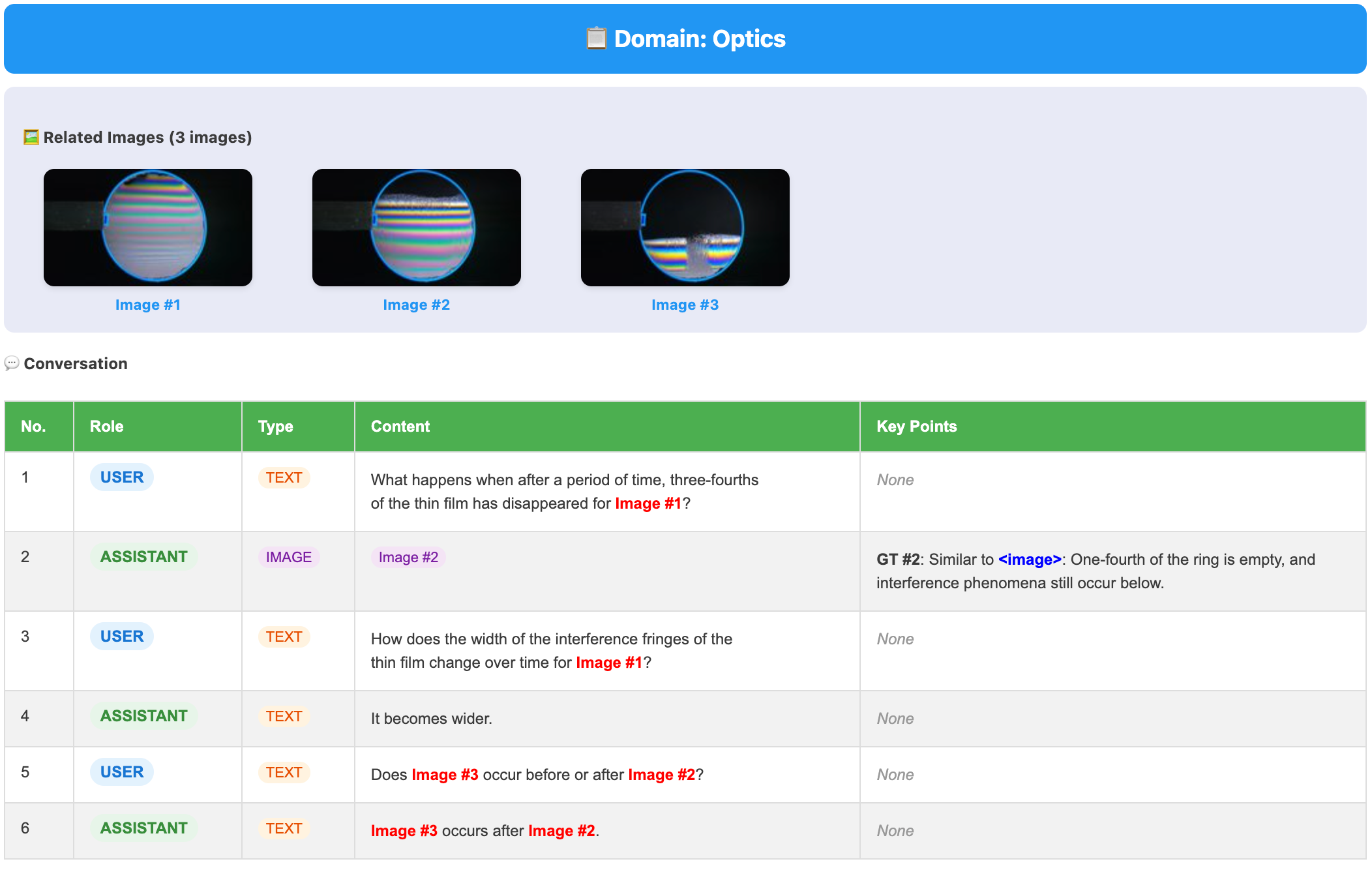}
  \caption{An example of optics domain testing the model's understanding of optical principles and phenomena.}
  \vspace{-5mm}  
  \label{fig:test_optics}
\end{figure*}

\begin{figure*}[t]
  \centering
  \includegraphics[width=0.9\linewidth]{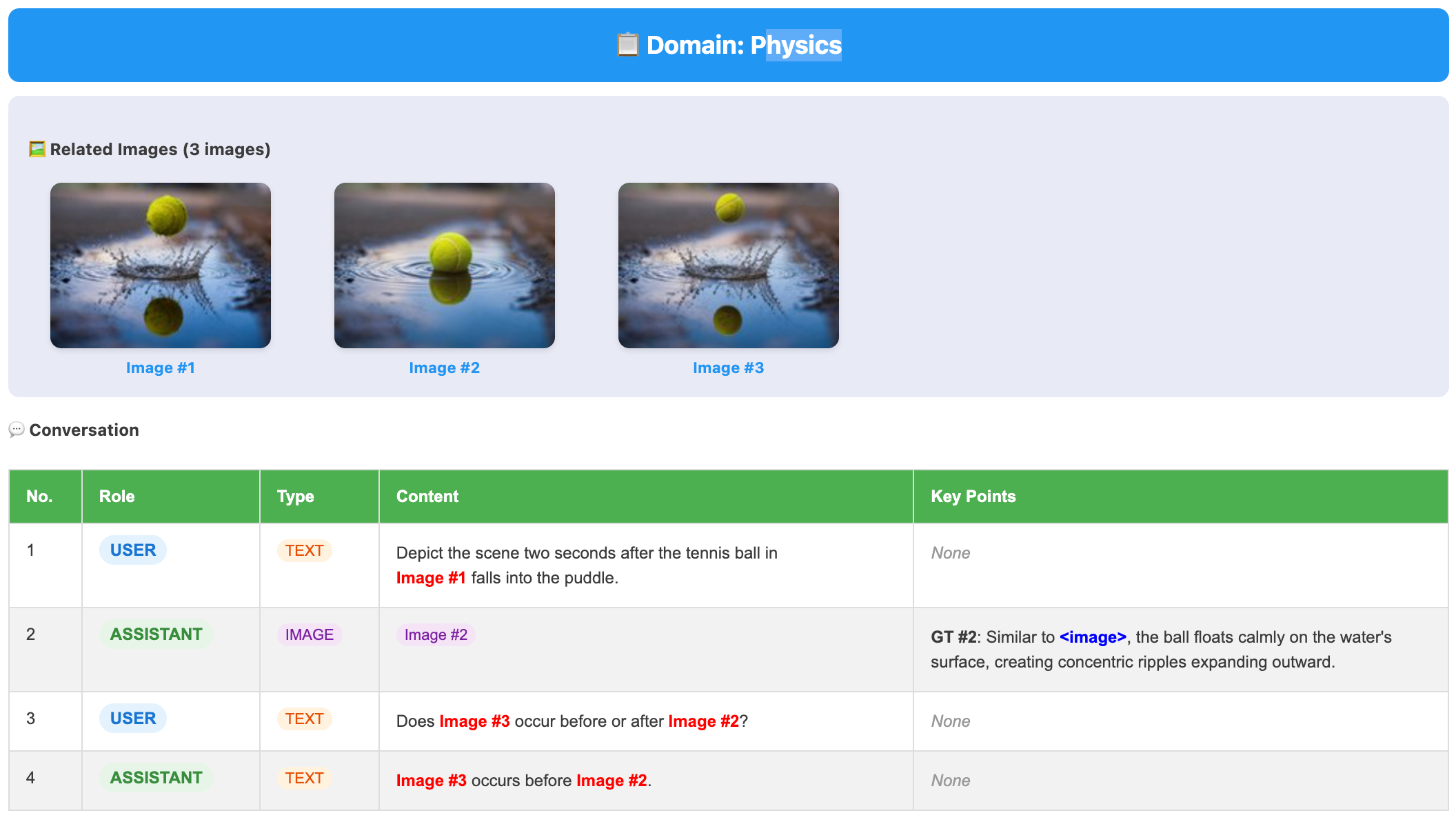}
  \caption{An example of physics domain testing the model's understanding of physical laws and principles.}
  \vspace{-5mm}  
  \label{fig:test_physics}
\end{figure*}

\begin{figure*}[t]
  \centering
  \includegraphics[width=0.9\linewidth]{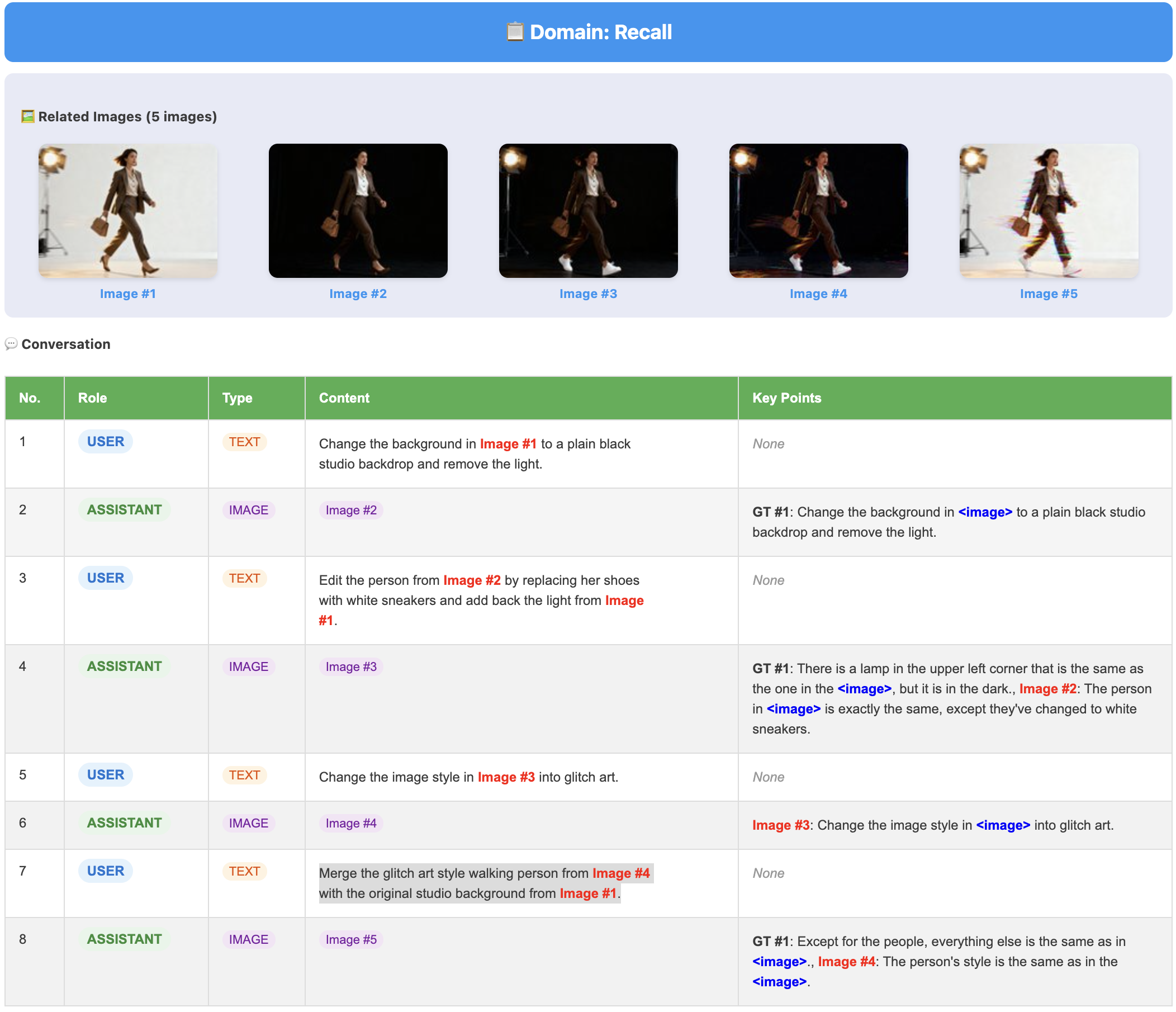}
  \caption{An example of recall task testing the model's memory and information retrieval capabilities.}
  \vspace{-5mm}  
  \label{fig:test_recall}
\end{figure*}

\begin{figure*}[t]
  \centering
  \includegraphics[width=0.9\linewidth]{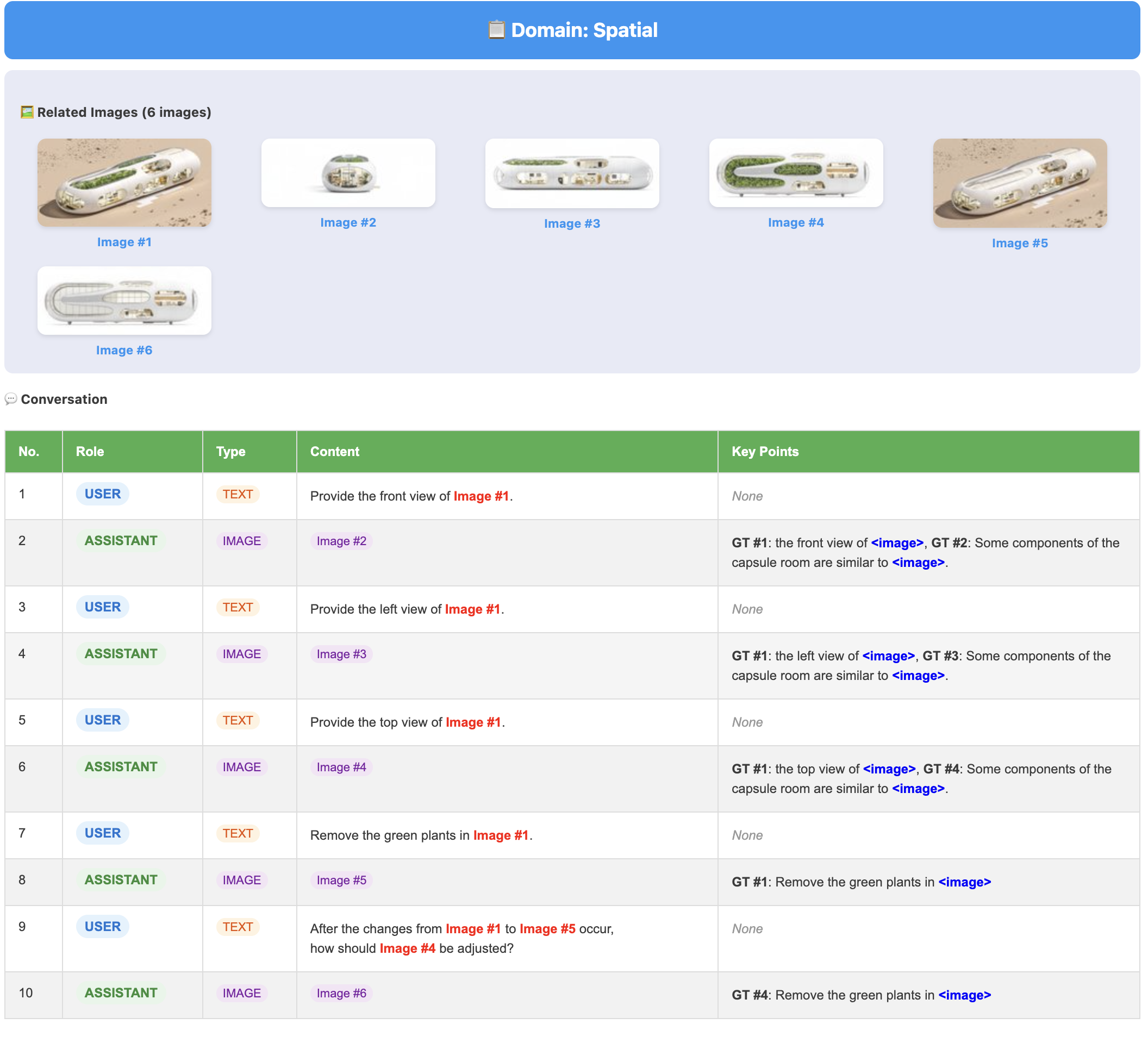}
  \caption{An example of spatial reasoning task testing the model's understanding of spatial relationships and transformations.}
  \vspace{-5mm}  
  \label{fig:test_spatial}
\end{figure*}

\begin{figure*}[t]
  \centering
  \includegraphics[width=0.9\linewidth]{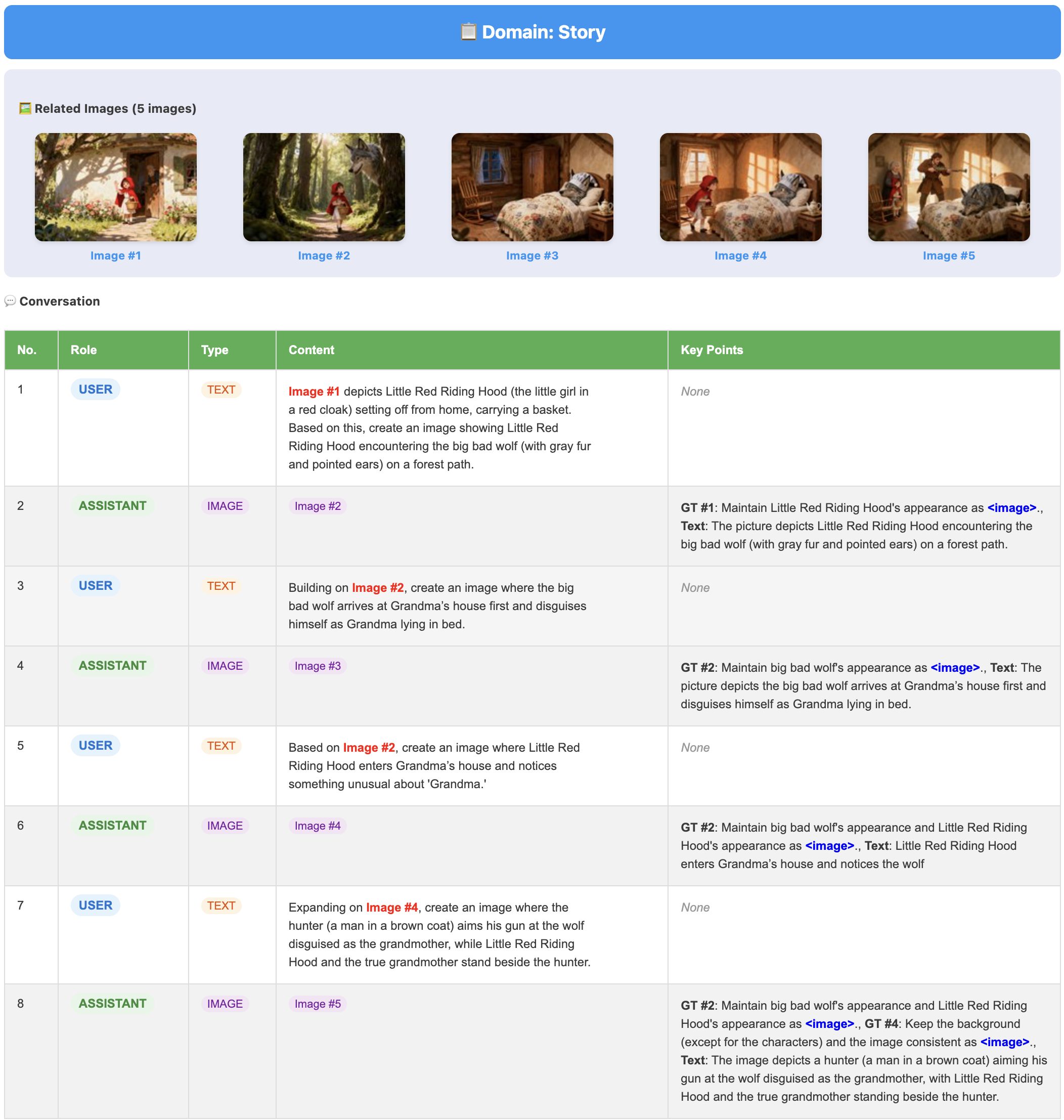}
  \caption{An example of story comprehension task testing the model's understanding of narratives and contexts.}
  \vspace{-5mm}  
  \label{fig:test_story}
\end{figure*}

\begin{figure*}[t]
  \centering
  \includegraphics[width=0.9\linewidth]{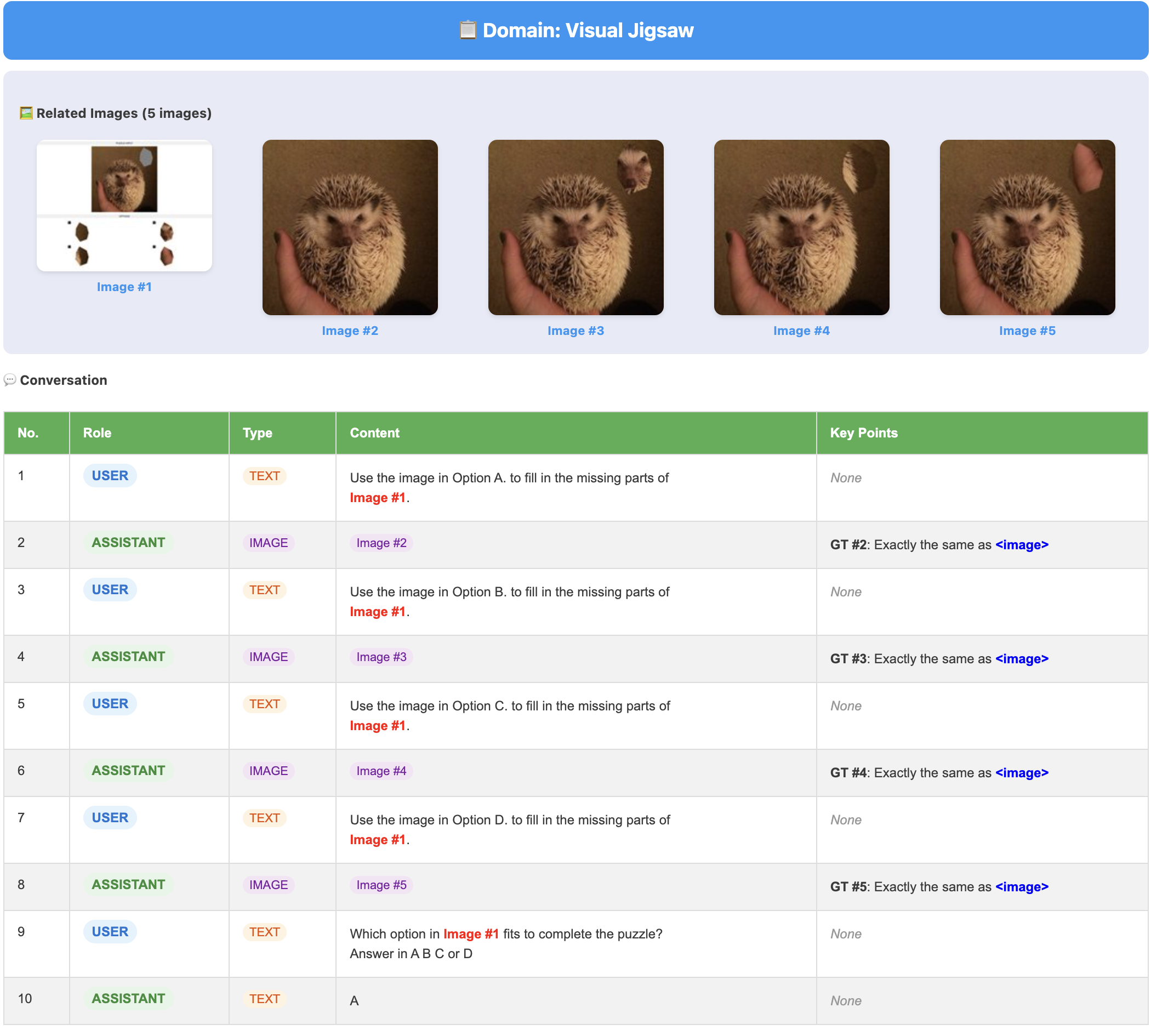}
  \caption{An example of visual jigsaw task testing the model's ability to understand and reconstruct visual patterns.}
  \vspace{-5mm}  
  \label{fig:test_visual_jigsaw}
\end{figure*}

This section presents the details of the examples shown in Figure~\ref{fig:test}. Figure~\ref{fig:test_astronomy} demonstrates astronomical concepts, while Figure~\ref{fig:test_biology} tests biological knowledge. Mathematical reasoning is evaluated in Figure~\ref{fig:test_mathematics}, and physical principles are examined in Figure~\ref{fig:test_physics}. The model's chemistry knowledge is assessed in Figure~\ref{fig:test_chemistry}, and fusion-related concepts in Figure~\ref{fig:test_fusion}. Geographic reasoning is presented in Figure~\ref{fig:test_geography}. The model's game understanding capabilities are tested through chess problems in Figure~\ref{fig:test_chess_game} and Minecraft scenarios in Figure~\ref{fig:test_minecraft}. Optical principles are demonstrated in Figure~\ref{fig:test_optics}. The model's memory and recall abilities are evaluated in Figure~\ref{fig:test_recall}, while spatial reasoning is tested in Figure~\ref{fig:test_spatial} and Figure~\ref{fig:test_visual_jigsaw}. Finally, narrative comprehension is assessed in Figure~\ref{fig:test_story}, and image editing capabilities in Figure~\ref{fig:test_edit}.


\begin{figure*}[t]
  \centering
  \includegraphics[width=0.9\linewidth]{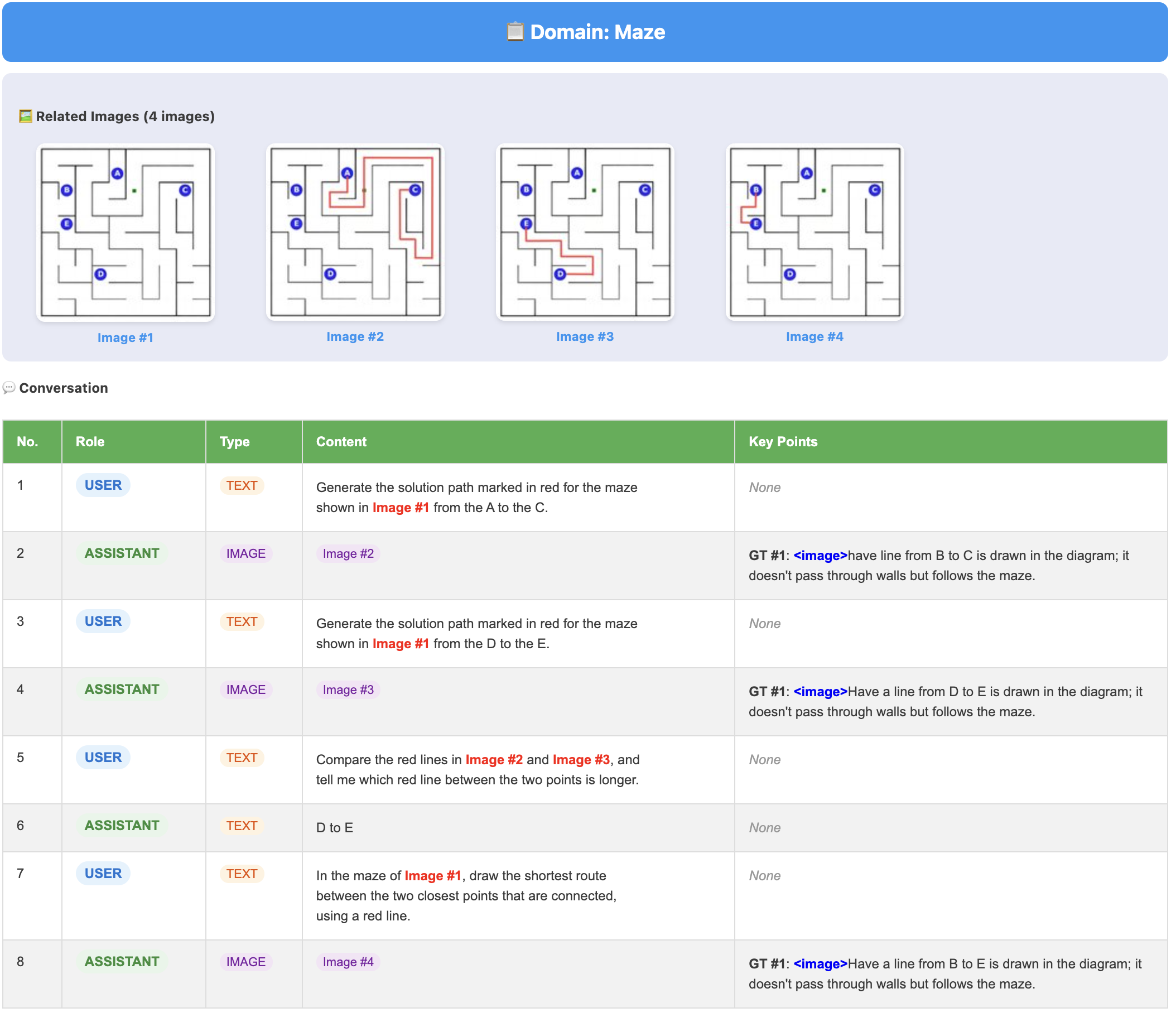}
  \caption{An example of maze-solving task testing the model's pathfinding and spatial reasoning abilities.}
  \vspace{-5mm}  
  \label{fig:test_maze}
\end{figure*}

\section{Broader Impact}\label{app:impact}
The broader impact of \name{} carries both potential benefits and risks upon deployment and release. Some considerations are unique due to the multimodal nature of UMMs while others reflect challenges common to image creation environments. Below, we outline risks and mitigation strategies for its release.

\noindent\textbf{Hallucination.}
Similar to other  models~\citep{deng2025bagel, li2025zebra, bai2024meissonic}, our approach extends and fine-tunes text-to-image generation models to obtain unified generation capabilities, which introduces potential hallucination issues~\citep{ji2023towards,zhao2025robot}. Analogous to existing methods, models trained on \trainname{} may produce outputs that deviate from user intentions or specified input conditions. This phenomenon raises significant concerns, particularly in commercial image applications where purchasing decisions rely on accurate visual representations, given that user requirements and expression modalities exhibit inherent variability. 

\noindent\textbf{Biases.}
Despite implementing human supervision and a multi-model ensemble pipeline to mitigate biases in our synthetically generated dataset, the inherent biases from the foundation models inevitably permeate our data collection process and subsequently propagate to our fine-tuned models. This propagation can yield biased retrieval results and inequitable representations across diverse cultural contexts. Multilingual processing introduces additional bias vectors through language alignment mechanisms, as demonstrated by~\citep{chow2024unified,gallegos2024bias}.

\end{document}